\documentclass[lettersize,journal]{IEEEtran}

\usepackage{amsmath,amsfonts}
\usepackage{algorithmic}
\usepackage{algorithm}

\usepackage{array}
\usepackage{textcomp}
\usepackage{stfloats}
\usepackage{url}
\usepackage{verbatim}
\usepackage{graphicx}
\usepackage[caption=false,font=footnotesize]{subfig}
\usepackage[table]{xcolor}
\usepackage[utf8]{inputenc}
\usepackage{pifont}
\usepackage{pgfplots}
\usepackage{pgfplotstable}

\usepackage{tikz}
\usetikzlibrary{patterns}
\pgfplotsset{compat=1.18}
\usepgfplotslibrary{polar}
\usepackage{tabularx}
\usepackage{booktabs}
\usetikzlibrary{arrows.meta,calc,decorations.text,positioning}
\usepackage[T1]{fontenc}
\usepackage{lmodern}
\usepackage{cite}
\usepackage{amssymb} 
\usepackage{arydshln}
\newcommand{\cmark}{\checkmark}
\newcommand{\xmark}{\times}
\newcommand{\partialmark}{\sim}
\usepackage{multirow}
\usepackage{listings}
\usepackage{xcolor}%
\usepackage[normalem]{ulem}
\usepackage{hyperref}
\usepackage{float}
\usepackage{makecell}
\usepackage{longtable}
\usepackage[most]{tcolorbox}
\usepackage{fontawesome5} 
\usepackage{amsthm}
\theoremstyle{definition}
\newtheorem{definition}{Definition}
\newenvironment{pquote}{%
  \begin{quote}\itshape\raggedright}%
  {\end{quote}}

\newtcolorbox[auto counter,number within=section]{observation}[1]{
    enhanced,
    breakable,
    colback=white,
    frame hidden,         
    borderline west={2pt}{0pt}{blue}, 
    boxsep=2mm,
    left=2mm,
    right=0mm,
    top=1mm,
    bottom=1mm,
    title=\faInfoCircle\ Observation~\thetcbcounter\ -- \parbox[t]{0.45\linewidth}{#1},
    coltitle=blue!70!black,
    fonttitle=\bfseries\itshape,
    attach boxed title to top left={xshift=0mm,yshift=-2mm},
    boxed title style={
        empty,
        frame empty,
        interior empty,
    },
}

\usepackage{amsmath,amssymb,amsfonts}
\usepackage{graphicx}
\usepackage{xcolor}
\usepackage[normalem]{ulem}

\usepackage{ifthen}
\newboolean{showcomments}
\setboolean{showcomments}{true} 
\ifthenelse{\boolean{showcomments}}
  {\newcommand{\nb}[2]{
    \fcolorbox{gray}{yellow}{\bfseries\sffamily\scriptsize#1}
    {\sf\small$\blacktriangleright$\textcolor{teal}{\textit{#2}}$\blacktriangleleft$}
  }
  
  }
  {\newcommand{\nb}[2]{}
  
  }

\begin{document}

\title{Adaptation Needs in Robotic Systems:\\
Assessing Behavior Trees and Their Enhancements}
\author{Mehran Rostamnia, Gianluca Filippone, Ricardo Caldas, Patrizio Pelliccione
\thanks{Mehran Rostamnia, Gianluca Filippone, Ricardo Caldas, and Patrizio Pelliccione are with Gran Sasso Science Institute (GSSI), L'Aquila, Italy
(e-mail: \{mehran.rostamnia, gianluca.filippone, ricardo.caldas, patrizio.pelliccione\}@gssi.it).
Mehran Rostamnia is the corresponding author.
}
}

\markboth{}%
{Rostamnia \MakeLowercase{\textit{et al.}}: Adaptation Needs in Robotic Systems:
Assessing Behavior Trees and Their Enhancements}


\maketitle

\IEEEpubid{%
  \begin{minipage}[t]{\textwidth}
    \mbox{}\\[8pt]
    \centering
    \scriptsize\itshape
    \copyright\ ``This work has been submitted to the IEEE for possible
    publication. Copyright may be transferred without notice, after which
    this version may no longer be accessible.''
  \end{minipage}%
}

\begin{abstract}
Robotic systems increasingly operate in dynamic, uncertain, and open-ended environments, where design-time assumptions may no longer hold, and adaptation becomes necessary to maintain effective and safe operation. Behavior Trees (BTs) are widely used in robotic control architectures due to their modularity, readability, and reactivity. This raises a central question: are BTs sufficient to meet the adaptation needs of modern robotic systems?

This paper investigates this question through a literature-driven study complemented by empirical validation. First, we derive a classification of robotic adaptation needs from the literature,
organizing them into six categories: Knowledge, Perception, Actuation, System, Mission, and Environment. Then, we analyze the capabilities and limitations of classical BTs with respect to these needs. 
Then, we characterize BT-based approaches for adaptation from the existing literature and organize them into four primary families, i.e., generation, extension, evolution, and refinement, including approaches that combine multiple families.

Our analysis shows that the modularity, flexibility, and reactivity of classical BTs are insufficient for adaptation needs involving runtime restructuring, reasoning under uncertainty, mission reinterpretation, learning, or integration with external knowledge and planning mechanisms. Enhanced BT approaches address several of these limitations, but to different extents and often with limitations of their own. Our findings relate adaptation needs to both the capabilities and limitations of classical and enhanced BTs, providing guidance on when classical BTs are sufficient, when enhanced mechanisms are needed, and which challenges remain or emerge for adaptive robotic control architectures.
\end{abstract}


\section{Introduction}

Robotic systems increasingly operate in dynamic, partially observable, and open-ended environments, where environmental changes, uncertainty, and evolving requirements affect their ability to perceive the world, act, and satisfy mission objectives~\cite{weyns2020introduction,loquercio2020general,du2011robot,lauri2022partially,hezavehi2021uncertainty,ramirez2012taxonomy,euRobotics2016MAR,askarpour2021robomax}.
In such settings, assumptions made at design time may no longer hold during operation: sensors may provide noisy or conflicting data, actuators may behave non-deterministically, resources may fail or change, missions may evolve, and humans or other agents may interact with the system in an unpredictable way and modify the environment. 

The violation of design-time premises creates a \textit{need to adapt} the robotic system's behavior, knowledge, configuration, or mission interpretation in order to maintain or improve mission achievement in response to changes~\cite{weyns2020introduction}. Throughout this paper, we refer to the conditions that trigger adaptation as \textit{adaptation needs}.
To cope with such needs, robotic systems require control architectures that allow for reacting to changing conditions at runtime while maintaining correct task execution. In this context, Behavior Trees (BTs) have become a prominent representation of robot behavior because they support modularity, hierarchy, reuse, and enable reactive execution, while offering an intuitive structure for task-level control~\cite{colledanchise2018behavior,iovino2022survey,ogren2022behavior,biggar2020principled}. 
Recent empirical work shows that BTs are used in a variety of robotics applications and, compared to traditional behavior modeling formalisms such as state machines, are particularly well suited for developing modular, reactive, and adaptable robot behaviors~\cite{ghzouli2023behavior,DRAGULE2025101330}. A broader comparative analysis of mission-specification formalisms, including BTs, state machines, Hierarchical Task Networks, and BPMN, further highlights that these formalisms offer different trade-offs in terms of control structure, expressiveness, and support for modeling robotic missions~\cite{Filippone2026Formalisms}. Within this landscape, the modularity and reactivity of BTs make them particularly attractive for structuring robot behavior in systems that must respond to changing execution conditions.

However, the same scenarios that motivate the use of BTs also expose their limitations. Classical BTs typically encode predefined control flows, condition checks, and recovery strategies. Consequently, the complexity of the overall system architecture tends to increase when adaptation requires runtime restructuring, explicit reasoning under uncertainty, mission reinterpretation, learning, or integration with planning and knowledge-management mechanisms. The growing body of work that extends BTs with planning, active inference, adaptive nodes, architectural adaptation, and uncertainty-aware mechanisms suggests that classical BTs alone may be insufficient for several forms of robotic adaptation~\cite{rovida2017extended,pezzato2023active,li2022towards,alberts2024rebet,rostamnia2025towards}.

Existing surveys have examined BT concepts, robotic applications, formal and control-theoretic properties, and their integrations with reinforcement learning and learning from demonstration~\cite{shin2024survey,ghzouli2023behavior,ogren2022behavior,biggar2020principled,iovino2022survey,DRAGULE2025101330}.
Nevertheless, these works do not systematically start from the adaptation needs of real-world robotic systems and analyze whether and how such needs can be addressed by classical BT features, linguistic enhancements of BTs, or online evolutionary algorithms. Consequently, it remains unclear which adaptation needs are naturally tackled by classical BTs, which require extensions or complementary mechanisms, and which remain inadequately addressed in practice and in the scientific literature.

This paper addresses this gap by investigating the following research questions:

\begin{itemize}
    \item \textbf{RQ1:} What adaptation needs characterize robotic systems?
    \item \textbf{RQ2:} To what extent can classical BT features address these adaptation needs, and what limitations emerge?
    \item \textbf{RQ3:} To what extent do approaches and techniques that have been proposed to enhance BTs overcome the limitations of classical BTs in addressing these adaptation needs, and what new limitations emerge?
\end{itemize}

To answer these questions, we first derived a classification of adaptation needs by combining uncertainty classifications from self-adaptive systems with robotics-specific literature. Second, we analyzed BT adaptation or evolution approaches in robotic systems proposed in the literature to identify the features and limitations of classical BTs with respect to those needs. Third, we categorized and assessed these approaches to determine how they compensate for the identified limitations of classical BTs and what limitations or trade-offs they introduce. Finally, we synthesized remaining challenges and identified the capabilities that future behavior modeling formalisms should provide to better support adaptation needs in robotic systems.
The literature-driven analysis is based on a corpus of 31 primary studies on BTs, adaptation, evolution, and robotics. 

The main contributions of this paper are:

\begin{itemize}
    \item A classification of robotic adaptation needs into six categories: Knowledge, Perception, Actuation, System, Mission, and Environment.
    
    \item An analysis of the capabilities and limitations of classical BTs in addressing the identified adaptation needs.
    
    \item A characterization of approaches proposed to enhance BTs for adaptation, including BT generation, extension, evolution, refinement, and hybrid approaches.
    
    \item A systematic mapping and assessment of these approaches against the identified adaptation needs, showing to what extent they overcome the limitations of classical BTs and identifying the limitations that remain or emerge.
    
    \item The identification of research gaps and future directions for BT-based adaptive robotic control architectures.
\end{itemize}

To validate and refine our findings, we complemented the literature-driven analysis with three empirical activities: validation by authors of the primary studies, a questionnaire with robotics and BT researchers and practitioners, and semi-structured interviews with domain experts. We collected feedback on the identified adaptation needs, the suitability of BT-based approaches to address adaptation needs, and the limitations of existing solutions. The responses were analyzed to assess the agreement between our literature-driven classification and practitioners' perspectives. These provided empirical evidence to support the identified challenges and research directions.

Our results show that, although classical BTs provide fundamental software engineering capabilities for adaptive robotic systems-- such as modularity, hierarchical decomposition, reusability, and reactive execution-- they are insufficient to fully support complex adaptation needs arising from uncertainty, changing missions, and dynamic interactions with humans and environments. The analysis reveals that current BT extensions, evolution, generation, and refinement approaches introduce complementary capabilities, but no single approach provides a complete solution. 
These findings highlight the need for future behavior modeling frameworks that treat adaptation as a first-class design concern and provide explicit integration with the planning, reasoning, learning, monitoring, and runtime evolution mechanisms required to realize it.
This work contributes to software engineering research on self-adaptive systems. It uses robotics as a demanding application domain to characterize the adaptation capabilities that behavioral control architectures need. These capabilities allow software-intensive systems to remain effective under changing operational conditions.

The remainder of this paper is organized as follows. 
Section~\ref{sec:background} introduces the background concepts. 
Section~\ref{sec:methodology} describes the research methodology. 
Section~\ref{sub:framework-rq1} presents the classification of adaptation needs for robotic systems, answering RQ1. 
Section~\ref{sub:framework-rq2} analyzes the features and limitations of classical BTs, answering RQ2. 
Section~\ref{sub:framework-rq3} categorizes adaptive BT approaches, answering RQ3. 
Section~\ref{sec:discussion} discusses the findings and open challenges. 
Section~\ref{sec:related-work} positions the paper with respect to related work. 
Finally, Section~\ref{sec:conclusion} concludes the paper.

\IEEEpubidadjcol
\section{Background}
\label{sec:background}

This section introduces the concepts needed to understand the analysis conducted in the paper. We first recall the basic structure of BTs, since they are the object of our review. We then clarify how we use the terms \emph{generation}, \emph{extension}, \emph{evolution}, and \emph{refinement}, because these terms are used to classify the approaches analyzed in RQ3. Finally, we introduce the notion of \emph{adaptation need}, which is central to the taxonomy presented in RQ1. A complete list of definitions adopted in this paper is reported in Appendix~\ref{app:definitions}.

\subsection{Behavior Trees (BTs)}

BTs are hierarchical control structures used to organize the behavior of autonomous agents. A BT represents behavior as a tree of modular nodes. Internal nodes define the control flow, while leaf nodes represent actions or conditions. Common control-flow nodes include sequence, fallback, parallel, and decorator nodes, which determine how actions and conditions are selected and executed~\cite{colledanchise2018behavior,iovino2022survey}.

BTs are widely used in robotics because they provide a readable and modular representation of task-level behavior. Their hierarchical structure allows complex missions to be decomposed into smaller sub-behaviors, while their execution semantics support reactivity through repeated ticking and condition evaluation~ \cite{biggar2020principled,ogren2022behavior}. These properties make BTs attractive for specifying robot behavior in systems that need to react to changing execution conditions.

However, classical BTs provide only limited support for advanced adaptation requirements. In many cases, the adaptation logic must already be encoded in the tree through conditions, fallback branches, or recovery subtrees. When adaptation requires runtime restructuring, reasoning under uncertainty, mission reinterpretation, quality-aware decision making, or integration with planning and learning, classical BTs may need to be extended or combined with additional mechanisms~\cite{biggar2020principled,iovino2022survey,ogren2022behavior}. This motivates the analysis conducted in Sections~\ref{sub:framework-rq2} and~\ref{sub:framework-rq3}.

\subsection{Creating and Modifying Behavior Trees}

In this paper, we distinguish four ways in which BTs can be created or modified to support adaptation: \emph{generation}, \emph{extension}, \emph{evolution}, and \emph{refinement}. This distinction is important because different forms of BT adaptation operate on different inputs and produce different kinds of changes.

\emph{BT generation} refers to processes that produce a BT from a higher-level specification, such as a task model, natural-language instruction, demonstration, plan, or user intent. In generation, there is no initial BT that is taken as the behavior to be modified, although existing BTs, fragments, templates, libraries, or design patterns may be used as prior knowledge to guide the generation process.

\emph{BT extension} refers to processes that change the BT formalism itself. This may include adding new types of nodes, modifying the semantics of existing nodes, introducing decorators, or integrating BT execution with additional reasoning mechanisms. Extension, therefore, changes what can be expressed within the BT language or how BT execution is interpreted.

\emph{BT evolution} refers to processes that start from an initial, often minimal or partial, BT and progressively modify its structure to obtain a tree that encodes more complex or improved behavior. Evolution is commonly associated with search-based or optimization-based techniques, such as genetic programming or grammatical evolution.

\emph{BT refinement} refers to processes that take an existing BT as input and update it to correct, improve, or modify its behavior without changing the underlying BT formalism. Refinement may modify nodes, parameters, subtrees, or execution conditions, and it may occur during design time or at runtime.

These four concepts are used in RQ3 to categorize how the literature enhances BTs for adaptation. Some approaches combine more than one form of change. For example, an approach may first generate an initial BT from a demonstration and then refine it through interaction or optimization.

\subsection{Adaptation Needs in Robotic Systems}

We use the term \emph{adaptation need} to denote a condition indicating that a robotic system should adapt its current behavior, knowledge, configuration, or mission interpretation to maintain or improve mission achievement in response to observed or anticipated changes in the system, its environment, available knowledge, perception, or objectives~\cite{salehie2009self,krupitzer2015survey}.

Adaptation needs may arise from uncertainty, failures, changing resources, mission updates, unreliable perception, non-deterministic actuation, or interactions with humans and other agents. In a nutshell, adaptation needs are the reason that leads a robotic system to perform adaptation.

In this paper, adaptation needs are used as the organizing lens for analyzing BTs. Rather than asking only whether BTs are modular, readable, or reactive, we ask whether these properties are sufficient to address the situations in which robotic systems need to adapt. Section~\ref{sub:framework-rq1} derives a classification of adaptation needs organized into six categories: \emph{Knowledge}, \emph{Perception}, \emph{Actuation}, \emph{System}, \emph{Mission}, and \emph{Environment}. These categories capture the main parts of a robotic system from which adaptation needs may emerge.

\section{Methodology}\label{sec:methodology}

This research aimed not only to review existing literature but also to incrementally construct and validate an artifact that supports the analysis of adaptation needs in robotic systems and their relationship with BTs. In this work, the artifact consisted of a set of structured classifications that progressively captured: (i) adaptation needs in robotic systems, (ii) the extent to which such needs are addressed by BT-based approaches, and (iii) the categories of BT-based adaptation or evolution approaches employed in robotics, and their relation with the addressed adaptation needs.

\begin{figure}
    \centering
    \includegraphics[width=\columnwidth]{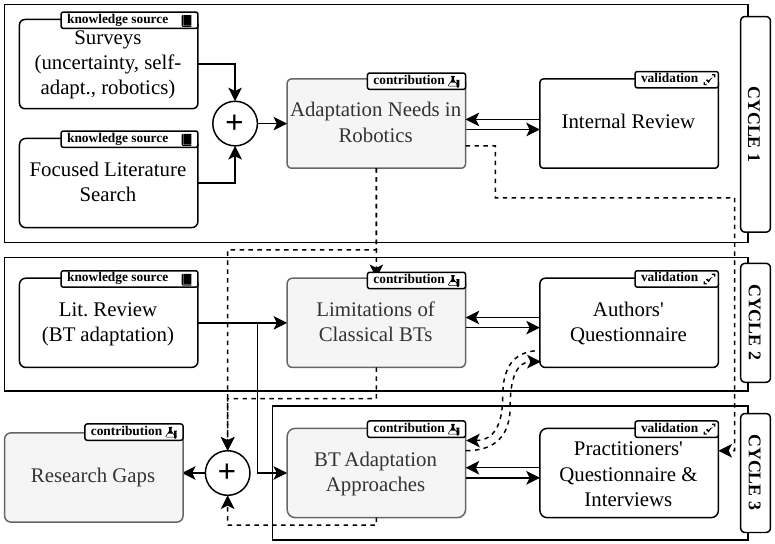}
    \caption{Research design for identifying adaptation needs, assessing classical BTs, characterizing BT-based adaptation approaches, and deriving research gaps.}
    \label{fig:methodology-dsr}
\end{figure}

The research methodology was organized into three iterative cycles. In each cycle, we first identified and framed the problem addressed in the cycle, then we constructed or refined the corresponding part of the artifact. At the end of each cycle, we validated the outcome of the cycle either through internal review or via questionnaires used to determine whether further refinement was needed. In particular, in addition to internal validation by the co-authors, we externally validated the findings through three empirical activities: (i) a targeted author validation in which we asked authors of the considered studies to assess the appropriateness and correctness of the extracted and mapped information; (ii) a questionnaire with researchers and practitioners knowledgeable in robotics and BTs regrading relevancy of adaptation needs and suitability of BTs-based approaches to address them; and (iii) semi-structured interviews with domain experts from both industry and academia to collect qualitative insights based on questionnars' answers. Figure~\ref{fig:methodology-dsr} provides an overview of the methodology adopted in this work.

The cycles are built one on top of another, with each addressing a distinct aspect of the study. Cycle~1 focused on identifying adaptation needs of robotic systems. Cycle~2 investigated how such needs were reflected in the literature on BTs and robotic adaptation. Cycle~3 categorized and assessed the approaches proposed to support robotic adaptation by enhancing or evolving BTs. Finally, the findings in the three cycles were synthesized to identify research gaps and open challenges to adaptation in BT-based robotic systems.

\subsection{Cycle 1: Problem Exploration and Identification of Adaptation Needs}\label{subsec:cycle1}

The first cycle addressed RQ1 by identifying and organizing the major adaptation needs that arise in robotic systems. The goal of this cycle was to build a robotics-specific classification of adaptation needs grounded in both self-adaptive systems research and robotics literature.

We first investigated how uncertainty and adaptation have been conceptualized in self-adaptive systems and robotics. We started from established literature on uncertainty in self-adaptive systems, including studies on uncertainty dimensions and descriptions~\cite{weyns2023towards,calinescu2020understanding}, uncertainty taxonomies~\cite{ramirez2012taxonomy,perez2014uncertainties}, and classification studies in self-adaptive systems~\cite{mahdavi2017classification,hezavehi2021uncertainty,esfahani2013uncertainty}. We also considered robotics-oriented sources, including studies on robotic uncertainty and modeling~\cite{askarpour2021robomax} and the Multi-Annual Roadmap for Robotics~\cite{euRobotics2016MAR}.
The rationale behind this step was that uncertainty is one of the main drivers of adaptation. Robotic systems operate in dynamic, partially observable, and often unpredictable environments. Hence, adaptation needs may arise from several sources, including limitations in knowledge, perception, actuation, system capabilities, mission specification, and environmental conditions.

We constructed an initial classification of adaptation needs for robotic systems. We first extracted uncertainty-related concepts and descriptions from the self-adaptive systems literature. Then, we interpreted these concepts in the context of robotics by identifying the corresponding elements of robotic systems affected by uncertainty.
Finally, to ground the classification in the robotics literature, we conducted targeted searches in Google Scholar, Scopus, and ACM Digital Library. For each initial candidate adaptation needs, we defined specific search strings combining general terms, such as \textit{``robotic''}, \textit{``adaptation''}, and \textit{``uncertainty''}, with terms related to the specific adaptation category under analysis. All search strings are available in the dedicated section of the replication package~\cite{Rostamnia2026Replication}. The purpose of these searches was to identify concrete examples, definitions, and evidence from the robotics literature that could support, refine, or challenge the initial classification.
We included studies that met at least one of the following criteria: (i) they discussed an adaptation need or adaptation strategy in the robotic domain, or (ii) they discussed adaptation in autonomous or self-adaptive systems while providing examples or applications in robotics. Through this process, we consolidated the adaptation needs into six main categories: \emph{Knowledge}, \emph{Perception}, \emph{Actuation}, \emph{System}, \emph{Mission}, and \emph{Environment}.

To validate the proposed classification, we performed an internal review at the end of this cycle. All co-authors independently reviewed the proposed categories, their definitions, and examples collected from the literature. The results were then discussed collectively in iterative review sessions. During these sessions, we assessed whether the classification was internally consistent, whether the categories were sufficiently distinct, and whether relevant adaptation needs were missing or misplaced. This validation led to several refinements. Some categories were redefined to better reflect robotics terminology, while additional needs were incorporated to improve coverage.
At the end of the three cycles, the classification was further assessed through a questionnaire directed to practitioners, in which experts on robotics assessed the proposed categories and evaluated the relevance of the identified adaptation needs to robotic systems. Further details on this validation step are provided in Section~\ref{sub:methodology-validation}.

\subsection{Cycle 2: Literature-Driven Analysis of BT Limitations and Adaptation Needs}
\label{subsec:cycle2}

The second cycle addressed RQ2 by investigating how the adaptation needs identified in Cycle~1 are reflected in the literature on BTs for robotic systems. The goal of this cycle was to understand the extent to which classical BTs are sufficient to address such needs, and why researchers have proposed extensions, adaptations, or alternative BT-based mechanisms.

First, we identified studies that propose, analyze, or evaluate mechanisms for adapting BT-based robotic behavior. Since the terminology used to describe such mechanisms varies considerably across the literature, we avoided constructing the search around specific technique labels such as \emph{generation}, \emph{refinement}, or \emph{extension}. Instead, we used broader terms intended to capture the general concepts of behavioral adaptation and evolution. We inspected the terminology used in studies closely aligned with our research questions and identified \emph{behavior tree/behaviour tree}, \emph{robotics}, \emph{adaptation}, and \emph{evolution} as recurring umbrella terms across different forms of BT change. Based on this analysis, we defined the following search string:

\begin{lstlisting}[language={},basicstyle=\ttfamily\small,frame=single]
("behavior tree" OR "behaviour tree")
AND (adapt* OR evol*)
AND robot*
\end{lstlisting}

The search string deliberately used \texttt{adapt*} and \texttt{evol*} as broad roots rather than enumerating individual adaptation techniques. The purpose was to retrieve studies that explicitly frame modifications to BT-based robotic behavior in terms of adaptation or evolution, while the subsequent screening determined the specific mechanism involved, including generation, extension, evolution, refinement, or combinations thereof. We therefore do not assume that the resulting corpus exhaustively captures every BT-based technique that could potentially support adaptation; rather, it provides the literature corpus used for the systematic analysis conducted in RQ2 and RQ3.
The search was conducted through SciVal and restricted to the robotics domain. The resulting candidate studies were then screened against the objectives of RQ2 and RQ3. We retained studies that propose, discuss, or evaluate a mechanism involving changes to, extensions of, generation of, or evolution of BT-based behavior in robotic systems. This process resulted in 31 primary studies, reported in the replication package~\cite{Rostamnia2026Replication}.

We analyzed the selected studies from three perspectives. First, we extracted the BT features that make BTs suitable for robotic control, such as modularity, reactivity, readability, and compositionality. Second, we identified limitations of classical BTs when they are used to address adaptation needs in robotic systems. Third, we mapped the adaptation needs from Cycle~1 to the selected BT-related studies.

This mapping allows us to examine which adaptation needs are explicitly addressed in the literature and which remain weakly covered. It also helps us identify why classical BTs alone may be insufficient to manage certain robotic adaptation needs. For example, some needs require runtime modification of the tree, integration with learning-based components, explicit handling of uncertainty, or mechanisms for updating knowledge and goals during execution.

The identified limitations of classical BTs and the proposed mapping were subsequently assessed through a questionnaire addressed to the authors of the selected studies, to verify whether our interpretation was consistent with the phenomena and limitations addressed in their work. Further details on this validation are provided in Section~\ref{sub:methodology-validation}.

\subsection{Cycle 3: Technique Characterization}
\label{subsec:cycle3}

The third cycle addressed RQ3 by categorizing the approaches used to enhance the adaptation capabilities of BTs. While Cycle~2 focused on identifying limitations and mapping adaptation needs to the literature, Cycle~3 analyzed how existing studies attempted to overcome these limitations in addressing adaptation needs.

In the first phase, we reused the corpus of primary studies identified in Cycle~2 and focused on the adaptation mechanisms proposed in each study.
We extracted technical information about the proposed solutions in the studies, including the techniques used to support adaptation, the type of robotic system considered, the adaptation triggers, the adaptation goals, and the level of human involvement.
Furthermore, we examined whether these studies modify the BT structure, extend the BT semantics, generate BTs from external specifications, evolve BTs through search-based techniques, refine existing BTs, or combine BTs with other methods such as planning, learning, runtime monitoring, or knowledge representation.

In the next phase, we categorized the approaches used in the selected studies. For each study, we extracted and organized information along the following dimensions: the type of BT modification, the adaptation need addressed, the adaptation trigger, the adaptation goal, the timing of adaptation, the degree of autonomy, and the level of human involvement. Then, we examined the extent to which the approaches were suitable to tackle adaptation needs in robotic systems. The document containing this technical information is available in the replication package~\cite{Rostamnia2026Replication}.

This analysis allows us to distinguish among different families of approaches, i.e., generating the BT, evolving it with new behaviors according to observed changes, refining it by updating nodes, or extending the formalism itself with new control structures.

The resulting characterization provides a structured view of how BT-based approaches currently support adaptation in robotic systems and where important gaps remain.
The characterization was assessed through expert validation, combining feedback from the authors of the primary studies on the assigned BT adaptation classes (from Cycle 2) with practitioners' assessments of the suitability of these classes to address the identified adaptation needs, as explained below.

\subsection{Validation}
\label{sub:methodology-validation}

At the end of Cycle 3, we validated the findings in three complementary steps. First, we contacted the authors of the primary studies to validate whether our mapping of their techniques to BT generation, extension, evolution, and refinement was appropriate. Second, we conducted a questionnaire with researchers and practitioners with expertise in robotics, BTs, autonomous systems, or related areas. This step aimed to empirically assess whether the adaptation needs identified in the literature were perceived as relevant, and whether the four classes of BT adaptation techniques were perceived as suitable to address them. Third, following the analysis of the practitioners' questionnaire responses and internal discussion among co-authors, we ran a round of in-depth semi-structured follow-up interviews.

\subsubsection{Hypotheses}

The validation was organized around six hypotheses.

\begin{description}
    \item[H1.] The mapping of primary studies to the four BT adaptation classes is consistent with the interpretation of the authors of those studies.

    \item [H2.] The limits of classical BTs to adaptation are correctly identified in the scientific literature.

    \item[H3.] The mapping of primary studies to the adaptation needs is consistent with the interpretation of the authors of those studies.
    \item[H4.] The classified adaptation needs are relevant in real robotic application scenarios.

    \item[H5.] Classical BTs are sufficient for addressing some adaptation needs, whereas others require BT enhancements or complementary mechanisms.

    \item [H6.]  The four categories of BT adaptation techniques, namely BT generation, BT extension, BT evolution, and BT refinement, are suitable to address at least some of the identified adaptation needs.
\end{description}

 Hypotheses H1--H3 were evaluated through the \emph{Author Questionnaire}, where we collected targeted feedback from the authors of the primary studies. Hypotheses H4--H6 were evaluated through the \emph{Practitioners' Questionnaire} and semi-structured interviews. These hypotheses provide validation evidence for the research questions as follows. H3 and H4 support RQ1 by assessing the literature and practitioners' points of view regarding the relevance of the adaptation needs classification. H2 and H5 support RQ2 by evaluating identified limitations and the sufficiency of BTs to solve adaptation needs. H1 and H6 support RQ3 by validating the classification of BT-based techniques and their suitability to address different adaptation needs. H3 also contributes to RQ3 by validating the link between primary studies and the adaptation needs they address.

\subsubsection{Author Questionnaire} 

To validate whether our classification of the papers collected in Cycle~2 reflects the authors' intent, we performed an author-based validation. For each primary study, we contacted one or more of the selected papers' authors and provided them with: (i) the identified limitations of classical BTs discussed in their paper, (ii) the adaptation class or classes assigned to it, (iii) the definitions of adaptation needs classes, (iv) a short explanation of our interpretation, and (v) the definitions of BT generation, BT extension, BT evolution, and BT refinement. We then asked the authors whether they agreed with the identified limitation, the assigned adaptation class, whether another class would be more appropriate, or whether the paper should be mapped to multiple classes.

We used the authors' replies to evaluate H1--H3. Each hypothesis was evaluated through a corresponding part of the validation. To evaluate H1, we asked authors whether their studies were correctly assigned to the proposed BT-based adaptation classes. To assess H2, we asked whether the extracted classical BT limitations accurately reflected the limitations discussed or implied in the studies and whether any limitation had been omitted or misinterpreted. To evaluate H3, we asked the authors whether the assigned adaptation needs accurately represented the needs addressed by the study and whether any assigned need should be removed or replaced. 

Author validation forms were sent to the first authors and, in the case of no reply from the first author, to co-authors. We received 13 responses out of 31 primary studies  (41.9\%). If the authors confirmed our classification, we marked the mapping as author-confirmed. If they disagreed, we revisited the paper and revised the mapping when their explanation showed that our interpretation was incomplete or inaccurate. If no reply was received, we kept the mapping based on our analysis, but we did not treat the absence of a reply as confirmation. Author validation was used as a member-checking procedure, not a statistically representative survey. The responses allowed us to identify and correct interpretation errors based on feedback from researchers who had direct knowledge of the related studies. The incomplete response coverage and the possibility of non-response bias were reported as threats to validation.

Based on the responses, the overall feedback largely supports the proposed mappings while providing several clarifications that improved our classifications. Most authors agreed with our characterization of the limitations of classical BTs. Regarding the classification of adaptation needs, the received feedback resulted in only minor refinements rather than substantial changes to the classification. The forms dedicated to each paper are available in the replication package~\cite{Rostamnia2026Replication}.

\subsubsection{Practitioners' Questionnaire}

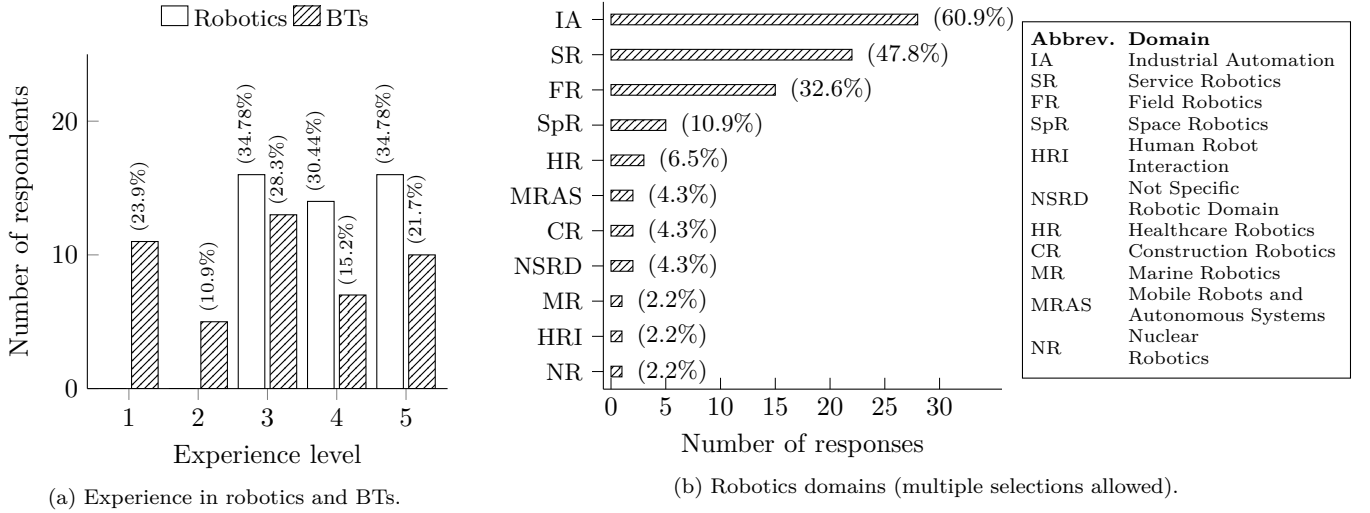
\begin{figure*}[t]
\centering

\begin{minipage}[t]{0.35\textwidth}
\centering
\subfloat[Experience in robotics and BTs.\label{fig:experience}]{
\begin{tikzpicture}
\begin{axis}[
    ybar,
    width=\linewidth,
    height=6cm,
    ymin=0,
    ymax=25,
    ylabel={Number of respondents},
    xlabel={Experience level},
    symbolic x coords={1,2,3,4,5},
    xtick=data,
    enlarge x limits=0.15,
    bar width=10pt,
    axis x line*=bottom,
    axis y line*=left,
    grid=none,
    legend style={
        at={(0.5,1.04)},
        anchor=south,
        draw=none,
        legend columns=2,
        font=\small
    },
    nodes near coords,
    every node near coord/.append style={
        font=\scriptsize,
        text=black,
        rotate=90,
        anchor=west,
        yshift=2pt
    },
    point meta=explicit symbolic,
]

\addplot+[
    draw=black,
    fill=white
]
coordinates{
(1,0)  []
(2,0)  []
(3,16) [(34.78\%)]
(4,14) [(30.44\%)]
(5,16) [(34.78\%)]
};

\addplot+[
    draw=black,
    pattern=north east lines,
    pattern color=black,
]
coordinates{
(1,11) [(23.9\%)]
(2,5)  [(10.9\%)]
(3,13) [(28.3\%)]
(4,7)  [(15.2\%)]
(5,10) [(21.7\%)]
};

\legend{Robotics, BTs}

\pgfplotsset{
    legend image code/.code={
        \draw[#1] (0cm,-0.08cm) rectangle (0.35cm,0.18cm);
    }
}

\end{axis}
\end{tikzpicture}
}
\end{minipage}
\hfill
\begin{minipage}[t]{0.63\textwidth}
\centering
\subfloat[Robotics domains (multiple selections allowed).\label{fig:domain}]{
\begin{tikzpicture}
\begin{axis}[
    xbar,
    clip=false,
    enlarge x limits=0.02,
    width=0.6\linewidth,
    height=6.7cm,
    xmin=0,
    xmax=35,
    xtick={0,5,10,15,20,25,30},
    xlabel={Number of responses},
    symbolic y coords={
        NR,
        HRI,
        MR,
        NSRD,
        CR,
        MRAS,
        HR,
        SpR,
        FR,
        SR,
        IA
    },
    ytick=data,
    yticklabel style={font=\small,align=right},
    enlarge y limits=0.05,
    axis x line*=bottom,
    axis y line*=left,
    tick style={black},
    bar width=4pt,
    nodes near coords,
    point meta=explicit symbolic,
    every node near coord/.append style={
        anchor=west,
        font=\small,
        xshift=2pt
    }
]
\addplot[
    draw=black,
    fill=white,
    pattern=north east lines
]
coordinates {
    (1,HRI)  [(2.2\%)]
    (1,NR)   [(2.2\%)]
    (2,MRAS) [(4.3\%)]
    (2,CR)   [(4.3\%)]
    (2,NSRD) [(4.3\%)]
    (3,HR)   [(6.5\%)]
    (15,FR)  [(32.6\%)]
    (5,SpR)  [(10.9\%)]
    (1,MR)   [(2.2\%)]
    (22,SR)  [(47.8\%)]
    (28,IA)  [(60.9\%)]
};
\node[
    anchor=north west,
    font=\scriptsize,
    align=left,
    draw=black,
    fill=white,
    inner sep=3pt
] at (rel axis cs:1.05,0.95) {
\begin{tabular}{@{}p{.87cm}l@{}}
\textbf{Abbrev.} & \textbf{Domain} \\
IA   & Industrial Automation \\
SR   & Service Robotics \\
FR   & Field Robotics \\
SpR  & Space Robotics \\
HRI & \makecell[l]{Human Robot\\ Interaction} \\
NSRD & \makecell[l]{Not Specific\\ Robotic Domain}\\
HR   & Healthcare Robotics \\
CR    & Construction Robotics \\
MR   & Marine Robotics \\
MRAS & \makecell[l]{Mobile Robots and\\ Autonomous Systems} \\
NR   & \makecell[l]{Nuclear\\ Robotics} \\
\end{tabular}
};
\end{axis}
\end{tikzpicture}
}
\end{minipage}

\caption{Participants' background (46 respondents). (a) Self-reported experience in robotics and BTs. (b) Robotics domains in which respondents have worked. Percentages in (b) do not sum to 100\% because respondents could select multiple domains.}
\label{fig:participant-background}

\end{figure*}

The practitioners' questionnaire aimed to assess the perceived relevance of the identified adaptation needs in robotic systems, the extent to which classical BTs were considered sufficient to address them, and the suitability of the four BT adaptation classes for tackling such needs.
The collection of data from the practitioners' questionnaire was carried out over a period of four weeks in July 2026. 
We distributed the questionnaire via Open Robotics Discourse\footnote{\url{https://discourse.openrobotics.org}}, speakers whose presentations at ROSCon talks were specifically dedicated to BTs\footnote{\url{https://roscon.ros.org}}, LinkedIn, and robotics-related online groups. We also directly invited researchers and practitioners working with BTs. We identified authors of scientific work focusing on BT adaptation through a search conducted on Scopus using the following query:
\begin{quote}
\small
(\text{``Behavior tree*''} OR \text{``Behaviour tree*''}) AND
\text{robotic*} AND (\text{generation} OR \text{evolution} OR \text{refinement} OR \text{generate} OR \text{evolve})
\end{quote}
We focused on papers published between 2024 and June 2026 that mentioned BTs in their titles, abstracts, or keywords. The papers identified through this process are listed in the replication package~\cite{Rostamnia2026Replication}. We also contacted participants in the author validation phase who had agreed to take part in a follow-up study. In total, we invited 87 potential participants via email, and we contacted 116 professionals through LinkedIn based on their reported expertise in BTs.
Overall, the questionnaire received 46 responses.
Figure~\ref{fig:participant-background} provides an overview of the questionnaire participation profile. Figure~\ref{fig:experience} presents the experience of the participants in robotics and BTs. The levels of experience range from 1 to 5, where 1 denotes individuals with little or no prior experience ($<$ 1 year); 3 denotes individuals with some practical or theoretical familiarity (1 -- 3 years); and 5 denotes substantial experience, typically several years of study, research, or practice ($>$ 3 years). In general, the respondents demonstrate strong robotics expertise, with 100\% reporting intermediate to expert experience (level 3--5). In contrast, the experience with BTs was more diverse. A group of respondents reported intermediate to expert experience (65.2\%). However, a portion of respondents reported limited experience (23.9\%). The distribution is reasonable because the first part of the questionnaire focused on general adaptation needs and, therefore, requires considerable robotic expertise. By contrast, participants who were less familiar with BTs could claim \emph{insufficient expertise} for the BT-related parts. The results indicate that the survey captures opinions of participants with substantial robotics expertise while representing a broad spectrum of familiarity with BTs.

We also collected the experience of the respondents across multiple application areas. Figure~\ref{fig:domain} reports the robotics domain in which the participants have worked. The respondents could select multiple domains; therefore, the percentages do not sum to 100\%. Industrial Automation was the most common domain (60.9\%), followed by Service Robotics (SR) with 47.8\% and Field Robotics (FR) with 32.6\%. Each remaining domain was represented by no more than 10.9\% of the respondents. 
Overall, the results indicate that the participant pool is composed of perspectives from various specialized application domains, thereby providing a broad and diverse view of the adaptation needs in robotic systems.

The questionnaire was structured into three main parts. The first part collected background information about the respondents as reported above, including their experience with robotics, their experience with BTs, the type of organizations in which they have worked, and the robotics domains they are familiar with.

The second part concerned the relevance of the adaptation needs. For the purpose of the questionnaire, we organized the questions around the categories of adaptation needs we identified in the scope of RQ1. 
For each adaptation need, respondents were asked to rate its relevance to robotic missions. The rating scale ranged from \emph{not relevant} to \emph{extremely relevant}. We also included an \emph{insufficient expertise} option to avoid forcing respondents to provide judgments outside their expertise. Additionally, respondents were asked whether and how each family of adaptation needs should primarily be addressed using BTs. They could select \emph{Using classical BTs is enough}, \emph{Mainly by extending the BT formalism}, \emph{Mainly by updating the BT, without changing the formalism}, \emph{Mainly by adapting external modules/models}, and \emph{By both adapting BT formalism and external modules}. We also included the \emph{None of the above} option and asked them to provide their own solution. 

The third part concerned the suitability of the adaptation techniques. We presented the definitions of \emph{BT generation}, \emph{BT extension}, \emph{BT evolution}, and \emph{BT refinement}. For each technique, respondents were asked to rate its suitability for addressing each of the five questionnaire families defined above. 
The rating scale ranged from \emph{not suitable} to \emph{completely suitable}, again including an \emph{insufficient expertise} option. 

At the end of each section, we provided open-ended questions to allow respondents to explain ratings, report disagreements, or identify missing needs or techniques.

We analyzed the closed questions quantitatively and the open-ended answers qualitatively. For the quantitative analysis, we treated the Likert-scale answers as ordinal data. We reported the distribution of answers for each adaptation need and for each technique--need pair. Answers marked as \emph{insufficient expertise} were not converted into numerical scores; instead, we reported them separately.

To evaluate H4, we inspected whether the adaptation needs received consistently positive relevance ratings. 
To evaluate H5, we inspected the type of BT-based solution each family of adaptation needs received. To evaluate H6, we compared the suitability profiles of the four techniques across the different adaptation-need families. This allowed us to identify whether respondents distinguished between the roles of generation, extension, evolution, and refinement.

We used the open-ended answers to interpret the quantitative results. In particular, we used them to identify cases where respondents disagreed with a category, considered a need ambiguous, or suggested that a technique is suitable only under specific assumptions.

\begin{table*}[!b]
    \centering
    \caption{Knowledge-driven adaptation needs.}
    \label{tbl:adapt_needs_knowledge}
    \small
    \begin{tabular}{
    p{0.3cm}|p{17cm}}
%
\multirow{4}{*}{\rotatebox[origin=c]{90}{\makebox[3.5cm][c]{Abstraction}}}
&  
 \begin{itemize}
    \item {\bf Definition:} Knowledge abstraction refers to the omission or oversimplification of real-world details of the robotic system itself, its operational environment, or the robot-environment interaction, in the knowledge representation maintained by the system~\cite{street2023formal,weyns2023towards,askarpour2021robomax}. Knowledge abstraction is one of the reasons for the reality gap~\cite{aljalbout2025reality}.
    \item {\bf Trigger}: Triggered by a discovered mismatch between the abstracted knowledge and the actual system state. 
    \item {\bf Object(s) of Adaptation:}  Robot's knowledge representation to reduce the reality gap~\cite{aljalbout2025reality}, the unpredictability of the robot action's outcome, or in its assessment of the current system environment state.
    \item {\bf Example:} To simplify the robot-world interaction knowledge, details of the wheel-ground contact may be omitted, introducing model mismatch that can lead to accumulated localization (pose) drift during mission execution~\cite{shuai2020research}.
  \end{itemize}
 \\
 \hline
 \multirow{4}{*}{\rotatebox[origin=c]{90}{\makebox[4.5cm][c]{Incompleteness}}}
 & 
 \begin{itemize}
    \item {\bf Definition:}  Knowledge incompleteness refers to the lack or unavailability of some required piece of information about the robotic system itself, its operational environment, or the actions/tasks to be performed~\cite{cornejo2020survey}. Knowledge incompleteness is one of the reasons for the reality gap~\cite{aljalbout2025reality}.
    \item {\bf Trigger}: Triggered by direct observation (e.g., through sensors) or inference mechanisms (e.g., analysis of its own behavior, benchmarking against expectations).
    \item {\bf Object(s) of Adaptation:} The robot's knowledge to reduce unpredictable system behavior, which may be consequent to strange behaviors, mission failure, or system error.
    \item {\bf Example:} A rover may encounter a new type of terrain that was not included in its original real-world environmental knowledge~\cite{rostamnia2025towards}. Since the terrain characteristic is unknown, the rover cannot accurately predict the effects of its actions (e.g., slippage).
  \end{itemize}
 \\
 \hline
 \multirow{4}{*}{\rotatebox[origin=c]{90}{\makebox[4.2cm][c]{Inaccuracy}}}
& 
 \begin{itemize}
    \item {\bf Definition:} Knowledge inaccuracy refers to a difference between the knowledge maintained by the robot and the actual state of the robotic system itself, its operational environment, or the robot-environment interaction, stemming from an inaccurate implementation of knowledge~\cite{davchev2022residual}.
    Knowledge inaccuracy is one of the reasons for the reality gap~\cite{aljalbout2025reality}.
     \item {\bf Trigger}: Triggered by inconsistencies between the predicted outcome based on the robot's knowledge and the actual observed outcomes in the real world.
    \item {\bf Object(s) of Adaptation:} The robot’s internal knowledge about system properties, environment characteristics, or task-related parameters used for perception, planning, and control.
    \item {\bf Example:} A robot may grasp an object using an inaccurate estimation of object mass or friction properties~\cite{feng2020center}. Since the internal knowledge does not represent the real object accurately, it may slip or fall from the robot's gripper. 
  \end{itemize}
 \\
 \hline
 \multirow{4}{*}{\rotatebox[origin=c]{90}{\makebox[4.2cm][c]{Drift}}}
& 
  \begin{itemize}
    \item {\bf Definition:} Knowledge drift refers to the mismatch between the knowledge maintained by the robot and the actual state of the robotic system or the physical world in which it operates~\cite{askarpour2021robomax}. It occurs when the initial assumptions about the knowledge used by the robot become outdated due to changes in the system, environment, or operational context over time.
    Knowledge drift is one of the reasons for the reality gap~\cite{aljalbout2025reality}. 
    \item {\bf Trigger}: Triggered by direct observation (e.g., through sensors) or inference mechanisms (e.g., analysis of own behavior, benchmark with expectations). 
    \item {\bf Object(s) of Adaptation}: Knowledge to reduce the reality gap~\cite{aljalbout2025reality}, different strategies, controllers, etc., to compensate for potential behavior degradations. 
    \item {\bf Example:} Knowledge drifts may manifest when a robot’s internal map indicates a corridor is free while its sensors detect an unexpected obstacle that had been added by the robot itself or other environmental factors during the mission execution, revealing a mismatch between the robot’s model and real-world phenomena and leading to degraded or unreliable behavior~\cite{qiao2024simultaneous}.
  \end{itemize} \\
\end{tabular}
\end{table*}

The validation results were used to refine the study and qualify our conclusions. Relevance ratings indicated which adaptation needs are broadly recognized by experts and which ones may be more domain-specific or less clearly formulated. Suitability ratings indicated how the four BT adaptation classes relate to different families of needs. Author feedback was used to improve the accuracy of the mapping between primary studies and adaptation classes.

We did not discard an adaptation need solely because it received lower ratings. A lower rating may indicate that the need is less common, less mature, or relevant only in specific robotic domains. Instead, we used the validation to identify where the classification is well supported and where further clarification or empirical evidence is needed.

\subsubsection{Semi-structured interviews}
To provide further insights into the results obtained from the practitioners' questionnaire, we conducted semi-structured interviews.
Participants were selected from the respondents in the practitioners' questionnaire according to (i) self-declared level of expertise, particularly individuals who rated their experience in robotics and BT expertise as 5 out of 5; (ii) the presence of particularly critical or insightful responses; and (iii) explicit consent to be contacted for follow-up questions.

We interviewed six BT experts in robotic systems and BTs. At the time of the interviews, two worked primarily in academia, both with seven years of experience in robotics, while four worked primarily in industry with between 5 and 20 years of experience. The interviewees had worked with a wide range of robotic domains and platforms. One of the academic participants focused on software engineering for robotic systems, including BTs, state machines, planning, quality, and safety, rather than on a specific robotic platform. The other one worked with service and industrial robots, with a particular focus on task planning, BT generation and learning, and collaborative manipulation. The four participants working in industry had varied backgrounds. One had worked on BT tooling and production systems involving automated vehicles, aerial robots, and humanoids. Another one had developed ROS~2-based task and mission management systems for mobile robots. The remaining two had experience in industrial manipulation, humanoid and quadruped robots, and the design of BT-based architecture for service, space, and industrial robotics.

The main part of the interview focused on four topics: (i) the relevance of the identified adaptation needs; (ii) the extent to which classical BTs can address these needs and which capabilities require external components; (iii) the suitability, strengths, and limitations of BT evolution, extension, generation, and refinement; and (iv) practical examples, unresolved challenges, and directions for future research. According to the semi-structured nature of the interviews, we asked follow-up questions to clarify the participants' reasoning and elicit examples from their experience. We used the insights provided by interviewees to further refine the conclusions of the study and to draw observations on the use of BTs across domains and adaptation needs.

\subsubsection{Threats to Validity}
Following the validity framework of Wohlin et al.~\cite{wohlin2012experimentation}, we discuss threats to validity of our study in terms of internal, external, and construct validity.

\emph{Internal validity} refers to the level of influence that extraneous variables may have on the design of the study. Since we did not investigate a causal treatment, the main internal threats related to researcher and respondent bias. The identification of adaptation needs, extraction of BT limitations, and mapping of primary studies to adaptation needs and BT-based technique classes required researcher interpretation. Thus, prior expectations may have influenced the extraction and classification process. To address this threat, we developed explicit definitions and classification criteria, conducted an internal review of the extracted information by co-authors, documented the rationale behind the mappings, and validated the mappings with the authors of the primary studies. When the authors suggested corrections, we re-examined the corresponding paper and checked it against our definitions before revising a mapping. Additionally, questionnaire responses may have been influenced by question wording, presentation order, fatigue, or respondents' prior familiarity with particular robotics domains or BT techniques. Similarly, responses to interview questions and their interpretation may have been influenced by the interviewer. We reduced these risks by providing participants with the same definitions and explanatory material, allowing them to indicate insufficient expertise, and conducting the interviews using a semi-structured protocol.

\emph{External validity} refers to the level of generalizability of the studies and participants included in the investigation. 
The search strategy may introduce terminology bias, since relevant approaches may support forms of adaptation without explicitly using terms derived from adaptation or evolution. We mitigated this risk by deriving the search terminology from studies aligned with our research questions and by complementing the corpus analysis with targeted literature searches used in the construction of the adaptation-needs classification. Nevertheless, the resulting corpus should not be interpreted as an exhaustive enumeration of all BT-based mechanisms that may contribute to robotic adaptation. 
Moreover, the practitioner sample may not represent all robotics domains equally. We mitigated this threat by recruiting participants from different communities and by reporting the domains represented in the responses. Author validation covered 13/31 studies (41.9\%). These responses provided direct validation evidence only for the studies whose authors responded. We did not generalize them to the other studies and did not treat non-response as confirmation. Similarly, we used interviews to obtain detailed insights, not intending them to be statistically representative findings. 

\emph{Construct validity} refers to whether the definitions and measurements used are suitable to answer research questions. Concepts such as \emph{adaptation need}, \emph{classical BT}, and the four BT-based approach classes may be interpreted differently. Furthermore, a study may exhibit characteristics of multiple technique classes or address several adaptation needs. To mitigate this, we provided explicit definitions, examples, and classification criteria. We allowed studies to be assigned to multiple classes and needs and asked the authors of the primary studies to assess our interpretations. To simplify the judgments, we defined relevance and suitability through questionnaire ratings. The experience reported by participants was only an approximation of their actual expertise. We addressed this issue by collecting background information and providing an \emph{insufficient expertise} option. 

\section{Adaptation Needs of Robotic Systems (RQ1)}
\label{sub:framework-rq1}

\begin{table*}[!b]
    \centering
    \caption{Perception-driven adaptation needs.}
    \label{tbl:adapt_needs_perception}
    \small
    \begin{tabular}{
    p{0.3cm}|p{17cm}}
 \multirow{4}{*}{\rotatebox[origin=c]{90}{\makebox[4.3cm][c]{Multiplicity}}}
& 
 \begin{itemize}
    \item {\bf Definition:} Perception multiplicity refers to heterogeneity and possible inconsistency of information coming from multiple sources of information (e.g., cameras, sensors, LiDAR), which may represent the same real-world entities or events in different ways or provide conflicting observations~\cite{zhang2022dynamic}.
    \item {\bf Trigger:} Triggered by the detection of inconsistent or conflicting information about the same entity, event, or state when combining observations from multiple sensors or information sources.
    \item {\bf Object(s) of Adaptation:} The robot’s data fusion mechanisms and internal knowledge used to integrate and reconcile information from multiple sensing sources.
    \item {\bf Example:} A robot may receive different estimations of the pedestrians' positions from its camera and LiDAR sensors~\cite{Berrio2022camera}. Since the two perception sources perceive and represent the environment differently, their observations may conflict. 
\end{itemize}
\\
\hline
 \multirow{4}{*}{\rotatebox[origin=c]{90}{\makebox[4.2cm][c]{Inaccuracy}}}
 &  
 \begin{itemize}
    \item {\bf Definition:} Perception inaccuracy refers to the discovery of incomplete, noisy, or unreliable observations of the environment, which stems from the inherent imperfection of sensors. This inaccuracy can stem either from internal stimuli (the system itself) or external stimuli (the system detects a problem that does not exist, or fails to detect a problem that actually occurs)~\cite{rabiee2023introspective}.
    \item {\bf Trigger:} Triggered by the detection of unreliable, noisy, or incomplete sensor observations, which may originate from internal sensing issues or external conditions that affect the sensor’s ability to correctly detect environmental states or events.
     \item {\bf Object(s) of Adaptation:} The robot’s perception and state estimation processes, including sensor interpretation, filtering, and decision-making mechanisms that rely on sensed data.
    \item {\bf Example:} A robot relying on LiDAR for obstacle detection may generate false positives due to sensor noise or fail to detect transparent obstacles, which results in state estimation unpredictability and degraded navigation performance~\cite{corso2022risk,elfes2002using}. 
\end{itemize}
 \\
 \hline
\multirow{4}{*}{\rotatebox[origin=c]{90}{\makebox[4.2cm][c]{Unavailability}}}
 & 
 \begin{itemize}
    \item {\bf Definition:} Perception unavailability refers to the unavailability of perception sources, such as cameras, sensors, and LiDAR, to maintain situational awareness and operational performance~\cite{mulubika2025approach}.
     \item {\bf Trigger:} Triggered by loss, failure, or temporary unavailability of one or more perception sources that normally provide information required for the system itself/environment perception, localization, or mapping.
    \item {\bf Object(s) of Adaptation:} The robot’s perception and sensing pipeline, including alternative sensing strategies, state estimation mechanisms, or operational behaviors that compensate for missing perception inputs.
    \item {\bf Example:} When cameras stop providing usable data due to sensor unavailability, the robot cannot continue to update its map or localize itself, leading to increased unpredictability in environment estimates and degraded operational performance~\cite{mulubika2025approach}.
\end{itemize}
\\
\end{tabular}
\end{table*}

\begin{table*}
    \centering
    \caption{Actuation-driven adaptation needs.}
    \label{tbl:adapt_needs_actuation}
    \small
    \begin{tabular}{
    p{0.3cm}|p{17cm}}
\multirow{4}{*}{\rotatebox[origin=c]{90}{\makebox[4.2cm][c]{Non-determinism}}}
& 
  \begin{itemize}
    \item {\bf Definition:} Non-determinism in actuation refers to non-deterministic behavior of the actuators, where commanded actions do not always produce the predicted outcomes. Mechanical imperfections and physical effects can introduce variability in how actions translate into real-world effects, which may change the world in unpredictable ways~\cite{euRobotics2016MAR,cully2015robots,phillips2020planning}.
    \item {\bf Trigger:} Triggered by observed deviations between commanded actuator actions and the actual physical outcomes produced by the robot’s actuators.
    \item {\bf Object(s) of Adaptation:} The robot’s control strategies, motion execution models, and action outcome predictions that depend on actuator behavior.
    \item {\bf Example:} When the controller of a robotic arm commands a small change in joint angle, the actual motion may vary because of backlash, stiction, and variable friction inside the gears and bearings~\cite{guida2022simulation}. This variability can cause the arm to reach slightly different positions than intended.
\end{itemize}
 \\
\hline
\multirow{4}{*}{\rotatebox[origin=c]{90}{\makebox[4.2cm][c]{Unavailability}}}
&  
 \begin{itemize}
    \item {\bf Definition:} Actuation unavailability refers to the unavailability or failure of actuators, which prevents the robot from executing commanded actions as intended and may compromise its ability to perform tasks~\cite{panda2025integrating}. 
     \item {\bf Trigger:} Triggered by the detection of an actuator failure or loss of actuator functionality, resulting in the robot’s inability to execute planned motions or actions as expected.
    \item {\bf Object(s) of Adaptation:} The robot’s control strategies, motion planning, and task execution mechanisms that rely on the availability and functionality of actuators. 
    \item {\bf Example:} In a legged robot, the failure of an actuator may affect the execution of a commanded motion. In this case, the failed leg moves passively while the robot's body undergoes unintended displacement, leading to uncertainty in task execution and requiring adaptation in motion strategy or gait~\cite{chen2015kinematic}. 
\end{itemize}
 \\
\end{tabular}
\end{table*}

Robotic systems rarely operate under fixed, fully known, and predictable conditions. Instead, they are exposed to uncertainty originating from their internal state, their sensing and actuation capabilities, their mission objectives, and the environment in which they operate. In this context, adaptation becomes necessary when the current behavior, configuration, or knowledge of the robotic system is no longer adequate to satisfy the mission under the observed operating conditions.

To answer RQ1, we identified the main reasons why a robotic system may need to engage in adaptation. We used uncertainty as the main lens for this analysis, since uncertainty directly affects how a robot perceives the world, reasons about its mission, selects actions, and evaluates the effects of those actions. In particular, we built on prior work on uncertainty in self-adaptive systems and robotic systems~\cite{weyns2020introduction,euRobotics2016MAR}, and we used the classification schema proposed by Mahdavi-Hezavehi et al.~\cite{mahdavi2017classification} as a starting point for identifying sources of uncertainty.

While existing uncertainty classifications provide a useful conceptual basis, they are not specific enough to directly capture the operational needs of robotic systems. Therefore, we reinterpreted uncertainty sources from a robotics perspective and derived a classification of adaptation needs. Each adaptation need describes a situation in which the robotic system may have to modify its knowledge, behavior, configuration, planning, perception, actuation strategy, or mission interpretation in order to continue operating effectively.

The resulting classification is organized into six categories: \emph{Knowledge}, \emph{Perception}, \emph{Actuation}, \emph{System}, \emph{Mission}, and \emph{Environment}. These categories correspond to the main elements that influence the behavior of a robotic system. Knowledge captures what the robot believes about itself and the world. Perception concerns how the robot obtains information from sensors and external sources. Actuation concerns how the robot physically executes actions. System refers to the internal resources, components, and communication mechanisms that support operation. Mission captures the objectives, the tasks to be executed, and quality goals assigned to the robot. Finally, Environment refers to the physical and social context in which the robot operates.

The classification is reported across six tables, each corresponding to one category of adaptation needs: Knowledge in Table~\ref{tbl:adapt_needs_knowledge}, Perception in Table~\ref{tbl:adapt_needs_perception}, Actuation in Table~\ref{tbl:adapt_needs_actuation}, System in Table~\ref{tbl:adapt_needs_system}, Mission in Table~\ref{tbl:adapt_needs_mission}, and Environment in Table~\ref{tbl:adapt_needs_environment}. For each adaptation need, the tables report four elements: a definition of the need, the object of adaptation, the trigger that may activate adaptation, and a representative example from the robotics literature.

Knowledge-driven adaptation needs arise when the robot's internal representation of itself, the environment, or the robot-environment interaction is no longer sufficient to support reliable decision-making and execution. As shown in Table~\ref{tbl:adapt_needs_knowledge}, these needs may originate from abstraction, incompleteness, inaccuracy, or drift. Although these phenomena are closely related, they capture different causes of mismatch between the robot's knowledge and reality. Abstraction is introduced intentionally to keep models manageable, incompleteness reflects missing information, inaccuracy captures incorrect information, and drift occurs when previously valid knowledge becomes misaligned with the current state of the world.

These needs are central to robotic adaptation because decisions depend on the quality and accuracy of the robot's knowledge. For example, planning, localization, manipulation, and task allocation all rely on assumptions about the current state of the system and its environment. When such assumptions become invalid, the robot may need to update its models, revise its state estimates, replan its actions, or compensate for the mismatch in order to continue achieving its intended objectives reliably.

Perception-driven adaptation needs, reported in Table~\ref{tbl:adapt_needs_perception}, concern the robot's ability to acquire and interpret information about itself and the surrounding environment. These needs are different from knowledge-driven needs because they originate in the sensing process rather than in the internal representation alone. A robot may have suitable models and task knowledge, but still fail to operate correctly if the data used to update those models is inconsistent, noisy, incomplete, or unavailable.

The three perception-related needs highlight different failure modes of the sensing pipeline. Multiplicity concerns the reconciliation of heterogeneous or conflicting information from multiple sources. Inaccuracy concerns unreliable or noisy observations. Unavailability concerns the partial or total loss of sensing capabilities. In all cases, adaptation may require changes in sensor fusion, filtering, state estimation, localization, or navigation strategies. 

Actuation-driven adaptation needs, summarized in Table~\ref{tbl:adapt_needs_actuation}, concern the execution side of robotic behavior. Even when a robot has accurate knowledge and reliable perception, commanded actions may not produce the expected physical effects. This can happen because actuators are affected by mechanical imperfections, wear, faults, delays, or interaction effects with the environment.

The classification distinguishes between non-deterministic actuation and actuation unavailability. Non-determinism captures situations in which actuators remain available, but their effects are not fully predictable. Unavailability captures situations in which one or more actuators can no longer be used as expected. These needs typically require adaptation in control, motion planning, redundancy management, or task execution strategies.

\begin{table*}
    \centering
    \caption{System-driven adaptation needs.}
    \label{tbl:adapt_needs_system}
    \small
    \begin{tabular}{
    p{0.3cm}|p{17cm}}
\multirow{4}{*}{\rotatebox[origin=c]{90}{\makebox[4.2cm][c]{Component failure}}}
& 
     \begin{itemize}
        \item {\bf Definition:} Component failure refers to a condition in which one or more components of the robotic system no longer function as expected. Failures may be identified through sensors, external components, stakeholder input, supporting tools, supervision, or observed abnormal system behavior~\cite{verma2006scalable,lu2024event,khalastchi2018fault,avizienis2004basic}.
        \item {\bf Trigger}: Triggered by the detection or diagnosis of abnormal behavior or system faults through monitoring, sensing, or analysis of the robot’s performance and interactions.
        \item {\bf Object(s) of Adaptation:} The robot’s operational configuration, task allocation, coordination mechanisms, or control strategies that must be adjusted to handle the detected fault and maintain system functionality.
        
        \item {\bf Example:}  In a distributed multi-robot system, a robot suddenly starts behaving abnormally due to an internal fault; then the rest of the robots detect the inconsistency in the robot's actions through monitoring, so the system isolates the faulty robot and reconfigures the team~\cite{arrichiello2017distributed}.
    \end{itemize}
     \\
\hline
\multirow{4}{*}{\rotatebox[origin=c]{90}{\makebox[3.8cm][c]{Changing resources}}}
&  
     \begin{itemize}
        \item {\bf Definition:} Changing resources in the system refers to changing, degrading, failing, or fluctuating available resources in the robotic system, such as power constraints, sensors, actuators, software, or hardware~\cite{harris2021online}.
        \item {\bf Trigger}: Triggered by the detection of reduced availability, degradation, failure, or fluctuation of system resources that are necessary for executing tasks or maintaining system performance.
        \item {\bf Object(s) of Adaptation:} The robot’s resource management strategies, task allocation mechanisms, and operational planning that depend on the availability and condition of system resources.
        
        \item {\bf Example:} In multi-robot task allocation for long-duration missions, robots may experience component failures or environmental conditions that make certain capabilities unavailable~\cite{notomista2021resilient}. 
    \end{itemize}
     \\
\hline
\multirow{4}{*}{\rotatebox[origin=c]{90}{\makebox[4.2cm][c]{New resources}}}
& 
     \begin{itemize}
        \item {\bf Definition:} New resources in the system refer to the availability of new resources in the robotic system, such as power constraints, sensors, actuators, software, and hardware~\cite{harris2021online}.
        \item {\bf Trigger}: Triggered by the detection or integration of new resources or components during runtime that were not originally part of the robot’s operational configuration.
        \item {\bf Object(s) of Adaptation:} The robot’s software architecture, perception pipelines, resource management mechanisms, and task planning processes that must incorporate and exploit the newly available resources.
        \item {\bf Example:} When a newly added sensor becomes available at runtime, the robotic software architecture must adapt to integrate the new data stream~\cite{braberman2015morph}. 
    \end{itemize}
     \\
\hline
\multirow{4}{*}{\rotatebox[origin=c]{90}{\makebox[4.2cm][c]{Communication impairment}}}
& 
     \begin{itemize}
        \item {\bf Definition:} Communication impairment in the system refers to communication issues, such as latency, packet loss, bandwidth limitations, interference, intermittent connectivity, synchronization errors, middleware failures, and cybersecurity threats, which constrain/degrade coordination/perception sharing in multi-robot systems or shared robotic systems with different stakeholders, and the system's ability to operate properly~\cite{gielis2022critical}. 
        \item {\bf Trigger}: Triggered by the detection of degraded, delayed, disrupted, or insecure communication affecting the transmission or synchronization of data among system components or collaborating robots.
        \item {\bf Object(s) of Adaptation:} The robot’s communication mechanisms, coordination strategies, data sharing protocols, and decision-making processes that depend on reliable information exchange.
        
        \item {\bf Example:} During drone inspection, high latency and packet loss in the wireless communication link can result in delayed or incomplete video transmission. This degraded communication can potentially compromise safe navigation and real-time decision-making~\cite{ilnytska2020loss}. 
    \end{itemize}
    \\
 
\end{tabular}
\end{table*}

System-driven adaptation needs, reported in Table~\ref{tbl:adapt_needs_system}, concern the internal organization and operational capabilities of the robotic system. These needs are not limited to physical robot components, but also include computational resources, software components, communication mechanisms, and coordination infrastructures. They are particularly relevant for long-running, distributed, or multi-robot systems, where resources and components may change during operation.

The system category includes component failures, changing resources, new resources, and communication impairments. These needs show that adaptation may be required not only when the external environment changes, but also when the robotic system itself changes. A robot may need to isolate a faulty component, reallocate tasks, exploit newly available capabilities, or modify its coordination strategy when communication becomes unreliable.

\begin{table*}
    \centering
    \caption{Mission-driven adaptation needs.}
    \label{tbl:adapt_needs_mission}
    \small
    \begin{tabular}{
    p{0.3cm}|p{17cm}}
 \multirow{4}{*}{\rotatebox[origin=c]{90}{\makebox[4.8cm][c]{Uncertain specification}}}
    & 
    \begin{itemize}
        \item {\bf Definition:} Uncertain mission specification refers to the lack of complete and precise specifications of the mission objectives and the quality of performance. This may arise from abstraction, incompleteness, and inaccuracies in the formulation of mission goals and qualitative criteria. Moreover, relationships may span across single missions/tasks and multiple missions/tasks, including those with the same hierarchical level with different priorities, as well as between global and local objectives~\cite{rostamnia2025towards,weyns2023towards,weyns2019software}.
        \item {\bf Trigger}: Triggered by encountering ambiguous, underspecified, incomplete, or conflicting mission objectives or performance criteria either before or during task execution, which prevents the robot from determining a single precise course of action.  
        \item {\bf Object(s) of Adaptation:} The robot’s mission specification, including its goal, priorities, constraints, together with goal interpretation, planning strategies, and decision-making mechanisms that translate high-level mission objectives and qualitative requirements into concrete operational behaviors.
        
        \item {\bf Example:} A robot receives an order to ``navigate efficiently and safely'' without a precise specification of what efficient or safe means in measurable terms (e.g., time, energy, distance to obstacle). Since these criteria are not explicitly defined, the robot must adapt to interpret and balance these objectives when executing navigation actions~\cite{chen2026learning}.
    \end{itemize}
    \\
\hline
\multirow{4}{*}{\rotatebox[origin=c]{90}{\makebox[4.2cm][c]{Future mission changes}}}
& 
     \begin{itemize}
        \item {\bf Definition:} Future mission changes refer to potential changes (addition, removal, or update) during the system's operation in the tasks of the mission or in the overall mission objectives defined at design time~\cite{bramblett2024robust,ghose2025ve,weyns2019software,zudaire2022assured}.
        \item {\bf Trigger}: Triggered by the detection of modifications to mission definitions, such as the addition, removal, or update of missions or subtasks during system operation.
        \item {\bf Object(s) of Adaptation:} The robot’s mission planning, task decomposition, scheduling, and coordination mechanisms that manage the execution of missions and their subtasks.
        
        \item {\bf Example:} During runtime, a robotic system may receive an updated mission in which a new subtask is created or an existing one is removed. The change can affect other missions executed in the shared environment and cause inconsistency in all missions~\cite{kwan2025onboard,bramblett2024robust}. 
    \end{itemize}\\    
\hline
\multirow{4}{*}{\rotatebox[origin=c]{90}{\makebox[4.2cm][c]{Outdated mission}}}
&  
     \begin{itemize}
        \item {\bf Definition:} Outdated mission refers to the condition in which a mission no longer needs to be executed because its objectives have already been achieved, are impossible to achieve, or are no longer required~\cite{weyns2019software}. 
        \item {\bf Trigger}: Triggered by the detection that a mission objective or performance requirement is no longer applicable, already satisfied, or cannot be meaningfully pursued during system operation.
        \item {\bf Object(s) of Adaptation:} The robot’s mission planning, goal management, and decision-making processes that determine which objectives to pursue and how to allocate effort.    
        \item {\bf Example:} In a multi-robot system, a robot receives a task to inspect a specific area. If the task has already been satisfied by another robot's presence in the environment, the original objective becomes redundant. 
    \end{itemize}
     \\
 \hline
\multirow{4}{*}{\rotatebox[origin=c]{90}{\makebox[4.2cm][c]{Uncertain quality goals}}}
     &  
     \begin{itemize}
        \item {\bf Definition:} Uncertain quality goals refer to failures, changes, conflicts, and additions of non-functional goals in robotic missions. These goals, such as safety, reliability, robustness, efficiency, security, and real-time performance, are sensitive to environmental dynamics, mission interdependencies, and shared resource consumption~\cite{honda2024replan,alberts2025software}.
        \item {\bf Trigger}: Triggered by the detection of changes, conflicts, degradation, or new additions in non-functional goals that affect how missions should be executed.
        \item {\bf Object(s) of Adaptation:} The robot’s decision-making, planning, and control strategies that balance and enforce non-functional requirements during task execution.  
        \item {\bf Example:} A hospital‑service robot may need to adapt when its planned path becomes blocked by a human worker or another robot. It must remain robust in low‑light conditions, compute efficient paths to save time and energy, and ensure safe physical interaction during manipulation tasks~\cite{brugali2019non}.
    \end{itemize}
     \\
 \hline
\multirow{4}{*}{\rotatebox[origin=c]{90}{\makebox[4.2cm][c]{Uncertain relationships}}}
     & 
     \begin{itemize}
        \item {\bf Definition:} Uncertain relationships refer to conflicting, implicit, unknown, or unclear dependencies and interactions among missions at a given time, which are hard to capture in a deterministic way. Missions may share the same physical environment where robots operate. Execution of a mission may cause inconsistencies and make other missions unreachable/uncomplishable, and have an impact on qualitative aspects. Additionally, missions may be executed in parallel or sequentially, which may share, consume, and occupy the same environmental and robotic resources. So, the execution of one mission can affect another mission~\cite{robmosys_separation_levels_concerns,mengi2023mission,wilde2024statistically}. 
        \item {\bf Trigger}: Triggered by the detection of conflicts, dependencies, or unintended interactions between missions, particularly when their execution affects shared resources, environment states, or each other’s feasibility.
        \item {\bf Object(s) of Adaptation:} The robot’s mission coordination, scheduling, and planning mechanisms that manage dependencies, conflicts, and interactions between multiple missions.   
        \item {\bf Example:} In a shared environment, one robot may move an object as part of a mission, while another robot depends on the object being in the original location for a different task. This interaction can degrade the performance of the second mission~\cite{debie2023swarm}. 
    \end{itemize}
     \\

\end{tabular}
\end{table*}

Mission-driven adaptation needs, reported in Table~\ref{tbl:adapt_needs_mission}, arise from uncertainty in what the robotic system is expected to achieve or how success should be evaluated. Unlike knowledge, perception, actuation, and system needs, mission-driven needs concern the objectives and constraints that guide behavior. These needs are especially important in open-ended or collaborative settings, where missions may be underspecified, modified at runtime, or affected by interactions with other missions.

The mission category includes uncertain specifications, future mission changes, outdated missions, uncertain quality goals, and uncertain relationships among missions. These needs highlight that adaptation may require more than replanning a path or compensating for a failed component. In some cases, the robot must reinterpret goals, revise priorities, resolve conflicts among objectives, or determine that a previously assigned mission is no longer relevant.

\begin{table*}
    \centering
    \caption{Environment-driven adaptation needs.}
    \label{tbl:adapt_needs_environment}
    \small
    \begin{tabular}{
    p{0.3cm}|p{17cm}}
\multirow{4}{*}{\rotatebox[origin=c]{90}{\makebox[4.2cm][c]{Execution context changes}}}
&  
 \begin{itemize}
        \item {\bf Definition:} Execution context changes refer to changes in the robot operational environment with which the robotic system interacts. The context changes may arise either in the form of static (e.g., a healthcare robot suddenly transferred to a different domain) or dynamic (e.g., a rover faces a new type of terrain). Due to ongoing changes beyond the control of the developer and robotic system, the context could not be known during system design~\cite{weyns2019software,larsen2024robotic,petrovska2022defining,sun2021uncertain}.
        \item {\bf Trigger}: Triggered by the detection of significant and unforeseen changes in the operating context or system structure that were not anticipated during design time.
        \item {\bf Object(s) of Adaptation:} The robot’s system architecture, behavior, mission, perception, actuation, knowledge of the environment, and interaction strategies that must adjust to operate effectively under new or significantly altered conditions.   
        \item {\bf Example:} A rover on Mars may encounter a completely new type of terrain due to wind effects on Mars. Because the environmental conditions differ from prior assumptions~\cite{rostamnia2025towards}.
    \end{itemize}
 \\
 \hline
 \multirow{4}{*}{\rotatebox[origin=c]{90}{\makebox[3.8cm][c]{Non-human agents coex.}}}
&
 \begin{itemize}
        \item {\bf Definition:}  
        Non-human agents coexistence refers to the presence of other non-human entities, such as robots, autonomous systems, or animals, whose behavior may be unpredictable, difficult to model, and/or capable of changing the environment in ways that affect the robot's mission~\cite{landgraf2021animal}.
        \item {\bf Trigger}: Triggered by the detection of changes in the environment or task conditions caused by the actions of other non-human agents, especially when their behavior deviates from expected or modeled patterns.
        \item {\bf Object(s) of Adaptation:} The robot’s perception, planning, and interaction strategies used to operate safely and effectively in environments shared with other autonomous or semi-autonomous agents.
        
        \item {\bf Example:}  In a shared workstation, another robot may unintentionally change the location of an object that must be picked by a cobot. 
    \end{itemize}
 \\
 \hline
  \multirow{4}{*}{\rotatebox[origin=c]{90}{\makebox[4.2cm][c]{Human agents coex.}}}
&
 \begin{itemize}
        \item {\bf Definition:} 
        Human agents coexistence refers to the presence of humans whose behavior, intentions, preferences, or interactions may be unpredictable, difficult to model, and/or capable of changing the environment, task conditions, or mission requirements in ways that affect the robot's mission~\cite{russell1995modern,zhang2023adaptive,ghadirzadehsensorimotor}. It also introduces safety-critical constraints, as the robot may need to adapt its behavior to avoid harmful interactions, maintain safe distances, and operate in a socially acceptable manner.

        \item {\bf Trigger}: Triggered by the detection of human presence, behavior changes, or interactions that deviate from expected patterns or introduce uncertainty, safety concerns, or ethical constraints.
        \item {\bf Object(s) of Adaptation:} The robot’s perception, planning, interaction, and decision-making strategies, particularly those related to human-aware behavior and ethical considerations.
        
        \item {\bf Example:} A robot operating in the environment in the presence of a human may need to adapt its behavior to secure human privacy. The robot must avoid capturing sensitive visual data or modify its sensing and navigation strategies when people are nearby.
    \end{itemize}
 \\

\end{tabular}
\end{table*}

Environment-driven adaptation needs, reported in Table~\ref{tbl:adapt_needs_environment}, concern changes and interactions that originate from the environment in which the robotic system operates. These needs capture the fact that a robot operates in a physical and social environment that is dynamic and cannot be fully specified at design time. The environment may change because of natural dynamics, human activity, other robots, animals, or unexpected contextual shifts.

This category distinguishes between execution context changes, non-human agents in the loop, and human agents in the loop. Execution context changes concern modifications in the operating conditions themselves. Coexistence of non-human agents introduces uncertainty through the behavior of other robots, animals, or autonomous entities. Coexistence of human agents introduces additional complexity because their behavior is difficult to predict and may involve safety, privacy, ethical, and social constraints. These needs often require adaptation at multiple levels, including perception, planning, interaction, and mission management.

Overall, the classification shows that adaptation needs in robotic systems are multidimensional. They do not arise only from failures or environmental changes, but also from mismatches between knowledge and reality, unreliable perception, non-deterministic actuation, resource variability, mission ambiguity, and interactions with humans or other agents. This observation is important for the remainder of the paper because it clarifies that adaptation cannot be reduced to a single mechanism, such as replanning or fault recovery. Different needs may require different forms of adaptation, different triggers, and different objects of change.

The classification also shows that adaptation needs are often interdependent. For instance, a perception failure may lead to inaccurate knowledge; inaccurate knowledge may cause an inappropriate plan; an inappropriate plan may violate a mission-level quality goal; and an unexpected environmental change may require simultaneous updates to perception, planning, and control. Therefore, robotic adaptation should be understood as a cross-cutting capability that may affect several parts of the system at once.

\subsection{Validation of the Adaptation Needs Classification}

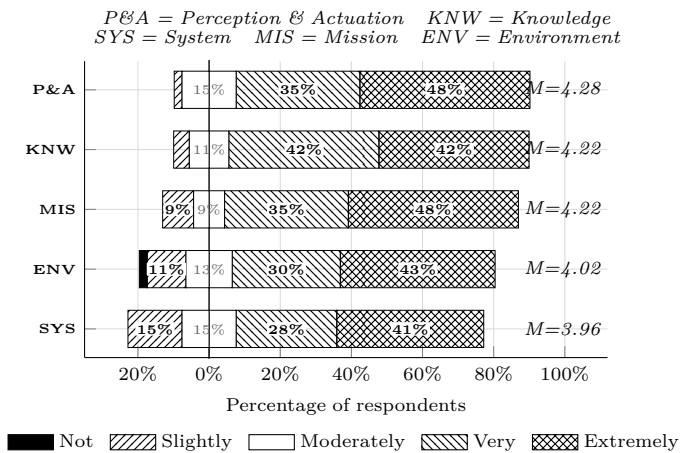
\begin{figure}[htb!]
    \centering
    \begin{tikzpicture}
\small
\begin{axis}[
    xbar stacked, width=8.5cm, height=5.5cm,
    xmin=-35, xmax=112, ytick={0,1,2,3,4},
    yticklabels={SYS,ENV,MIS,KNW,P\&A},
    yticklabel style={font=\tiny\bfseries},
    xlabel={Percentage of respondents}, xlabel style={font=\scriptsize},
    xtick={-20,0,20,40,60,80,100}, xticklabel style={font=\scriptsize},
    xticklabel={\pgfmathparse{abs(\tick)}\pgfmathprintnumber[fixed,precision=0]{\pgfmathresult}\%},
    axis x line*=bottom, axis y line*=none,
    grid=major, major grid style={line width=0.2pt, draw=gray!30},
    enlarge y limits=0.12, bar width=14pt, clip=false,
    legend style={at={(0.5,-0.22)}, anchor=north, draw=none,
        legend columns=5, font=\scriptsize,
        /tikz/every even column/.append style={column sep=4pt}},
    legend image code/.code={\draw[#1] (0cm,-0.06cm) rectangle (0.25cm,0.14cm);},
]
\addplot+[draw=black, line width=0.3pt, fill=white, forget plot]
  coordinates{(-7.6,0)(-6.5,1)(-4.35,2)(-5.55,3)(-7.6,4)};
\addplot+[draw=black, line width=0.3pt, pattern=north east lines, pattern color=black, forget plot]
  coordinates{(-15.2,0)(-10.9,1)(-8.7,2)(-4.4,3)(-2.2,4)};
\addplot+[draw=black, line width=0.3pt, fill=black, forget plot]
  coordinates{(0,0)(-2.2,1)(0,2)(0,3)(0,4)};
\addplot+[draw=black, line width=0.3pt, fill=white, forget plot]
  coordinates{(7.6,0)(6.5,1)(4.35,2)(5.55,3)(7.6,4)};
\addplot+[draw=black, line width=0.3pt, pattern=north west lines, pattern color=black, forget plot]
  coordinates{(28.3,0)(30.4,1)(34.8,2)(42.2,3)(34.8,4)};
\addplot+[draw=black, line width=0.3pt, pattern=crosshatch, pattern color=black, forget plot]
  coordinates{(41.3,0)(43.5,1)(47.8,2)(42.2,3)(47.8,4)};
\addlegendimage{area legend, draw=black, fill=black}
\addlegendentry{Not}
\addlegendimage{area legend, draw=black, pattern=north east lines, pattern color=black}
\addlegendentry{Slightly}
\addlegendimage{area legend, draw=black, fill=white}
\addlegendentry{Moderately}
\addlegendimage{area legend, draw=black, pattern=north west lines, pattern color=black}
\addlegendentry{Very}
\addlegendimage{area legend, draw=black, pattern=crosshatch, pattern color=black}
\addlegendentry{Extremely}
\draw[black,line width=0.5pt](axis cs:0,-0.5)--(axis cs:0,4.5);
\coordinate(s0)at(axis cs:-15.2,0);\coordinate(s1)at(axis cs:-12,1);\coordinate(s2)at(axis cs:-8.7,2);
\coordinate(m0)at(axis cs:0,0);\coordinate(m1)at(axis cs:0,1);\coordinate(m2)at(axis cs:0,2);
\coordinate(m3)at(axis cs:0,3);\coordinate(m4)at(axis cs:0,4);
\coordinate(v0)at(axis cs:21.8,0);\coordinate(v1)at(axis cs:21.7,1);\coordinate(v2)at(axis cs:21.8,2);
\coordinate(v3)at(axis cs:26.7,3);\coordinate(v4)at(axis cs:25,4);
\coordinate(e0)at(axis cs:56.6,0);\coordinate(e1)at(axis cs:58.7,1);\coordinate(e2)at(axis cs:63,2);
\coordinate(e3)at(axis cs:68.9,3);\coordinate(e4)at(axis cs:66.3,4);
\coordinate(mean0)at(axis cs:86,0);\coordinate(mean1)at(axis cs:86,1);
\coordinate(mean2)at(axis cs:86,2);\coordinate(mean3)at(axis cs:86,3);\coordinate(mean4)at(axis cs:86,4);
\end{axis}
\foreach \c/\t in {s0/15\%,s1/11\%,s2/9\%}{
  \node[font=\tiny\bfseries, text=black, fill=white, inner sep=0.5pt, rounded corners=0.5pt] at (\c) {\t};}
\foreach \c/\t in {m0/15\%,m1/13\%,m2/9\%,m3/11\%,m4/15\%}{
  \node[font=\tiny, text=black!60] at (\c) {\t};}
\foreach \c/\t in {v0/28\%,v1/30\%,v2/35\%,v3/42\%,v4/35\%}{
  \node[font=\tiny\bfseries, text=black, fill=white, inner sep=0.5pt, rounded corners=0.5pt] at (\c) {\t};}
\foreach \c/\t in {e0/41\%,e1/43\%,e2/48\%,e3/42\%,e4/48\%}{
  \node[font=\tiny\bfseries, text=black, fill=white, inner sep=0.5pt, rounded corners=0.5pt] at (\c) {\t};}
\foreach \c/\t in {mean0/M=3.96,mean1/M=4.02,mean2/M=4.22,mean3/M=4.22,mean4/M=4.28}{
  \node[font=\scriptsize\itshape, anchor=west] at (\c) {\t};}
\node[anchor=south,font=\scriptsize\itshape]at(3.6,4.25)
  {P\&A = Perception \& Actuation\quad KNW = Knowledge};
\node[anchor=south,font=\scriptsize\itshape]at(3.6,4.00)
  {SYS = System\quad MIS = Mission\quad ENV = Environment};
\end{tikzpicture}
    \caption{Respondent's perception of the relevance of Adaptation Needs in robotics.}
    \label{fig:rq1_validation}
\end{figure}

As discussed in Section~\ref{sub:methodology-validation}, we validated the proposed classification among the authors and through the practitioners' questionnaire and semi-structured interviews. We collected responses from 46 practitioners and ran six interviews. Practitioners were asked to rate the relevance of each category of adaptation needs on a five-point Likert scale (from Not relevant to Extremely relevant). As shown in Figure~\ref{fig:rq1_validation}, all five categories received high relevance ratings (means between 3.96 and 4.28 on the 5-point scale).

A Friedman test across all five categories found no statistically significant difference ($\chi^2\!=\!2.42$, $p\!=\!0.66$, $n\!=\!45$)\footnote{One respondent was excluded due to an \emph{Insufficient expertise} response on knowledge-related needs.}, indicating that practitioners broadly endorsed all categories rather than strongly prioritizing some over others. Only one pairwise Wilcoxon signed-rank comparison reached statistical significance: perception- and actuation-related needs vs.\ environment-related needs ($W\!=\!68$, $p\!=\!0.046$). The higher variance observed for system-related ($SD\!=\!1.09$) and environment-related ($SD\!=\!1.11$) needs reflects practitioner comments that the relevance of these categories depends heavily on the application domain.

Of the 46 respondents, 16 (35\%) provided open-ended comments. Three recurring themes emerged.
First, practitioners emphasized that adaptation needs are \emph{interdependent} rather than isolated: environmental changes frequently trigger perception uncertainty, which subsequently affects knowledge and decision-making. 


 
 
 In this context, they emphasized that robots operating in dynamic and uncertain environments must continuously adapt to changes in the terrain, the lighting, the weather, the moving obstacles, or sensor inaccuracies and incomplete knowledge. One practitioner illustrated the runtime contingencies that must be considered:

 \begin{pquote}
      \textit{``What happens if my sensor is disconnected, what happens if my batteries are low, and what happens if I fail to do the grasping''}. 
 \end{pquote}
 
 Moreover, several respondents, particularly those working with outdoor mobile robots and autonomous vehicles, described these adaptation needs as extremely relevant because they directly affect robot safety and decision-making and require frequent runtime adaptation: 

 \begin{pquote}
    {\it ``If the battery is low, you have to decide if you are able to go back to recharge or go to an emergency mode where you lie down, because collapsing is not acceptable''}.
 \end{pquote}

Second, respondents highlighted \emph{domain-dependence}.
Human-agent and non-human agent coexistence were considered essential for collaborative robots, autonomous vehicles, and multi-robot systems, but less critical in structured industrial environments. Similarly, environmental adaptation was regarded as significantly more important for outdoor robots than for robots operating indoors. Mission-related and system-related needs, while still regarded as important, were perceived as less critical for continuous adaptation. In fact, mission changes typically occur at a higher planning level, whereas system-related issues can often be anticipated through engineering design: 


\begin{pquote}
    {\it``In industry, the majority of what I saw was that the system is not something dynamic. The typical approach is to shut down the robot and restart it if something changes there because you have to deal with some safety concerns''}.
\end{pquote}

This is consistent with the higher variance observed for system-related ($SD\!=\!1.09$) and environment-related ($SD\!=\!1.11$) needs.

Overall, the practitioner feedback demonstrates that the proposed classification is comprehensive and representative of real-world robotic adaptation challenges. The uniformly high ratings across categories, combined with the absence of statistically significant differences, suggest that the classification captures a balanced set of concerns rather than overemphasizing any single dimension of adaptation.

\begin{tcolorbox}[
    enhanced,
    breakable,
    colback=gray!4!white,
    colframe=gray!55!black,
    coltitle=white,
    colbacktitle=black!55!black,
    title=RQ1: Adaptation Needs of Robotic Systems,
    fonttitle=\bfseries,
    boxrule=0.8pt,
    arc=1mm
]
We identified six categories of adaptation needs in robotic systems: \emph{Knowledge}, \emph{Perception}, \emph{Actuation}, \emph{System}, \emph{Mission}, and \emph{Environment}. These needs arise when uncertainty, failures, incomplete information, changing resources, evolving objectives, or dynamic operating conditions make the current robotic behavior inadequate. The classification shows that adaptation is a cross-cutting capability. It may involve updating knowledge, compensating for unreliable perception or actuation, reconfiguring resources, revising mission interpretation, or adapting interactions with humans and other agents. Practitioners highlight the importance of considering {\em interdependence} and {\em domain-dependence} of the adaptation needs. 
\end{tcolorbox}
\section{Features and Limitations of BTs (RQ2)}
\label{sub:framework-rq2}

In this section, we analyze whether classical BTs are sufficient to support the adaptation needs identified in RQ1. We first identify the key features of classical BTs reported in the literature. We subsequently discuss their main limitations with respect to adaptation in robotic systems. We then map the adaptation needs from RQ1 to these features and limitations. Finally, we present validation to prove our findings.

\subsection{Identifying BT Features}

BTs have become popular in robotics because they combine structural clarity with execution-time reactivity. Their tree structure decomposes complex behavior into smaller sub-behaviors, while their ticking semantics allow the robot to repeatedly evaluate conditions and select actions according to the current execution state~\cite{colledanchise2018behavior}. These properties make BTs attractive not only as a modeling notation, but also as an execution mechanism for robotic systems.

A first group of features concerns the organization of behavior. BTs are inherently \emph{modular}: each subtree can be treated as a behavioral component that can be developed, tested, replaced, and reused independently~\cite{colledanchise2018behavior}. This modularity supports fault isolation, customization, and incremental upgrades, which are important properties in robotic software architectures~\cite{wolf2023modularity}. Closely related to modularity are \emph{reusability}, \emph{maintainability}, and \emph{extensibility}. Since behaviors are represented as composable subtrees, existing behaviors can be reused in different parts of a BT, modified locally, or extended with new nodes without redesigning the entire control structure~\cite{colledanchise2018behavior,ghzouli2023behavior}.

A second group of features concerns human understanding and structural abstraction. BTs expose the structure of decision-making through a hierarchical tree, which makes complex behaviors easier for developers and operators to inspect, communicate, and debug. This supports \emph{readability}, i.e., the ability of humans to understand the behavioral structure, and \emph{interpretability}, i.e., the ability to reason about why specific actions or tasks are selected during execution~\cite{colledanchise2018behavior,tziafas2022enhancing}. Since readability mainly benefits human comprehension and maintenance rather than runtime adaptation, we do not include it in the adaptation mapping. By contrast, \emph{hierarchical decomposability}, listed as\emph{decomposability} in the table~\ref{tbl:map_adaptation_needs_to_BT_features_limitations}, can support adaptation more directly, since it allows complex missions to be represented and modified at multiple levels of abstraction, from high-level goals to low-level actions~\cite{darvish2020hierarchical}.

A third group of features concerns execution. BTs are \emph{reactive} because their execution is driven by repeated ticking and condition evaluation, allowing the selected behavior to change according to the current state of the system and environment~\cite{colledanchise2018behavior}. They are also \emph{flexible} in the sense that their modular and hierarchical organization allows developers to adapt the same behavioral structure to different tasks or environments with limited restructuring~\cite{ghzouli2023behavior}. 

Overall, these features explain why BTs are often used as the behavioral backbone of robotic systems. At the same time, they do not imply that classical BTs are sufficient for all forms of adaptation. Therefore, the next subsection analyses the limitations that arise when BTs are used in dynamic, uncertain, and human-centred robotic environments.

\subsection{Identifying BT Limitations}

Although classical BTs provide a scalable and modular structure for robot control, their use in adaptive and dynamic robotic systems presents several limitations. These limitations affect runtime flexibility, uncertainty handling, task resumption, quality-awareness, and integration with planning and learning frameworks. Table~\ref{tbl:bt_reasons} summarizes the main limitations identified in the analyzed literature.

\begin{table}[t]
\centering
\caption{Limitations of adaptability and extensibility of BTs.}
\label{tbl:bt_reasons}
\small
\begin{tabular}{p{0.9cm}|p{6.9cm}}
\hline
\textbf{Limit.} & \textbf{Description} \\
\hline
L1 & Static control flow in classical BTs limits runtime flexibility~\cite{pezzato2023active,rovida2017extended}. \\
\hline
L2 & Fixed execution order prevents flexible adaptation to runtime changes~\cite{rovida2017extended,fusaro2021human}. \\
\hline
L3 & Classical BTs often operate under deterministic assumptions, which are unsuitable for uncertain environments~\cite{fusaro2021human,rostamnia2025towards,abiyev2016robot,behery2023human}. \\
\hline
L4 & Difficulty handling partial observability and incomplete state information~\cite{li2022towards,pezzato2023active}. \\
\hline
L5 & Human-centered and collaborative industrial environments introduce uncertainty and variability not captured by classical BTs~\cite{behery2023human,fusaro2021human}. \\
\hline
L6 & Suboptimal execution and lack of explicit representation of the purpose and intent of behaviors~\cite{rovida2017extended}. \\
\hline
L7 & Modeling conditions as explicit nodes is unsuitable for some planning-based domains~\cite{rovida2017extended}. \\
\hline
L8 & Challenges in representing and managing highly complex behaviors~\cite{pezzato2023active}. \\
\hline
L9 & Limited explainability, particularly in diagnosing and explaining execution failures~\cite{lemasurier2024reactive}. \\
\hline
L10 & Lack of mechanisms for task resumption after interruptions or failures~\cite{el2021resume}. \\
\hline
L11 & Architectural adaptation of BTs often focuses on functional goals while neglecting quality requirements~\cite{alberts2024rebet,wang2020extending}. \\
\hline
L12 & Difficulty handling continuous-valued states and actions~\cite{abiyev2016robot}. \\
\hline
\end{tabular}
\end{table}

\color{black}

\newcommand{\rot}[1]{\rotatebox{90}{\parbox{2cm}{\centering\scriptsize #1}}}
\setlength\dashlinedash{0.6pt}
\setlength\dashlinegap{1.2pt}
\setlength\arrayrulewidth{0.4pt}

\begin{table}[htbp]
\centering
\renewcommand{\arraystretch}{1.4}
\caption{Mapping adaptation needs to BT features and limitations. \textbf{Legend:}
$\cmark$~sufficient; $\partialmark$~partial/implicit; $\xmark$~no support; $\square$~no information.}
\label{tbl:map_adaptation_needs_to_BT_features_limitations}
\resizebox{\columnwidth}{!}{%
\begin{tabular}{p{.12cm}|p{2.1cm}|p{.12cm}:p{.12cm}:p{.12cm}:p{.12cm}:p{.12cm}:p{.12cm}:p{.01cm}:p{.12cm}:p{1.5cm}}
 & \textbf{Subtype}
 & \rotatebox{90}{Modularity} 
 & \rotatebox{90}{Interpretability} 
 & \rotatebox{90}{Reactivity} 
 & \rotatebox{90}{Reusability} 
 & \rotatebox{90}{Maintainability} 
 & \rotatebox{90}{Extensibility} 
 & \rotatebox{90}{
 Decomposability}
 & \rotatebox{90}{Flexibility}
 & \textbf{BT limit.} \\ \hline

\multirow{4}{*}{\rotatebox[origin=c]{90}{\makebox[0.2cm][c]{Knowledge}}}
& Abstraction     & \cellcolor{gray!20}&   \cellcolor{gray!20}& \cellcolor{gray!20} &\cellcolor{gray!20} & \cellcolor{gray!20} & \cellcolor{gray!20} & \cellcolor{gray!20} 
&  \cellcolor{gray!20}  & \cellcolor{gray!20}-\\ \cline{2-11}
& Incompleteness  &  $\cmark$ &  & $\partialmark$ & $\cmark$ & $\cmark$ &  $\cmark$ & $\cmark$ &  &  L4\\ \cline{2-11}
& Inaccuracy      &  &  & $\cmark$  &  &  &  &  &  &  -\\ \cline{2-11}
& Drift      & $\cmark$ & $\cmark$ & $\cmark$ & $\cmark$ & $\cmark$ &  &  & $\cmark$  & -\\ \hline
\multirow{3}{*}{\rotatebox[origin=c]{90}{\makebox[0.1cm][c]{Percep.}}}
 & Multiplicity & \cellcolor{gray!20} &   \cellcolor{gray!20}& \cellcolor{gray!20} & \cellcolor{gray!20} & \cellcolor{gray!20} & \cellcolor{gray!20} &\cellcolor{gray!20}  & \cellcolor{gray!20} &  \cellcolor{gray!20}-\\ \cline{2-11}
 & Inaccuracy & \cellcolor{gray!20} & \cellcolor{gray!20}  & \cellcolor{gray!20} & \cellcolor{gray!20} & \cellcolor{gray!20} & \cellcolor{gray!20} & \cellcolor{gray!20} &  \cellcolor{gray!20} & \cellcolor{gray!20}-\\ \cline{2-11}
 & Unavailability & \cellcolor{gray!20}  & \cellcolor{gray!20} & \cellcolor{gray!20} & \cellcolor{gray!20} & \cellcolor{gray!20} & \cellcolor{gray!20} & \cellcolor{gray!20} & \cellcolor{gray!20} & \cellcolor{gray!20}-\\ \hline
 \multirow{2}{*}{\rotatebox[origin=c]{90}{\makebox[1.2cm][c]{Act.}}}
 & Non-determinism &  &  & $\cmark$  &  &  &  &  &   & -\\ \cline{2-11}
 & Unavailability  & \cellcolor{gray!20} & \cellcolor{gray!20} & \cellcolor{gray!20} &  \cellcolor{gray!20} & \cellcolor{gray!20} & \cellcolor{gray!20} & \cellcolor{gray!20} & \cellcolor{gray!20} &  \cellcolor{gray!20}-\\ \hline
 \multirow{4}{*}{\rotatebox[origin=c]{90}{\makebox[2.5cm][c]{System}}}
 & Component failure  & $\cmark$ &  & $\partialmark$  &  &  &  &  & $\xmark$ &  L10, L11\\ \cline{2-11}
 & Changing resources & $\cmark$ &  &  &  & $\cmark$ &  & $\cmark$ &  &  \\ \cline{2-11}
 & New resources  & \cellcolor{gray!20} & \cellcolor{gray!20} & \cellcolor{gray!20} & \cellcolor{gray!20} & \cellcolor{gray!20} & \cellcolor{gray!20} & \cellcolor{gray!20} & \cellcolor{gray!20}  & \cellcolor{gray!20}-\\ \cline{2-11}
 & Communication impairment & \cellcolor{gray!20} & \cellcolor{gray!20} & \cellcolor{gray!20} & \cellcolor{gray!20} & \cellcolor{gray!20} & \cellcolor{gray!20} & \cellcolor{gray!20} & \cellcolor{gray!20} & \cellcolor{gray!20}-\\ \hline 
\multirow{5}{*}{\rotatebox[origin=c]{90}{\makebox[3.7cm][c]{Mission}}}
 & Uncertain specification   & $\partialmark$ &  & $\partialmark$ & $\cmark$ &  & $\cmark$ & $\cmark$ & $\partialmark$  & L1, L2, L3, L5, L6, L7\\ \cline{2-11}
 & Future mission changes  & \cellcolor{gray!20} &  \cellcolor{gray!20} & \cellcolor{gray!20} & \cellcolor{gray!20} & \cellcolor{gray!20} & \cellcolor{gray!20} & \cellcolor{gray!20} & \cellcolor{gray!20}  & \cellcolor{gray!20}-\\ \cline{2-11}
 & Outdated mission & \cellcolor{gray!20} & \cellcolor{gray!20} & \cellcolor{gray!20} & \cellcolor{gray!20} & \cellcolor{gray!20} & \cellcolor{gray!20} & \cellcolor{gray!20} &  \cellcolor{gray!20} & \cellcolor{gray!20}-\\ \cline{2-11}
 & Uncertain quality goals & &  &  &  &  &  &  & $\xmark$  & L11 \\ \cline{2-11}
 & Uncertain relationships  & \cellcolor{gray!20} & \cellcolor{gray!20} & \cellcolor{gray!20} & \cellcolor{gray!20} & \cellcolor{gray!20} & \cellcolor{gray!20} & \cellcolor{gray!20} & \cellcolor{gray!20} & \cellcolor{gray!20}- \\ \cline{2-11}
\hline
\multirow{3}{*}{\rotatebox[origin=c]{90}{\makebox[2.5cm][c]{Environment}}}
 & Execution context changes  & $\cmark$ &  & $\partialmark$  & $\cmark$ & $\cmark$ & $\cmark$ & $\cmark$  &$\partialmark$ & L1, L2, L3, L4, L8, L11\\ \cline{2-11}
 & Non-human agents coexistence  & $\cmark$ &  & $\xmark$  &  &  &  &  &  &  L3, L8, L12\\ \cline{2-11}
 & Human agents coexistence  & $\cmark$ &  & $\partialmark$  & $\cmark$ & $\cmark$ & $\cmark$ & & $\partialmark$  & L1, L4, L5, L10\\ \cline{2-11}
\hline
\end{tabular}}
\end{table}

Classical BTs suffer from limited runtime flexibility due to their static control flow (L1). Execution usually follows a fixed order (L2), which makes it difficult for the system to adapt when task priorities, action costs, or environmental conditions change at runtime. As a result, introducing new skills or modifying behavior often requires restructuring the tree.

Another major limitation concerns uncertainty and partial observability. Classical BTs are often designed under deterministic assumptions (L3), which makes them less suitable for real-world robotic scenarios where sensor information may be incomplete, noisy, or unavailable. They also struggle with incomplete state information and partial observability (L4), often requiring external replanning or reasoning mechanisms to maintain correct behavior.

Classical BTs are also limited in human-centered and collaborative settings (L5). They lack explicit mechanisms for representing human intentions, availability, preferences, and variability, which are critical in human-robot interaction. This limits their ability to adapt actions according to human behavior or changing collaboration conditions. Moreover, classical BTs do not explicitly represent the purpose or intent of behaviors (L6), and their treatment of conditions as explicit nodes can create mismatches with planning-based domains where conditions are embedded in actions (L7).

As systems grow more complex, BTs may become large and difficult to manage (L8), especially when reactivity is achieved through hard-coded recovery branches. Classical BTs also provide limited support for explaining failures (L9): they may indicate where execution failed, but not necessarily why the failure occurred. Furthermore, they provide weak support for interruptions and task resumption (L10), since execution may need to restart or rely on ad hoc state-retention mechanisms. Finally, classical BTs usually focus on functional goals and provide limited support for quality requirements such as energy consumption, safety, robustness, or efficiency (L11), and they are less suitable for continuous-valued states and actions that require nuanced, non-crisp decisions (L12).

\subsection{Mapping Adaptation Needs to BT Features and Limitations}

Table~\ref{tbl:map_adaptation_needs_to_BT_features_limitations} maps the adaptation needs identified in RQ1 to the features and limitations of classical BTs. The table indicates whether the classical features of BT are sufficient ($\cmark$), partially sufficient ($\partialmark$), unsupported ($\xmark$), or whether explicit evidence was not found in the studies analyzed ($\square$). The last column links unsupported or partially supported needs to the limitations summarized in Table~\ref{tbl:bt_reasons}. Partially sufficient indicates that, while several studies have reported the feature as sufficient to address the needs, at least one study has reported a limitation regarding that feature. Unsupported indicates that at least one study has reported a limitation associated with the feature, with no study reporting evidence to the contrary. Furthermore, the absence of evidence in some cells does not necessarily mean that classical BTs cannot address the corresponding need; rather, it indicates that the analyzed literature does not provide explicit support for that feature--need relation.

\subsection{Validation of BT Support for the Families of Adaptation Needs}

\begin{figure}[t]
    \centering
    \begin{tikzpicture}
\small
\begin{axis}[
    xbar stacked, width=8.5cm, height=5.5cm,
    xmin=0, xmax=100, ytick={0,1,2,3,4},
    yticklabels={ENV,MIS,SYS,KNW,P\&A},
    yticklabel style={font=\tiny\bfseries},
    xlabel={Percentage of respondents}, xlabel style={font=\scriptsize},
    xtick={0,20,40,60,80,100},
    xticklabel={\pgfmathprintnumber[fixed,precision=0]{\tick}\%},
    xticklabel style={font=\scriptsize},
    axis x line*=bottom, axis y line*=none,
    grid=major, major grid style={line width=0.2pt, draw=gray!30},
    enlarge y limits=0.12, bar width=14pt, clip=false,
    legend style={at={(0.5,-0.20)}, anchor=north, draw=none, legend columns=6, font=\scriptsize,
        /tikz/every even column/.append style={column sep=2pt}},
    legend image code/.code={\draw[#1] (0cm,-0.06cm) rectangle (0.22cm,0.14cm);},
]
\addplot+[draw=black, line width=0.3pt, fill=black]
  coordinates {(8.7,0)(13.0,1)(10.9,2)(15.2,3)(4.3,4)};
\addplot+[draw=black, line width=0.3pt, pattern=north east lines, pattern color=black]
  coordinates {(10.9,0)(28.3,1)(19.6,2)(17.4,3)(8.7,4)};
\addplot+[draw=black, line width=0.3pt, pattern=dots, pattern color=black]
  coordinates {(17.4,0)(19.6,1)(15.2,2)(13.0,3)(6.5,4)};
\addplot+[draw=black, line width=0.3pt, pattern=north west lines, pattern color=black]
  coordinates {(15.2,0)(13.0,1)(19.6,2)(21.7,3)(39.1,4)};
\addplot+[draw=black, line width=0.3pt, pattern=crosshatch, pattern color=black]
  coordinates {(39.1,0)(21.7,1)(26.1,2)(26.1,3)(32.6,4)};
\addplot+[draw=black, line width=0.3pt, fill=white]
  coordinates {(8.7,0)(4.3,1)(8.7,2)(6.5,3)(8.7,4)};
\legend{Classical, Update, Extend, External, BT+Ext., None}
\draw[black!60,line width=0.4pt](axis cs:4.3,3.72)--(axis cs:4.3,4.28);
\draw[black!60,line width=0.4pt](axis cs:19.6,3.72)--(axis cs:19.6,4.28);
\draw[black!60,line width=0.4pt](axis cs:58.7,3.72)--(axis cs:58.7,4.28);
\draw[black!60,line width=0.4pt](axis cs:91.3,3.72)--(axis cs:91.3,4.28);
\draw[black!60,line width=0.4pt](axis cs:15.2,2.72)--(axis cs:15.2,3.28);
\draw[black!60,line width=0.4pt](axis cs:45.7,2.72)--(axis cs:45.7,3.28);
\draw[black!60,line width=0.4pt](axis cs:67.4,2.72)--(axis cs:67.4,3.28);
\draw[black!60,line width=0.4pt](axis cs:93.5,2.72)--(axis cs:93.5,3.28);
\draw[black!60,line width=0.4pt](axis cs:10.9,1.72)--(axis cs:10.9,2.28);
\draw[black!60,line width=0.4pt](axis cs:45.6,1.72)--(axis cs:45.6,2.28);
\draw[black!60,line width=0.4pt](axis cs:65.2,1.72)--(axis cs:65.2,2.28);
\draw[black!60,line width=0.4pt](axis cs:91.3,1.72)--(axis cs:91.3,2.28);
\draw[black!60,line width=0.4pt](axis cs:13.0,0.72)--(axis cs:13.0,1.28);
\draw[black!60,line width=0.4pt](axis cs:61.0,0.72)--(axis cs:61.0,1.28);
\draw[black!60,line width=0.4pt](axis cs:74.0,0.72)--(axis cs:74.0,1.28);
\draw[black!60,line width=0.4pt](axis cs:95.7,0.72)--(axis cs:95.7,1.28);
\draw[black!60,line width=0.4pt](axis cs:8.7,-0.28)--(axis cs:8.7,0.28);
\draw[black!60,line width=0.4pt](axis cs:37.0,-0.28)--(axis cs:37.0,0.28);
\draw[black!60,line width=0.4pt](axis cs:52.2,-0.28)--(axis cs:52.2,0.28);
\draw[black!60,line width=0.4pt](axis cs:91.3,-0.28)--(axis cs:91.3,0.28);
\coordinate(p4a)at(axis cs:8.65,4);\coordinate(p4b)at(axis cs:16.25,4);
\coordinate(p4c)at(axis cs:39.1,4);\coordinate(p4d)at(axis cs:75.0,4);\coordinate(p4e)at(axis cs:95.65,4);
\coordinate(k3a)at(axis cs:7.6,3);\coordinate(k3b)at(axis cs:23.9,3);\coordinate(k3c)at(axis cs:39.1,3);
\coordinate(k3d)at(axis cs:56.5,3);\coordinate(k3e)at(axis cs:80.45,3);\coordinate(k3f)at(axis cs:96.75,3);
\coordinate(s2a)at(axis cs:5.45,2);\coordinate(s2b)at(axis cs:20.7,2);\coordinate(s2c)at(axis cs:38.1,2);
\coordinate(s2d)at(axis cs:55.4,2);\coordinate(s2e)at(axis cs:78.3,2);\coordinate(s2f)at(axis cs:95.65,2);
\coordinate(m1a)at(axis cs:6.5,1);\coordinate(m1b)at(axis cs:27.15,1);\coordinate(m1c)at(axis cs:51.1,1);
\coordinate(m1d)at(axis cs:67.4,1);\coordinate(m1e)at(axis cs:84.8,1);
\coordinate(e0a)at(axis cs:4.35,0);\coordinate(e0b)at(axis cs:14.15,0);\coordinate(e0c)at(axis cs:28.3,0);
\coordinate(e0d)at(axis cs:44.6,0);\coordinate(e0e)at(axis cs:71.75,0);\coordinate(e0f)at(axis cs:95.65,0);
\end{axis}
\foreach \c/\t in {p4a/9\%,p4b/7\%,p4c/39\%,p4d/33\%,p4e/9\%,
  k3a/15\%,k3b/17\%,k3c/13\%,k3d/22\%,k3e/26\%,k3f/7\%,
  s2a/11\%,s2b/20\%,s2c/15\%,s2d/20\%,s2e/26\%,s2f/9\%,
  m1a/13\%,m1b/28\%,m1c/20\%,m1d/13\%,m1e/22\%,
  e0a/9\%,e0b/11\%,e0c/17\%,e0d/15\%,e0e/39\%,e0f/9\%}{
  \node[font=\tiny\bfseries, text=black, fill=white, inner sep=0.5pt, rounded corners=0.5pt] at (\c) {\t};
}
\node[anchor=south,font=\scriptsize\itshape] at (3.6,4.25)
  {P\&A = Perception \& Actuation\quad KNW = Knowledge};
\node[anchor=south,font=\scriptsize\itshape] at (3.6,4.00)
  {SYS = System\quad MIS = Mission\quad ENV = Environment};
\end{tikzpicture}
    \caption{Practitioner responses on how each family of adaptation needs should be addressed using BTs.}
    \label{fig:rq2_validation}
\end{figure}
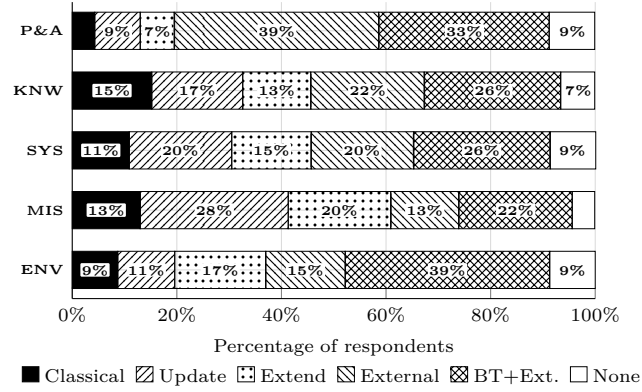

As shown in Figure~\ref{fig:rq2_validation}, classical BTs alone were considered insufficient to address adaptation needs across all five categories: only 4--15\% of respondents considered classical BTs sufficient, as BTs operate primarily at the execution layer and lack native mechanisms for representing or exploiting uncertainty. Nevertheless, respondents indicated that BTs can address adaptation needs to varying degrees, with the role of the BT differing significantly across families ($\chi^2\!=\!24.66$, $p\!=\!0.076$, $\text{dof}\!=\!16$). Practitioners consistently emphasized that BTs excel at orchestrating adaptive behaviors once the required information is available, while external components remain responsible for generating, updating, or interpreting that information.

Knowledge-related adaptation needs were generally considered addressable through BTs, particularly when the required knowledge can be obtained or updated by external modules. Respondents indicated that BTs effectively react to changes such as knowledge drift or inaccuracies once these changes become observable, whereas knowledge acquisition, abstraction, and reasoning remain outside the scope of the execution layer. Accordingly, 48.8\% of respondents favored a BT-centric approach (classical, update, or extend BT), while 23.3\% favored external modules and 27.9\% favored a combined strategy. Knowledge-related adaptation is thus achieved through the integration of BTs with external reasoning components rather than through the BT formalism alone.

Perception- and actuation-related adaptation needs received the strongest consensus on the limitations of BTs. This family had the highest reliance on external modules: 78.6\% of respondents chose ``external modules'' or ``both BT + external,'' significantly more than the 49.1\% average across other families (Fisher exact test: $OR\!=\!3.80$, $p\!<\!0.001$). Participants agreed that perception uncertainty, sensor failures, and actuator adaptation must primarily be handled by dedicated perception algorithms, controllers, and hardware-specific components. BTs remain valuable for coordinating recovery strategies and selecting appropriate behaviors based on the outputs of these modules, but they do not improve perception or actuation capabilities themselves. 
One interviewee explained:
\begin{pquote}
    {\it ``The execution layer needs to be able to observe an abnormal state and then take actions, but it is not responsible for addressing it or compensating for it.''}
\end{pquote}

System-related adaptation needs were viewed as more challenging because they often require coordination across multiple software layers. Although respondents agreed that BTs can effectively react to observable failures through recovery behaviors, they also highlighted that resource monitoring, middleware adaptation, and architectural changes require support from external runtime mechanisms. 

\begin{observation}{Industrial Vision}
    In industrial robotic systems, dynamic reconfiguration is uncommon because safety certification discourages runtime modification of system components. Instead, robots are often stopped and restarted whenever significant configuration changes occur. Industrial robotics is dominated by safety requirements, reliability constraints, certification, and predictable behavior. Consequently, BTs mainly coordinate predefined recovery procedures rather than dynamically restructuring the robotic system.
\end{observation}

Mission-related adaptation needs generated more nuanced opinions. This family was the most BT-centric: 63.6\% of the respondents favored a BT-internal approach, significantly more than the average of 40.2\% across other families (Fisher's exact test: $OR\!=\!2.60$, $p\!=\!0.007$). Interviewees emphasized that uncertain, incomplete, or ambiguous mission specifications cannot be resolved at the execution layer because they represent inputs to the BT rather than runtime events. In practice, mission clarification should be handled by planning or reasoning components before execution begins, after which BTs effectively organize and execute the resulting tasks. This distinction was expressed directly by one interviewee: 
\begin{pquote}
    {\it ``If the mission is incomplete, that is an input, not an output. [...] It is a precondition that needs to be met to have a correct definition of the mission.''}
\end{pquote}

Finally, environment-related adaptation needs emerged as one of the strongest application areas for BTs. This family had the highest rate of combined approaches (39.1\% selected ``both BT formalism + external modules''), reflecting that environmental changes are observable runtime events that naturally trigger BT reactivity, while complex adaptations still benefit from external perception and reasoning modules. Both the survey and interview responses consistently identified adaptation of the environment as a major strength of BTs. One interviewee explained,
\begin{pquote}
{\it The whole point of BTs is to have reactive behaviors. It works really well when you have a human or an external agent that is disrupting your plan, because you can react quickly with a BT, but you can react [only] if you have a branch for it.''}
\end{pquote}

\begin{tcolorbox}[
    enhanced,
    breakable,
    colback=gray!4!white,
    colframe=gray!55!black,
    coltitle=white,
    colbacktitle=black!55!black,
    title=RQ2: Features and Limitations of BTs,
    fonttitle=\bfseries,
    boxrule=0.8pt,
    arc=1mm
]
Classical BTs support robotic adaptation through modularity, hierarchy, reactivity, reusability, maintainability, extensibility, and readability. These features make them suitable for structuring known tasks, composing reusable behaviors, reacting to monitored conditions, and encoding predefined recovery strategies.

However, classical BTs are only partially sufficient for modern robotic adaptation. Their limitations include static execution structures, deterministic assumptions, weak support for partial observability, limited representation of intent, quality goals, human behavior, and continuous-valued decisions, as well as limited interruption handling and task resumption. Thus, BTs often need extensions or integration with planning, learning, monitoring, or reasoning mechanisms to address dynamic missions, uncertain environments, resource changes, and human-centered scenarios.
\end{tcolorbox}

\section{Adaptive BTs: Approaches and Enhancements (RQ3)}
\label{sub:framework-rq3}

This section answers RQ3 by characterizing the main approaches used to enhance BTs for adaptation. The goal is not only to list existing techniques, but also to understand how they compensate for the limitations of classical BTs identified in RQ2. We organize the analyzed studies into four primary families of BT enhancement: \textit{BT evolution}, \textit{BT extension}, \textit{BT generation}, and \textit{BT refinement}. Some approaches combine these families, particularly BT generation and refinement, and we term them \textit{hybrid} approaches.  
Table~\ref{tbl:map_adaptation_needs_to_approaches_suitability} presents the suitability of each family of adaptive BT approaches for addressing the identified adaptation needs. The table indicates whether different categories of BTs to address adaptation needs are suitable ($\cmark$), partially suitable ($\partialmark$), not suitable ($\xmark$), or whether no explicit evidence was found in the studies analyzed ($\square$). In our analysis, a BT-based approach is: \emph{Suitable} if it is sufficient as the primary mechanism to address the adaptation need; \emph{Partially suitable} if it plays an important role, but must be combined with external modules (e.g., perception, planning, controllers, knowledge bases) or only addresses part of the adaptation. A \emph{Not suitable} approach cannot contribute (at all) to tackling an adaptation need; it is orthogonal to the problem.
The number of papers that tackled specific adaptation needs alone is not the criterion. Furthermore, during analysis, some studies fell into more than one family. In particular, some approaches both generate a BT from external information, such as demonstrations, symbolic models, natural-language instructions, or learned policies, and refine the generated or existing BT through optimization, interaction, or runtime correction. We discuss these cases as \textit{hybrid} approaches in the text, while representing them in the table through entries in both the \textit{Generation} and \textit{Refinement} columns.

\newcommand{\tcite}[1]{\scriptsize\cite{#1}}

\newcommand{\evolgray}{%
  \cellcolor{gray!20}[\tabcolsep][2\tabcolsep]\strut
}

 \begin{table}[htbp]
\caption{Mapping adaptation needs to BT adaptation and evolution approaches. \textbf{Legend:}
$\cmark$~suitable; $\partialmark$~partially suitable; $\xmark$~not suitable; $\square$~no information.}
\label{tbl:map_adaptation_needs_to_approaches_suitability}
\resizebox{\columnwidth}{!}{%

\begin{tabular}{p{0.1cm}|p{1.8cm}|p{1.35cm}|p{1.5cm}|p{1.5cm}|p{0.7cm}}
\hline
& \textbf{Subtype} & Evol. & Exten. & Gen. & Refin.  \\
\hline
\multirow{4}{*}{\rotatebox[origin=c]{90}{\makebox[0.5cm][c]{Knowled.}}}
 & Abstraction & \cellcolor{gray!20} & \cellcolor{gray!20} &\cellcolor{gray!20} & \cellcolor{gray!20}\\
 \cline{2-6}
 & Incompleteness & \multicolumn{1}{@{}l|@{}}{$\cmark$(\tcite{hallawa2017instinct})} & $\cmark$(\tcite{li2022towards}) & $\partialmark$(\tcite{iovino2023framework,scherf2023interactively}) & \multicolumn{1}{@{}l@{}}{$\cmark$(\tcite{scherf2023interactively})} \\
 \cline{2-6}
 & Inaccuracy&  & & $\partialmark$(\tcite{scherf2023interactively}) & \multicolumn{1}{@{}l@{}}{$\cmark$(\tcite{scherf2023interactively})}\\
 \cline{2-6}
 & Drift& \multicolumn{1}{@{}l|@{}}{$\cmark$(\tcite{scheper2016behavior})} & & $\partialmark$(\tcite{ruiz2022automating}) & \\
 \cline{2-6}
\hline
\multirow{3}{*}{\rotatebox[origin=c]{90}{\makebox[0.2cm][c]{Percep.}}}
 & Multiplicity & \cellcolor{gray!20}& \cellcolor{gray!20} & \cellcolor{gray!20} & \cellcolor{gray!20}  \\
 \cline{2-6}
 & Inaccuracy &\cellcolor{gray!20} & \cellcolor{gray!20} & \cellcolor{gray!20} & \cellcolor{gray!20}  \\
 \cline{2-6}
 & Unavailability & \cellcolor{gray!20} & \cellcolor{gray!20} &\cellcolor{gray!20} & \cellcolor{gray!20} \\
  \hline
\multirow{2}{*}{\rotatebox[origin=c]{90}{\makebox[0.8cm][c]{Act.}}}
 & Non-determinism & &  & $\partialmark$(\tcite{scherf2023interactively}) & \multicolumn{1}{@{}l@{}}{$\cmark$(\tcite{scherf2023interactively})}  \\
 \cline{2-6}
 & Unavailability & \cellcolor{gray!20} & \cellcolor{gray!20} & \cellcolor{gray!20}&  \cellcolor{gray!20}\\
  \hline
\multirow{4}{*}{\rotatebox[origin=c]{90}{\makebox[2.2cm][c]{System}}}
 & Component failure &  &$\cmark$(\tcite{el2021resume,wang2020extending})&  $\partialmark$(\tcite{segura2017integration}) &  \\
 \cline{2-6}
 & Changing resources & \multicolumn{1}{@{}l|@{}}{$\cmark$(\cite{chen2023co})} &  & &  \\
 \cline{2-6}
 & New resources & \cellcolor{gray!20}&  \cellcolor{gray!20}& \cellcolor{gray!20} &  \cellcolor{gray!20} \\
 \cline{2-6}
 & Communication impairment &\cellcolor{gray!20} & \cellcolor{gray!20} & \cellcolor{gray!20} &  \cellcolor{gray!20} \\
  \hline 
\multirow{5}{*}{\rotatebox[origin=c]{90}{\makebox[3.0cm][c]{Mission}}}
 & Uncertain specification & \multicolumn{1}{@{}l|@{}}{$\cmark$(\tcite{montague2024hierarchical,hallawa2020evolving})} & $\partialmark$(\tcite{fusaro2021human,rovida2017extended,behery2023human,rostamnia2025towards}) & $\cmark$(\tcite{iovino2023framework,colledanchise2018learning,deng2023learning}) & \multicolumn{1}{@{}l@{}}{$\cmark$(\tcite{deng2023learning})} \\
 \cline{2-6}
& Future mission changes & \cellcolor{gray!20}& \cellcolor{gray!20} & \cellcolor{gray!20} &  \cellcolor{gray!20} \\
 \cline{2-6}
 & Outdated mission &\cellcolor{gray!20} & \cellcolor{gray!20} & \cellcolor{gray!20} & \cellcolor{gray!20}  \\
 \cline{2-6}
 & Uncertain quality goals & \cellcolor{gray!20} &  \cellcolor{gray!20} &\cellcolor{gray!20} & \cellcolor{gray!20}\\
 \cline{2-6}
 & Uncertain relationships & \cellcolor{gray!20} & \cellcolor{gray!20} & \cellcolor{gray!20} & \cellcolor{gray!20}\\
\hline
\multirow{3}{*}{\rotatebox[origin=c]{90}{\makebox[2.2cm][c]{Environment}}}
 & Execution context changes& \multicolumn{1}{@{}l|@{}}{$\cmark$(\tcite{hallawa2017instinct})} & $\partialmark$ (\tcite{pezzato2023active,fusaro2021human,li2022towards,alberts2024rebet,wang2020extending,rostamnia2025towards}) & $\partialmark$(\tcite{wen2024auction,verma2021automatic,zhou2024llm,herranz2022decentralised,segura2017integration,scherf2023interactively,iovino2023framework}) & \multicolumn{1}{@{}l@{}}{$\cmark$(\tcite{scherf2023interactively})}\\
 \cline{2-6}
 & Non-human agents coexistence & &  $\cmark$(\tcite{abiyev2016robot}) &  &   \\
 \cline{2-6}
 &Human agents coexistence & & $\cmark$(\tcite{pezzato2023active,fusaro2021human,el2021resume,behery2023human,li2022towards})  & $\partialmark$(\tcite{ruiz2022automating,chen2024integrating,verdaguer2025boosting})& \multicolumn{1}{@{}l@{}}{$\cmark$(\tcite{verdaguer2025boosting})} \\
\hline
\end{tabular}}
\end{table}

\subsection{BT Evolution}

Evolution-based approaches use search and optimization to synthesize or improve BTs when complete models, predefined control structures, or deterministic specifications are not available. These approaches are especially useful for addressing knowledge incompleteness, uncertain mission specifications, changing resources, and execution context changes.

For knowledge incompleteness, evolutionary techniques reduce the need for a fully specified mathematical or symbolic model. Hallawa et al.~\cite{hallawa2017instinct} use Evolutionary Algorithms (EA) and Grammatical Evolution (GE) to evolve BTs over constrained pools of actions, conditions, and control nodes. Their approach links hardware parameters to high-level objectives and lets the system learn appropriate behavior from performance feedback. This allows BTs to be synthesized even when task knowledge is incomplete.

To address knowledge drift in robotic systems, evolutionary techniques can be used to enable runtime adaptations as well. Scheper et al.~\cite{scheper2016behavior} organize behaviors into intelligible sub-behaviors, so that users can identify which parts of the robot’s internal knowledge no longer match the actual environment and manually adapt thresholds, control parameters, or actions accordingly.

Evolutionary techniques also support adaptation to changing resources. Chen et al.~\cite{chen2023co} represent the robot body as a set of components and the behavior as a modular BT whose grammar is generated from the actions available for that body. When resources change, the system recomputes the action set, regenerates a grammar constrained to the current body, and re-evolves or selects a BT that remains valid under the new resource configuration.

Evolution has also been used to address uncertain mission specifications. Hallawa et al.~\cite{hallawa2020evolving} combine the Instinct Evolution Scheme (IES) with GE. IES reduces the solution space by extracting Pareto-optimal hardware configurations and translating them into action and condition pools, while GE evolves BTs from those constrained pools. Montague et al.~\cite{montague2024hierarchical} use GP to evolve low-level behavioral primitives, such as ``go-to-food'', ``reduce-density'', and ``go-away-from-nest'', and then evolve a high-level arbitrator that combines these behaviors dynamically according to environmental conditions and swarm states.

For execution context changes, evolution allows BTs to adapt to runtime sensory conditions and dynamic environments. Hallawa et al.~\cite{hallawa2017instinct} evolve BTs using GE so that agents can autonomously choose among hardware configurations, compression techniques, sensing strategies, and power-management behaviors according to runtime sensory inputs. Herranz et al.~\cite{herranz2022decentralised} use evolved BT controllers within a two-stage adaptive control loop for robot swarms, where each robot first generates desired actions and then negotiates final actions with nearby robots through decentralized communication.

Overall, evolution-based approaches increase autonomy and reduce manual design effort. However, they often require carefully designed fitness functions, may incur high computational cost, and may not provide formal guarantees of optimality or safety.

\begin{observation}{BT Evolution}
    Evolution-based BT approaches are applicable if changes are localized and simulation environments exist.
\end{observation}

\subsection{BT Extension}

Extension-based approaches modify the BT formalism or execution semantics by adding new node types, decorators, reasoning mechanisms, or integration layers. These approaches directly address the limitations of classical BTs related to static control flow, deterministic assumptions, limited task resumption, lack of quality requirements, and weak support for human-centered or partially observable environments.

For knowledge incompleteness and partial observability, Li et al.~\cite{li2022towards} extend Reactive BTs with dynamic subtree expansion. When a condition fails because of partial observability, the BT is expanded by inserting a subtree whose postcondition satisfies the failed predicate. The approach also reorders conflicting branches and uses dependence-aware rollback to preserve consistency. This allows online recovery without requiring a complete world model.

For component failures, El et al.~\cite{el2021resume} introduce H-nodes, namely Sequence and Fallback nodes with history memory. These nodes allow a robot to pause and resume execution from the correct point after interruption or failure, rather than restarting the entire task. Wang et al.~\cite{wang2020extending} extend BTs with a market/auction layer for multi-robot systems, enabling failure detection, task reassignment, and cost adaptation when robots malfunction, either partially or completely.

Several extensions focus on uncertain mission specification. Rovida et al.~\cite{rovida2017extended} propose extended BTs (eBTs), combining BTs with HTN and PDDL-based planning. Abstract goals are expanded into hierarchical skills and primitives, and the resulting tree can be reorganized at runtime to optimize time and resource use. Fusaro et al.~\cite{fusaro2021human} introduce custom nodes, such as \textit{SequenceCosts} and \textit{KeepRunningUntilSuccess}, to continuously evaluate task-related costs and reorder actions according to the minimum execution cost. Behery et al.~\cite{behery2023human} extend BTs with Human Action Nodes (HANs), integrating BTs, PDDL action models, and a CLIPS dispatcher to support collaborative human--robot assembly. Rostamnia et al.~\cite{rostamnia2025towards} introduce adaptable nodes as placeholders for uncertain portions of a mission, using FRETISH requirements to generate condition-checking subtrees and support runtime selection and hot swapping of alternative behaviors.

Extension-based approaches are also used for execution context changes. Pezzato et al.~\cite{pezzato2023active} introduce leaf nodes that specify desired states and combine BTs with Active Inference for online action selection and probabilistic reasoning. Fusaro et al.~\cite{fusaro2021human} use cost-based extensions to reprioritize actions as environmental and human-related conditions change. Li et al.~\cite{li2022towards} introduce dynamic rollback and subtree collapsing to restore execution dependencies when environmental changes or limited perception invalidate the current plan. Alberts et al.~\cite{alberts2024rebet} use ReBeT to modify both task behavior and software architecture at runtime in response to environmental, resource, and quality-related changes. Wang et al.~\cite{wang2020extending} split a single robot BT into perception, allocation, and execution subtrees, enabling dynamic task perception, distributed auction-based task allocation, and adaptive execution. Rostamnia et al.~\cite{rostamnia2025towards} use adaptable nodes to select alternative behaviors based on runtime environmental conditions.

Finally, extension approaches have been proposed for agents coexisting with the robot. Abiyev et al.~\cite{abiyev2016robot} combine BTs with fuzzy inference to handle ambiguous interactions with other agents, preserving a clear high-level strategy while translating noisy perceptions into graded actions instead of brittle discrete decisions. In human-agent coexistence scenarios, Behery et al.~\cite{behery2023human}, Pezzato et al.~\cite{pezzato2023active}, Fusaro et al.~\cite{fusaro2021human}, El et al.~\cite{el2021resume}, and Li et al.~\cite{li2022towards} extend BTs to account for human actions, interruptions, collaboration dynamics, runtime cost changes, and partial observability.

Overall, BT extensions are the most direct way to address limitations of the classical formalism. They preserve the BT structure while adding mechanisms for planning, memory, quality-awareness, probabilistic reasoning, cost optimization, human interaction, and runtime reconfiguration. Their main drawback is that each extension may change the semantics of BTs, potentially reducing standardization, analyzability, and comparability across approaches.

\subsection{BT Generation}

Generation-based approaches automatically synthesize BTs from demonstrations, task specifications, plans, symbolic knowledge, natural-language instructions, learned policies, or prior BT knowledge. These approaches are especially useful when no initial BT exists or when manual BT design would be too costly.

For knowledge incompleteness, Iovino et al.~\cite{iovino2023framework} combine Learning from Demonstration (LfD) with Genetic Programming (GP) to incrementally complete BTs when task knowledge is missing. Demonstrations provide partial task knowledge, while GP explores alternative structures to improve the generated BTs.

To address knowledge drift, Ruiz et al.~\cite{ruiz2022automating} integrate planning and ontology-based reasoning with BTs. Their proposed framework, upon discovery of real-world mismatches, updates or regenerates the symbolic, geometric, or ontological knowledge used for planning. Then, the BT executor adapts the execution flow by regenerating the tree at runtime. 

Regarding component failure in robotic systems, generation approaches focus on replanning and runtime recovery. Segura et al.~\cite{segura2017integration} address component failures by integrating Hierarchical Task Network (HTN) planning and BTs to enable continuous monitoring, reactive recovery, and dynamic replanning. When a failure occurs, the BT detects the failure and either triggers a predefined reactive behavior or initiates replanning from the current execution point rather than restarting the entire mission. 

Generation has also been used to address uncertain mission specifications. Iovino et al.~\cite{iovino2023programming} combine BTs, LfD, and GP to semi-automatically generate robot behavior. LfD enables non-expert users to teach subtasks through demonstrations, while GP evolves and optimizes the BT by exploring alternative solutions when missions change or the system gets stuck in local optima. Colledanchise et al.~\cite{colledanchise2018learning} incrementally learn BTs by mapping environmental conditions to actions using a greedy learning strategy and GP when action effects and condition meanings are initially unknown.

For execution context changes, generation approaches synthesize or regenerate BTs according to runtime conditions. Wen et al.~\cite{wen2024auction} combine market-based task allocation with automated behavior generation in multi-agent systems, enabling dynamic task reallocation without human intervention or expert pre-programming. Iovino et al.~\cite{iovino2023framework} allow robots to learn behaviors through demonstrations and improve them through GP, supporting adaptation when the execution context is not fully known at design time. Verma et al.~\cite{verma2021automatic} generate BTs from symbolic task plans and geometric motion plans to dynamically repair failed actions or regenerate trajectories when the environment changes. Zhou et al.~\cite{zhou2024llm} combine LLMs with BTs through a dynamic BT update algorithm that can add new actions with higher execution priority during runtime. Segura-Muros et al.~\cite{segura2017integration} combine HTN planning for high-level deliberation with BTs for online execution monitoring and adaptive recovery.

Generation approaches also support human-agent coexistence by translating human input into executable behavior structures. Ruiz et al.~\cite{ruiz2022automating} convert sensor outputs into symbolic facts, generate PDDL domain and problem files, and update BT XML when humans modify the environment. Chen et al.~\cite{chen2024integrating} use LLMs to translate ambiguous human instructions into formal goals and then generate a cost-minimizing BT for the specified goal.

Overall, generation-based approaches reduce manual modeling effort and make BT construction more accessible. They are particularly useful when task specifications are incomplete, when users are non-experts, or when runtime replanning is required. However, generated BTs may be difficult to read, reuse, verify, or maintain, especially when produced by opaque learning or LLM-based components.

\subsection{BT Refinement}

Refinement-based approaches start from an existing BT and modify it to repair, improve, or adapt its behavior. Unlike generation, refinement assumes that a BT already exists; unlike extension, it does not necessarily modify the BT formalism itself.

Scherf et al.~\cite{scherf2023interactively} propose ILBERT, which treats adaptation as interactive online refinement. BTs are learned from imperfect demonstrations and then refined through real-time interaction. When unseen states occur, the system detects failures, applies backchaining, and inserts corrective subtrees. This approach addresses knowledge incompleteness by repairing BTs when new situations are encountered.

The same approach also addresses knowledge inaccuracy and actuation non-determinism. For knowledge inaccuracy, ILBERT continuously checks learned action conditions against sensor observations at runtime. When mismatches or unseen states are detected, the robot asks the user for corrective input and updates the BT together with its continuous pre- and postcondition ranges~\cite{scherf2023interactively}. For actuation non-determinism, the approach performs interactive repair through three steps: detecting deviations via runtime postcondition failures, adapting condition ranges or requesting corrective demonstrations, and inserting or extending a subtree using backchaining. This enables local correction of behavior, although human-guided repair can introduce delays.

Overall, refinement approaches are useful when a BT is already available but must be corrected or adapted after deployment. They support localized repair and can preserve the existing BT structure. However, they often depend on runtime monitoring, user feedback, or additional demonstrations, and their effectiveness depends on how accurately failures and mismatches can be detected.

\subsection{Hybrid}

Some approaches combine BT generation and refinement. These approaches first create an initial BT and then improve it through search, interaction, validation, or optimization. This combination is particularly relevant because an initial input, such as a demonstration, a grammar, a natural-language instruction, or a task specification, is rarely sufficient to capture the full complexity of a real-world robotic environment. Failure cases, edge cases, hidden dependencies, changing conditions, and unexpected interactions may only become visible during execution or validation. Refinement is therefore used to correct, complete, or optimize the generated BT so that it becomes more robust and deployable.

Deng et al.~\cite{deng2023learning} propose an adaptive BT learning framework that combines GE, GP, and reusable BT knowledge resources. The framework first uses GE to solve simpler tasks by constraining the search space with user-defined grammars. If the task is more complex, the system switches to GP, which explores a broader solution space without depending heavily on predefined grammars. The framework also uses manually designed and generated BTs as prior knowledge to accelerate learning for future tasks.

Verdaguer et al.~\cite{verdaguer2025boosting} combine LLMs, BTs, and GP to automatically generate, validate, and optimize robot behaviors. In this framework, LLMs synthesize initial BTs, while GP refines and improves them. This combination reduces the manual effort of initial BT construction while using evolutionary optimization to improve the robustness and efficiency of the generated behavior.

Overall, the frequent combination of generation and refinement suggests that BT generation alone is often insufficient for adaptation in realistic robotic settings. Generation can provide an initial behavioral structure, but refinement is needed to address incompleteness, suboptimality, failure cases, and deployment-specific constraints. Hybrid approaches are therefore promising because they balance automation and correction. However, they also inherit limitations from both families: generated BTs may be unreliable or hard to interpret, while refinement may require additional computation, validation, or human involvement.

The analyzed approaches show that enhancing BTs for adaptation is not achieved through a single technique. Instead, different adaptation needs require different enhancement mechanisms. Evolution is useful when behavior must be discovered or optimized under incomplete models or uncertain goals. Extension is effective when the BT formalism must be enriched with memory, planning, cost reasoning, quality requirements, fuzzy decisions, human-aware nodes, or runtime reconfiguration. Generation reduces manual effort by synthesizing BTs from plans, demonstrations, symbolic knowledge, natural language, or learned policies. Refinement supports localized repair of existing BTs when execution failures, mismatches, or unseen states occur. Hybrid approaches attempt to obtain the benefits of automatic synthesis while improving the resulting BTs through optimization or interaction.

At the same time, these techniques introduce important trade-offs. Evolutionary and learning-based approaches can improve adaptability but may be computationally expensive and dependent on task-specific fitness functions or training data. Planning-based generation and extensions improve deliberation and recovery, but require accurate symbolic models and synchronization between planning and BT execution. Human-in-the-loop refinement improves flexibility and interpretability, but depends on the availability and quality of human feedback. LLM-based techniques lower the barrier for BT generation, but raise concerns about hallucination, domain mismatch, and verification. Finally, extensions increase expressiveness but may reduce the simplicity and standard semantics that make classical BTs attractive.

Therefore, the main conclusion for RQ3 is that adaptive BTs should be understood as a family of enhanced BT-based architectures rather than as a single formalism. Classical BTs provide the execution structure, but adaptation is typically achieved by combining them with planning, learning, monitoring, reasoning, human feedback, quality management, or runtime architectural mechanisms.

\subsection{Validation of the Suitability of BT-based Approaches}
 
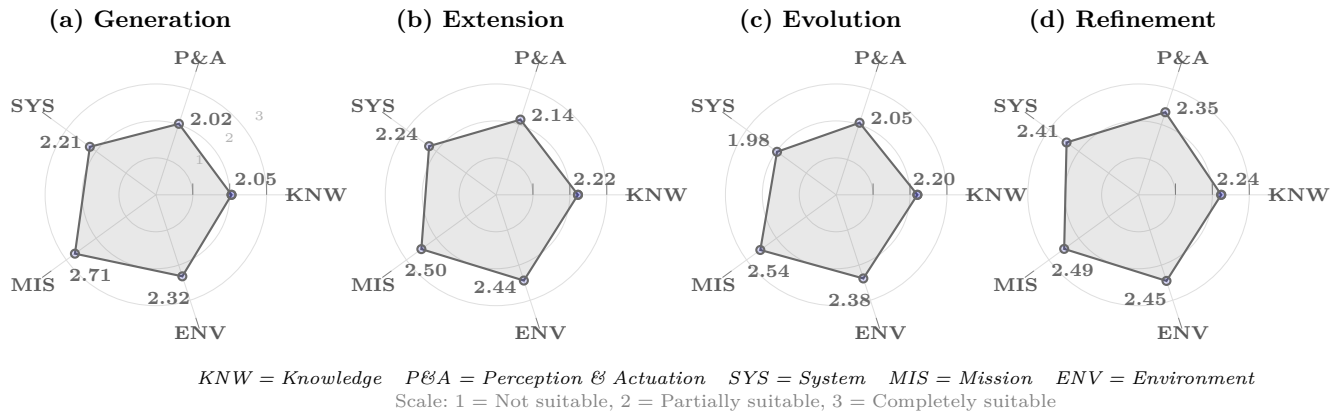
\begin{figure*}[t]
    \centering
    \begin{tikzpicture}
\definecolor{cGen}{HTML}{696969}\definecolor{cExt}{HTML}{696969}
\definecolor{cEvol}{HTML}{696969}\definecolor{cRef}{HTML}{696969}
\pgfplotsset{radarstyle/.style={
    width=5.0cm, height=5.0cm,
    xtick={0,72,144,216,288},
    xticklabels={KNW,P\&A,SYS,MIS,ENV},
    xticklabel style={font=\scriptsize\bfseries, color=black!65,inner sep=0pt},
    ymin=0, ymax=3.5, ytick={1,2,3}, yticklabels={},
    y grid style={line width=0.3pt, gray!25},
    x grid style={line width=0.3pt, gray!25},
    grid=both, axis line style={draw=none},
}}

\begin{scope}[shift={(0,0)}]
\begin{polaraxis}[radarstyle,at={(0,0)}]
\addplot+[cGen,thick,mark=*,mark size=1.5pt,fill=cGen,fill opacity=0.15,draw opacity=1]
  coordinates{(0,2.05)(72,2.02)(144,2.21)(216,2.71)(288,2.32)(360,2.05)}\closedcycle;
\coordinate(g1)at(axis cs:0,2.3);\coordinate(g2)at(axis cs:72,2.15);
\coordinate(g3)at(axis cs:144,2.4);\coordinate(g4)at(axis cs:216,2.95);
\coordinate(g5)at(axis cs:288,2.5);
\coordinate(gs1)at(axis cs:36,1);\coordinate(gs2)at(axis cs:36,2);\coordinate(gs3)at(axis cs:36,3);
\end{polaraxis}
\node[font=\scriptsize\bfseries,cGen,anchor=south,xshift=0.2cm]at(g1){2.05};
\node[font=\scriptsize\bfseries,cGen,anchor=west]at(g2){2.02};
\node[font=\scriptsize\bfseries,cGen,anchor=south east,yshift=-0.2cm,xshift=0.1cm]at(g3){2.21};
\node[font=\scriptsize\bfseries,cGen,anchor=north,xshift=0.3cm]at(g4){2.71};
\node[font=\scriptsize\bfseries,cGen,anchor=east,yshift=-0.2cm,xshift=0.2cm]at(g5){2.32};
\node[font=\tiny,gray!60,anchor=south west]at(gs1){1};
\node[font=\tiny,gray!60,anchor=south west]at(gs2){2};
\node[font=\tiny,gray!60,anchor=south west]at(gs3){3};
\node[font=\small\bfseries]at(1.5,4.0){(a) Generation};
\end{scope}

\begin{scope}[shift={(4.5,0)}]
\begin{polaraxis}[radarstyle,at={(0,0)}]
\addplot+[cExt,thick,mark=*,mark size=1.5pt,fill=cExt,fill opacity=0.15,draw opacity=1]
  coordinates{(0,2.22)(72,2.14)(144,2.24)(216,2.50)(288,2.44)(360,2.22)}\closedcycle;
\coordinate(x1)at(axis cs:0,2.47);\coordinate(x2)at(axis cs:72,2.27);
\coordinate(x3)at(axis cs:144,2.42);\coordinate(x4)at(axis cs:216,2.73);
\coordinate(x5)at(axis cs:288,2.62);
\end{polaraxis}
\node[font=\scriptsize\bfseries,cExt,anchor=south,xshift=0.1cm]at(x1){2.22};
\node[font=\scriptsize\bfseries,cExt,anchor=west]at(x2){2.14};
\node[font=\scriptsize\bfseries,cExt,anchor=south east,yshift=-0.1cm,xshift=0.05cm]at(x3){2.24};
\node[font=\scriptsize\bfseries,cExt,anchor=north,xshift=0.15cm]at(x4){2.50};
\node[font=\scriptsize\bfseries,cExt,anchor=east]at(x5){2.44};
\node[font=\small\bfseries]at(1.5,4.0){(b) Extension};
\end{scope}

\begin{scope}[shift={(9.0,0)}]
\begin{polaraxis}[radarstyle,at={(0,0)}]
\addplot+[cEvol,thick,mark=*,mark size=1.5pt,fill=cEvol,fill opacity=0.15,draw opacity=1]
  coordinates{(0,2.20)(72,2.05)(144,1.98)(216,2.54)(288,2.38)(360,2.20)}\closedcycle;
\coordinate(ev1)at(axis cs:0,2.45);\coordinate(ev2)at(axis cs:72,2.18);
\coordinate(ev3)at(axis cs:144,2.15);\coordinate(ev4)at(axis cs:216,2.77);
\coordinate(ev5)at(axis cs:288,2.56);
\end{polaraxis}
\node[font=\scriptsize\bfseries,cEvol,anchor=south,xshift=0.1cm]at(ev1){2.20};
\node[font=\scriptsize\bfseries,cEvol,anchor=west]at(ev2){2.05};
\node[font=\scriptsize\bfseries,cEvol,anchor=south east,xshift=0.1cm,yshift=-0.1cm]at(ev3){1.98};
\node[font=\scriptsize\bfseries,cEvol,anchor=north,xshift=0.2cm]at(ev4){2.54};
\node[font=\scriptsize\bfseries,cEvol,anchor=east,xshift=0.2cm,yshift=-0.2cm]at(ev5){2.38};
\node[font=\small\bfseries]at(1.5,4.0){(c) Evolution};
\end{scope}

\begin{scope}[shift={(13.0,0)}]
\begin{polaraxis}[radarstyle,at={(0,0)}]
\addplot+[cRef,thick,mark=*,mark size=1.5pt,fill=cRef,fill opacity=0.15,draw opacity=1]
  coordinates{(0,2.24)(72,2.35)(144,2.41)(216,2.49)(288,2.45)(360,2.24)}\closedcycle;
\coordinate(r1)at(axis cs:0,2.49);\coordinate(r2)at(axis cs:72,2.48);
\coordinate(r3)at(axis cs:144,2.59);\coordinate(r4)at(axis cs:216,2.72);
\coordinate(r5)at(axis cs:288,2.63);
\end{polaraxis}
\node[font=\scriptsize\bfseries,cRef,anchor=south,xshift=0.1cm]at(r1){2.24};
\node[font=\scriptsize\bfseries,cRef,anchor=west]at(r2){2.35};
\node[font=\scriptsize\bfseries,cRef,anchor=south east,xshift=0.1cm,yshift=-0.1cm]at(r3){2.41};
\node[font=\scriptsize\bfseries,cRef,anchor=north,xshift=0.2cm]at(r4){2.49};
\node[font=\scriptsize\bfseries,cRef,anchor=east,yshift=-0.15cm,xshift=0.1cm]at(r5){2.45};
\node[font=\small\bfseries]at(1.5,4.0){(d) Refinement};
\end{scope}

\node[font=\scriptsize\itshape,anchor=north]at(9.25,-0.5)
  {KNW = Knowledge\quad P\&A = Perception \& Actuation\quad SYS = System\quad MIS = Mission\quad ENV = Environment};
\node[font=\scriptsize,color=gray!98,anchor=north]at(9.25,-0.8)
  {Scale: 1 = Not suitable, 2 = Partially suitable, 3 = Completely suitable};
\end{tikzpicture}
    \caption{Mean suitability of four BT adaptation approaches across five families of adaptation needs, as rated by 46 practitioners on a three-point scale (1\,=\, Not suitable, 2\,=\, Partially suitable, 3\,=\, Completely suitable). Each panel shows one approach; the polygon shape reveals its suitability profile across families.}
    \label{fig:rq3_validation}
\end{figure*}

Respondents to the practitioners' questionnaire rated the suitability of each BT adaptation approach (Generation, Extension, Evolution, Refinement) for each family of adaptation needs on a three-point scale (\emph{Not suitable}, \emph{Partially suitable}, \emph{Completely suitable}). Figure~\ref{fig:rq3_validation} summarizes the mean suitability ratings. Across all approaches, mission-related needs received the highest suitability ratings ($M\!=\!2.49$--$2.71$, 60--74\% rated completely suitable), while perception- and actuation-related needs received the lowest ratings ($M\!=\!2.02$--$2.35$, 22--43\% completely suitable). Refinement was the highest-rated approach overall (48.8\% completely suitable, $M\!=\!2.39$), followed by Extension (44.9\%, $M\!=\!2.31$), Generation (40.7\%, $M\!=\!2.26$), and Evolution (38.2\%, $M\!=\!2.23$).
 
\emph{BT Evolution}, based on the literature, is suitable for adaptation needs that require automatic synthesis or optimization of robot behavior under uncertainty. However, its suitability is confined to specific adaptation families---knowledge~\cite{hallawa2017instinct,scheper2016behavior}, mission~\cite{montague2024hierarchical,hallawa2020evolving}, system~\cite{chen2023co}, and environment~\cite{hallawa2017instinct}---and it is not a comprehensive solution for all categories of robotic adaptation needs. Practitioner ratings confirm this pattern: suitability varied significantly across families (Friedman $\chi^2\!=\!25.65$, $p\!<\!0.001$, $n\!=\!37$), with mission-related needs rated highest ($M\!=\!2.54$, 61\% completely suitable) and system-related needs rated lowest ($M\!=\!1.98$, 34\%).

Practitioners acknowledged that the ability of evolutionary methods to search for improved BT structures makes them attractive for addressing adaptation needs. Evolution can optimize BTs for slowly changing environments by evaluating candidate behaviors over representative scenarios, while also incorporating constraints such as battery consumption or resource usage into the fitness function. Nevertheless, practitioners expressed reservations about online evolution. Beyond their computational cost, they emphasized that evolutionary search fundamentally relies on iterative fitness evaluation and trial-and-error optimization, making it structurally different from the immediate, deterministic responses required to handle operational changes such as knowledge inconsistencies, sensor degradation, or hardware failures. One practitioner directly highlighted the risk of this trial-and-error process:
\begin{pquote}
    {\it ``When you do reinforcement learning, the robot needs to try things. Online, in the real world, this is dangerous. [...] In the real environment, you cannot really undo all the actions.''}
\end{pquote}
Such situations instead require reactive execution mechanisms provided by the BT itself (e.g., fallback behaviors), rather than an online optimization process.

Practitioners also highlighted the difficulty of validating the safety of evolved behaviors and of encoding complex adaptation objectives into fitness functions. Consequently, they viewed BT Evolution primarily as an offline design-time technique for discovering or optimizing BTs, with only limited applicability to online adaptation. Therefore, BT Evolution is most suitable when the required adaptation can be achieved by optimizing the BT before deployment, without relying on continuous runtime re-optimization or other online adaptation mechanisms.

\emph{BT Extension} approaches, based on the literature, are suitable for adaptation needs that primarily require enriching BT execution semantics (e.g., knowledge incompleteness~\cite{li2022towards}, component failures~\cite{el2021resume,wang2020extending}, uncertain quality goals~\cite{rostamnia2025towards,fusaro2021human}, and non-human~\cite{abiyev2016robot} and human-agent coexistence~\cite{li2022towards,fusaro2021human,el2021resume}). However, they are only partially suitable for adaptation needs that require reasoning beyond BT execution, such as uncertain mission specification and execution-context changes. In these cases, BT extensions improve the adaptability of execution but still depend on external modules---such as planners, perception systems, probabilistic inference engines, knowledge bases, or architectural monitoring---to determine or support the adaptation. 
Practitioner ratings showed moderate variation across families (Friedman $\chi^2\!=\!9.72$, $p\!=\!0.045$, $n\!=\!38$), with mission-related needs rated highest ($M\!=\!2.50$, 62\% completely suitable) and perception- and actuation-related needs rated lowest ($M\!=\!2.14$, 31\%). 

Practitioners largely agreed with this execution-centric view, but further emphasized that BT extensions primarily increase the expressiveness of the BT formalism rather than providing a complete adaptation framework. One practitioner illustrated this point by describing many extensions as convenient abstractions over classical BT constructs:
\begin{pquote}
    {\it ``You can do pretty much everything with Parallel, Fallback, and Sequence. [...] The other [constructs] can be implemented with the classical nodes; they are probably syntactic sugar.''}
\end{pquote}
They generally considered BT extensions suitable when adaptation can be expressed through richer execution logic (e.g., mission execution, fault handling, or human interaction). In contrast, they viewed BT extensions as partially suitable when adaptation fundamentally depends on external capabilities---such as perception, knowledge acquisition, planning, learning, or resource management---because these capabilities remain outside of the BT itself. This suggests that BT extensions should be understood as execution-level adaptation mechanisms that complement, rather than substitute for, higher-level reasoning and system infrastructure.

\emph{BT Generation} is partially suitable for addressing most of the adaptation needs in robotic systems. Its primary strength lies in automatically synthesizing or regenerating executable robot behaviors from demonstrations~\cite{iovino2023framework,scherf2023interactively}, symbolic knowledge~\cite{chen2024integrating}, learned policies~\cite{zhou2024llm}, planners~\cite{ruiz2022automating,segura2017integration,wu2022rbt,verma2021automatic}, or natural-language instructions~\cite{zhou2024llm,chen2024integrating}. However, for most adaptation families, including knowledge, system, and environment, the BT is the product of adaptation rather than the mechanism performing the adaptation. External components, such as planners, knowledge bases, ontologies, LLMs, runtime monitors, or learning algorithms, first determine how the robot should adapt, after which BT generation constructs the corresponding executable behavior. This distinction was explicitly described by one practitioner: 
\begin{pquote}
    {\it ``If the mission is uncertain, you need a learning or planning module that provides the BT as a result.''}
\end{pquote}
The notable exception is uncertain mission specification, where generating the BT is itself the adaptation objective. In these cases, BT generation directly addresses the uncertainty by synthesizing the required behavior without relying on a pre-existing BT, making generation approaches suitable for this adaptation need. Practitioner ratings strongly reflect this distinction: suitability varied significantly across families (Friedman $\chi^2\!=\!32.91$, $p\!<\!0.001$, $n\!=\!37$), with mission-related needs rated highest ($M\!=\!2.71$, 74\% completely suitable---the highest single cell in the entire matrix) and perception- and actuation-related needs rated lowest ($M\!=\!2.02$, 22\%).

Practitioners distinguished between generating robot behavior and performing adaptation. BT generation is considered suitable only when the adaptation problem itself is to synthesize a new behavior (mission adaptation). For all other adaptation families, adaptation first requires external reasoning---such as updating knowledge, interpreting perception, monitoring system resources, or analyzing environmental changes---after which BT generation produces or regenerates the corresponding execution policy. Consequently, BT generation is viewed as a behavior synthesis mechanism rather than a complete adaptation mechanism.

\begin{observation}{BT Generation}
BT Generation is less suitable for adaptation needs-- such as perception- and system-related adaptation, because BTs cannot repair perception algorithms or faulty hardware. 
\end{observation}

\emph{BT Refinement}, based on the literature, is considered suitable to address adaptation needs in robotics. BT Refinement assumes an existing BT and directly modifies it to accommodate new situations, repair failures, or improve behaviors. Consequently, the BT remains the central adaptation artifact. Runtime monitoring, user feedback, demonstrations~\cite{scherf2023interactively}, or evolutionary optimization~\cite{deng2023learning,verdaguer2025boosting} provide information on when and how to adapt, but the adaptation itself is realized through modifications to the BT structure or its execution conditions. Therefore, refinement is suitable for knowledge-related, actuation-related, mission-related, and environment-related adaptation needs where an existing BT can be incrementally repaired or improved. Uniquely among the four approaches, practitioner ratings for Refinement did not vary significantly across families (Friedman $\chi^2\!=\!4.73$, $p\!=\!0.316$, $n\!=\!36$), confirming its broad applicability. Refinement also had the lowest ``Not suitable'' overall rate (9.7\%, compared to 14--15\% for the other three approaches).

Practitioners agreed that refinement excels at adapting execution to changes in the environment, system behavior, and mission execution by updating parameters, conditions, recovery strategies, or replacing subtrees without redesigning the whole BT. One practitioner characterized this incremental process as follows:
\begin{pquote}
    {\it ``Refinement is actually [a] proper use of the BT. You start with the tree [...], and then you extend [and] enrich [its leaves] when you realize that you need more things.''}
\end{pquote}
However, they considered BT refinement only partially suitable for knowledge-, actuation-, and perception-related adaptation needs. Although refinement can update thresholds, conditions, action parameters, or locally modify decision logic based on new information, it does not fundamentally acquire, represent, or reason about new knowledge, nor can it adapt perception independently of the underlying sensing and control algorithms. These adaptations still depend on external perception, learning, or knowledge-management mechanisms, while refinement incorporates their outcomes into the existing BT. 

Practitioners also emphasized that refinement preserves the intent of an existing BT. Consequently, it is most appropriate for incremental adaptation rather than radical behavioral changes. If the robot's mission, domain assumptions, or operating conditions change substantially, refinement alone may become insufficient, and regeneration or redesign of the BT may be required. Thus, refinement should be viewed as a complementary adaptation mechanism that continuously improves an existing BT rather than replacing approaches such as BT generation for fundamentally new behaviors.

\begin{observation}{BT Refinement}
Refinement optimizes existing behaviors but is not suitable for fundamentally changing mission objectives. For instance, for mission-related adaptation needs, replacing abstract actions with more detailed subtrees can be effectively accomplished with BT refinement (e.g., drive $\rightarrow$ drive under harsh terrain conditions). 
\end{observation}

\begin{tcolorbox}[
    enhanced,
    breakable,
    colback=gray!4!white,
    colframe=gray!55!black,
    coltitle=white,
    colbacktitle=black!55!black,
    title=RQ3: Enhancement of BTs for adaptation,
    fonttitle=\bfseries,
    boxrule=0.8pt,
    arc=1mm

]
Existing approaches enhance BTs for adaptation through four main families-- \emph{evolution}, \emph{extension}, \emph{generation}, \emph{refinement}-- as well as \emph{hybrid} approaches. Evolution discovers or optimizes BTs under incomplete models or uncertain goals. Extension enriches BTs with new semantics, nodes, decorators, memory, planning, quality-awareness, or human-aware mechanisms. Generation synthesizes BTs from demonstrations, plans, symbolic knowledge, natural language, learned policies, or prior BT knowledge. Refinement repairs or improves existing BTs during or after execution. Hybrid approaches first synthesize BTs and then improve them through optimization, interaction, or runtime correction.

These techniques substantially improve the adaptability of BT-based robotic systems, but they also introduce trade-offs in computational cost, scalability, interpretability, verification, dependence on models or human input, and preservation of BT simplicity. Therefore, adaptive BTs are best understood as enhanced BT-based architectures in which classical BTs provide the modular execution structure, while adaptation is achieved through integration with planning, learning, monitoring, reasoning, quality management, human feedback, or runtime architectural mechanisms. More broadly, these approaches reveal that BTs can play different roles in adaptation: they can act as the adaptation mechanism, constitute the artifact being adapted, or represent the executable product of adaptation performed by external mechanisms.
\end{tcolorbox}

\section{Discussion}
\label{sec:discussion}

This section synthesizes the findings from the analysis of adaptation needs, classical BT features and limitations, and adaptive BT approaches. The discussion is grounded in the two mappings produced in RQ2 and RQ3. Table~\ref{tbl:map_adaptation_needs_to_BT_features_limitations} shows whether classical BT features are sufficient, partially sufficient, or insufficient for the identified adaptation needs, while Table~\ref{tbl:map_adaptation_needs_to_approaches_suitability} shows which families of adaptive BT approaches have been used to address those needs. Taken together, the two mappings show that BTs are a useful behavioral backbone, but not a complete adaptation framework.

\subsection{Main Findings}

Our first finding is that classical BTs support adaptation only when the adaptation space is sufficiently bounded. Their modularity, hierarchy, reactivity, maintainability, and reuse make them effective for representing known tasks, composing reusable behaviors, checking monitored conditions, and encoding predefined recovery strategies. 
This is particularly visible for adaptation needs whose effects can be handled through localized behavioral changes, including some cases arising from knowledge incompleteness, knowledge inaccuracy, knowledge drift, actuation non-determinism, and execution-context changes.

Our second finding is that several adaptation needs expose structural limitations of classical BTs. Mission uncertainty, execution-context changes, component failures, changing resources, and human-agent coexistence are only partially supported because they require more than predefined conditions and fallback branches. These needs are associated with limitations such as static control flow, fixed execution order, deterministic assumptions, weak handling of partial observability, limited task resumption, lack of explicit intent representation, and insufficient support for quality requirements. This explains why the literature often addresses them through extensions, generation, refinement, evolution, or integration with planning and learning mechanisms.

Our third finding is that some needs remain weakly addressed. The mappings show limited explicit support for perception-driven adaptation, actuation unavailability, new resources, communication impairment, future mission changes, outdated missions, uncertain quality goals, and uncertain mission relationships. This does not mean that BTs cannot be integrated with mechanisms addressing these needs; rather, it indicates that the analyzed BT literature rarely treats them as first-class adaptation concerns.

Our fourth finding concerns the role that the BT itself plays in the adaptation process. Our analysis reveals three distinct roles. First, the BT can be the \emph{mechanism of adaptation}, when its reactive execution directly selects among predefined behaviors in response to changing conditions. Second, the BT can be the \emph{object of adaptation}, when its structure or execution logic is modified through refinement or evolution. Third, the BT can be the \emph{product of adaptation}, when an external mechanism, such as a planner, learning component, knowledge-based reasoner, or LLM, determines how the system should adapt and generates a corresponding BT for execution. BT extensions may span these roles by incorporating additional adaptation capabilities into the BT formalism itself. This distinction clarifies that ``BT-based adaptation’’ does not necessarily imply that adaptation is performed by the BT: in many approaches, the BT remains the structured execution layer within a broader adaptive architecture.

Finally, enhanced BT approaches broaden the applicability of BTs but introduce new trade-offs. Evolution supports the discovery and optimization of behavior under incomplete models or uncertain goals. Extension enriches BTs with mechanisms such as memory, quality-awareness, human-aware nodes, fuzzy reasoning, planning, or runtime reconfiguration. Generation synthesizes BTs from plans, demonstrations, symbolic models, natural language, or learned policies. Refinement repairs existing BTs after failures, mismatches, or unseen states. These approaches improve adaptability, but they also raise concerns about computational cost, scalability, interpretability, verification, and dependence on models or human input.

\subsection{Research Directions}

\emph{Perception-driven adaptation} emerges as one of the clearest gaps. Table~\ref{tbl:map_adaptation_needs_to_BT_features_limitations} does not report explicit evidence that classical BT features support perception multiplicity, perception inaccuracy, or perception unavailability, and Table~\ref{tbl:map_adaptation_needs_to_approaches_suitability} does not identify approaches in our corpus that explicitly target these needs. Future research should investigate how BTs can be integrated with perception monitoring, active perception, adaptive sensor fusion, and uncertainty-aware state estimation. For industrial systems, this means that BT-based control should not assume perception outputs are always reliable; perception quality should itself become an adaptation trigger.

\emph{Distributed and resource-aware adaptation} is also underdeveloped. Although some multi-robot approaches address allocation, coordination, and execution in dynamic settings~\cite{wang2020extending,wen2024auction,herranz2022decentralised}, communication impairment is rarely modeled as a first-class adaptation need. Similarly, actuation unavailability and newly available resources receive limited explicit treatment, even though they are central in long-running and safety-critical robotic systems. Future work should investigate BTs connected to explicit capability models, communication-quality monitors, and fallback coordination protocols, so that behavior can be adapted when actuators, sensors, software services, or communication channels become unavailable or newly available.

\emph{Mission and quality management} require stronger support. The literature addresses uncertain mission specification more extensively than future mission changes, outdated missions, uncertain mission relationships, or uncertain quality goals. This suggests that current BT-based adaptation research often focuses on generating or modifying behavior for a given mission, while paying less attention to the mission lifecycle during long-running operations. Quality requirements are partially addressed by approaches such as ReBeT~\cite{alberts2024rebet}, but safety, efficiency, robustness, energy consumption, timeliness, privacy, and security are still often encoded indirectly through conditions, decorators, or external modules. Future BT approaches should make mission revision, mission conflict detection, and quality-aware trade-offs first-class modeling and execution concerns.

\emph{Human-aware adaptation} remains a central challenge. Classical BTs can react to monitored conditions, but they do not explicitly represent human intentions, preferences, availability, interruptions, safety constraints, or interaction costs. Existing work addresses parts of this challenge through cost-aware nodes, Human Action Nodes, history-aware nodes, dynamic expansion, planning integration, and LLM-based generation~\cite{fusaro2021human,behery2023human,el2021resume,li2022towards,chen2024integrating,verdaguer2025boosting}. However, human-aware adaptation requires more than reactive execution: it requires reasoning about collaboration, safety, task ownership, privacy, and changing human behavior.

Several cross-cutting challenges affect all these directions. First, adaptive BTs increasingly depend on planning, learning, and reasoning. Planning-based approaches require accurate symbolic models and reliable world-state representations; learning-based approaches require data, reward or fitness functions, and computational resources~\cite{chen2024integrating,yang2023robot,iovino2023framework}; and LLM-based approaches raise concerns about hallucination, domain mismatch, validation, and reliability~\cite{zhou2024llm,verdaguer2025boosting,chen2024integrating}. Second, adaptive BTs must remain scalable and maintainable. Large, generated, evolved, or runtime-modified BTs may be difficult to understand, debug, validate, and reuse~\cite{yang2023robot,verma2021automatic}. Third, runtime restructuring introduces dependability concerns: when BTs are modified during execution, the system must preserve consistency with the current execution state and avoid unsafe or unreachable behavior~\cite{romeiro2026towards, rovida2017extended,montague2024hierarchical,wen2024auction,rostamnia2025towards}. These concerns point to the need for verification and assurance mechanisms for adaptive BTs, including runtime monitoring, safety-invariant checking, and validation of generated or refined subtrees~\cite{caldas2024runtime, luckcuck2019formal}.

\begin{observation}{Interview insight: Emerging role of BTs and LLMs} 
 Interview participants anticipated a changing division of responsibilities between BTs and LLMs. BTs provide structured mission control and orchestration for robot behavior, while learned models and LLMs perform increasingly sophisticated decision-making and action execution. They envisioned that BTs should be kept as a higher layer to orchestrate the robot behaviors, and LLMs should be used at a lower level. Current BTs contain many fine-grained actions, e.g., look at object, move arm, close gripper, and grasp, but future low-level sequences may disappear. One BT leaf node could invoke an intelligent model that performs the complete grasping procedure. This can provide richer contextual reasoning than decomposing every step into many separate BT nodes. Furthermore, this can add the benefit of fewer fine-grained BT nodes, handling complex execution inside leaf nodes, and larger intelligent action modules. Besides, to ensure deterministic behavior, there should be an agentic component on top of BT to decide, reason about, and interpret the mission to avoid uncertainties in mission specifications and relationships.   

\end{observation}

\emph{Layered integration of BTs and learned models} emerges as an additional research direction. Future research should investigate architectures in which agentic components interpret missions and propose adaptations, BTs provide structured orchestration, and learned models implement complex execution skills. A key challenge is to define explicit boundaries between these responsibilities: generative or learned components can expand the space of possible behaviors, but should not implicitly acquire authority over mission constraints or physical execution. Such architectures may require extensions to the BT formalism and well-defined interfaces through which intelligent components expose execution states, uncertainty estimates, assumptions, failure modes, and interruption semantics. This information can support independent validation and runtime monitoring, so that generated adaptations and learned behaviors remain consistent with mission and safety requirements.

A related direction concerns \emph{behavioral languages designed for both machine manipulation and human understanding}. As behaviors are increasingly generated or modified automatically at runtime, these languages should make explicit which parts of a mission may change and which constraints must remain invariant. Future behavioral languages should preserve desirable properties of BTs, such as hierarchy, modularity, and reactivity, while providing machine-oriented representations for generation and controlled runtime modification together with human-readable projections for inspection, validation, and assurance.

\subsection{Implications}

For software engineering and robotics researchers, the main implication is that BT adaptation should not be studied only as a tree-transformation problem. It is a broader software engineering challenge concerning software evolution, requirements engineering, architecture, and runtime monitoring. Adaptation needs originate from knowledge, perception, actuation, system resources, mission objectives, and the environment that may interact with multiple software and physical components. New BT-based approaches should therefore state which adaptation needs they address, how adaptation is triggered, what part of the system is adapted, and whether the mechanism relies on classical BT features, BT extensions, generation, refinement, evolution, or external reasoning.

For practitioners, particularly robotic software engineers, autonomy architects, and verification and safety engineers, the mappings help clarify when classical BTs are sufficient and when enhanced mechanisms are required. Classical BTs are appropriate when robot behavior can be specified through known tasks, explicit conditions, and predefined recovery strategies. Enhanced BT mechanisms are needed when the robot must cope with incomplete knowledge, partial observability, resource changes, evolving missions, human intervention, quality-related trade-offs, or runtime context shifts. In such cases, BTs should be treated as part of a broader adaptive architecture rather than as a complete adaptation solution.

From a broader industrial software engineering perspective, these findings also affect how BT-based robotic systems should be architected and maintained. Adaptation should not be confined to the behavioral model, but treated as a system-level concern spanning perception, knowledge management, planning, resource management, monitoring, and execution. This requires explicit interfaces between the BT and these components, together with mechanisms for tracing adaptation decisions and managing changes to behavior throughout the system lifecycle. In safety- and mission-critical applications, dynamically generated, refined, or evolved behaviors also raise assurance and configuration-management concerns: engineers need to determine which parts of the behavior may change at runtime, under which conditions, and how modified behaviors are validated before or during execution. The choice of a BT adaptation mechanism therefore becomes an architectural and lifecycle decision, rather than only a choice of behavior representation.

These findings also have implications for the engineering tools used to develop and operate BT-based robotic systems. When BTs are generated, refined, evolved, or reconfigured at runtime, they can no longer be treated only as static design artifacts. Tool support should therefore cover the lifecycle of evolving BTs, including versioning and provenance of behavioral changes, traceability between adaptation triggers and resulting modifications, runtime inspection and debugging, and validation of generated or modified behaviors. In safety- and mission-critical settings, such support should also enable engineers to determine which version of a BT was executed, why an adaptation occurred, and what evidence supports the acceptability of the resulting behavior.

The choice of enhancement technique should follow the adaptation need. Evolution is suitable when behavior must be discovered or optimized, but it requires careful fitness design and computational resources. Extension is useful when classical BT semantics lack necessary constructs, but it may reduce standardization and analyzability. Generation reduces manual design effort, but generated BTs require validation and often refinement. Refinement supports localized repair, but may depend on runtime monitoring, human feedback, or additional demonstrations. Combined generation--refinement is promising because it balances automation and correction, but it inherits the challenges of both families.

\subsection{Summary}

The main message is that adaptive robotic behavior cannot be achieved by classical BTs alone. It needs to be integrated with mechanisms for monitoring, reasoning, learning, and runtime adaptation. Classical BTs provide a readable, modular, and reactive execution structure, but mainly for adaptation needs that can be anticipated and encoded through predefined structures. Enhanced BT approaches broaden this support through evolution, extension, generation, and refinement. Therefore, the future of adaptive BTs is likely to involve hybrid architectures in which BTs remain the behavioral backbone, while complementary mechanisms detect adaptation needs, select appropriate responses, and ensure that the resulting behavior remains safe, explainable, maintainable, and aligned with mission and quality requirements.

\section{Related work}\label{sec:related-work}

\begin{table*}[t]
\centering
\caption{Positioning of our work with respect to related studies.}
\label{tab:related-work-comparison}

\small
\setlength{\tabcolsep}{4pt}
\renewcommand{\arraystretch}{1.08}

\begin{tabular}{@{}
    >{\raggedright\arraybackslash}p{2.65cm}
    >{\centering\arraybackslash}p{0.95cm}
    >{\raggedright\arraybackslash}p{5.6cm}
    >{\raggedright\arraybackslash}p{7.9cm}
@{}}
\toprule
\textbf{Study} &
\textbf{Year} &
\textbf{Main focus} &
\textbf{Difference from our work} \\
\midrule

Iovino et al.~\cite{iovino2022survey}
& 2022
& Survey of BTs in robotics and game AI, including concepts, applications, design patterns, synthesis, and learning.
& Provides a broad overview of BTs, but does not organize the analysis around robotic adaptation needs or assess classical BT limitations with respect to such needs. \\

\"{O}gren and Sprague~\cite{ogren2022behavior}
& 2022
& Review of BTs from a robot-control perspective, including modularity, hierarchy, reactivity, planning, and control.
& Focuses on BT principles and control-oriented properties rather than on adaptation needs and adaptation-oriented BT techniques. \\

Biggar et al.~\cite{biggar2020principled}
& 2020
& Analysis of BTs as a model of action selection, with emphasis on modularity, reactivity, and formal properties.
& Studies BTs as an action-selection formalism, but does not provide a robotics-specific taxonomy of adaptation needs or map such needs to BT capabilities. \\

Shin and Jung~\cite{shin2024survey}
& 2024
& Survey of BT-based task planning for autonomous robotic systems, especially reinforcement learning and learning from demonstration.
& Focuses on learning-based task planning, whereas our work covers broader adaptation needs and different forms of BT enhancement. \\

Ghzouli et al.~\cite{ghzouli2023behavior}
& 2023
& Empirical study of BTs and state machines in ROS-based robotics applications.
& Studies language concepts, usage patterns, and reuse in robotic software, but does not analyze adaptation needs or classify BT adaptation mechanisms. \\

Iovino et al.~\cite{iovino2023programming}
& 2023
& Comparison of the programming effort required to implement and modify BTs and finite state machines.
& Focuses on implementation and modification effort, but does not investigate why robotic systems need adaptation or how BT-based approaches address such needs. \\

Dragule et al.~\cite{DRAGULE2025101330}
& 2025
& Empirical study on the effects of specifying robotic missions with BTs and state machines, focusing on comprehension, correctness, and usability.
& Studies mission specification effects from a human and language-usability perspective, while our work focuses on adaptation needs and adaptation mechanisms. \\

Filippone et al.~\cite{Filippone2026Formalisms}
& 2026
& Comparative analysis of BTs, state machines, Hierarchical Task Networks, and BPMN for robotic mission specification and execution, considering control structures, expressiveness, limitations, and tool support.
& Compares alternative formalisms for specifying robotic missions, whereas our work focuses on adaptation needs and assesses how classical and enhanced BTs address such needs and what limitations remain or emerge. \\

\textit{Our work}
& --
& Taxonomy and literature-driven analysis of robotic adaptation needs and the capabilities of classical and enhanced BTs in addressing them.
& Identifies and classifies robotic adaptation needs, assesses the capabilities and limitations of classical BTs, and characterizes and evaluates BT generation, extension, evolution, refinement, and hybrid approaches in terms of how they address these needs and what limitations remain or emerge. \\

\bottomrule
\end{tabular}
\end{table*}

This section positions our study with respect to prior work that is related in scope and purpose. Since this paper analyzes BT-based adaptation approaches as primary studies, we do not discuss individual approaches that generate, extend, evolve, or refine BTs in this section. Those studies are examined when answering RQ2 and RQ3. Instead, we focus on prior surveys, overviews, and empirical studies that characterize BTs, compare BTs with state machines, or investigate their use for robotic behavior and mission specification.

\subsection{Surveys and Overviews on Behavior Trees}

Several surveys and overview papers have studied BTs in robotics, artificial intelligence, and control. Iovino et al.~\cite{iovino2022survey} provide a broad survey of BTs in robotics and game AI, covering BT concepts, application domains, design patterns, and methods for BT synthesis and learning. \"{O}gren and Sprague~\cite{ogren2022behavior} review BTs from a robot-control perspective, emphasizing their modularity, hierarchy, reactivity, and connections with planning and control. Biggar et al.~\cite{biggar2020principled} analyze BTs as a principled model of action selection and discuss their relation to modularity and reactivity. More recently, Shin and Jung~\cite{shin2024survey} survey BT-based task planning for autonomous robotic systems, with a specific focus on reinforcement learning and learning from demonstration.

These studies provide important foundations for understanding BTs and their role in robotic systems. However, their main objective is to review BT principles, applications, formal properties, or integrations with learning and planning. They do not start from the adaptation needs of robotic systems, nor do they systematically analyze whether such needs are supported by classical BTs, require BT extensions, or remain open challenges.

\subsection{Empirical Studies on BTs and State Machines}

A second line of work compares BTs with state machines, which remain a common formalism for specifying robotic behavior. Ghzouli et al.~\cite{ghzouli2023behavior} empirically study BTs and state machines in robotics applications by analyzing domain-specific languages and open-source ROS projects. Their work provides evidence on how these formalisms are used in practice, including their language concepts and reuse patterns. Iovino et al.~\cite{iovino2023programming} compare the programming effort required to implement and modify BTs and finite state machines for robotic applications, showing how BT modularity can affect development and maintenance effort. Dragule et al.~\cite{DRAGULE2025101330} investigate the effects of specifying robotic missions with BTs and state machines, focusing on comprehension, correctness, and usability. 
Filippone et al.~\cite{Filippone2026Formalisms} broaden this comparison by systematically analyzing BTs alongside state machines, Hierarchical Task Networks, and BPMN as formalisms for robotic mission specification and execution. Their study compares the formalisms in terms of their underlying control structures and mission concepts, expressiveness and limitations in modeling real-world missions, and available tool support, providing guidance on their respective strengths and shortcomings for single- and multi-robot mission specification.

These studies are particularly relevant from a software engineering perspective because they analyze BTs as artifacts used to specify, understand, and modify robotic behavior and missions. Nevertheless, their focus differs from ours. They compare BTs with alternative formalisms primarily in terms of their modeling characteristics, practical use, development effort, expressiveness, comprehension, and usability. In contrast, our work focuses specifically on adaptation: we identify the adaptation needs of robotic systems, assess the capabilities and limitations of classical BTs with respect to these needs, and analyze to what extent existing BT-based approaches overcome these limitations and what limitations remain or emerge.

\subsection{Positioning of This Study}

Our study complements prior surveys and empirical studies by connecting two perspectives that have mostly been treated separately: adaptation in robotic systems and BT-based behavior specification. Rather than reviewing BTs in general, or comparing BTs with state machines only in terms of usability or programming effort, we ask whether BTs are sufficient to meet the adaptation needs of modern robotic systems.

To this end, we first derive a taxonomy of adaptation needs from the robotics and self-adaptive systems literature. We then use this taxonomy to analyze the capabilities and limitations of classical BTs, and to characterize existing BT-based adaptation approaches. Table~\ref{tab:related-work-comparison} summarizes how our work differs from existing surveys and empirical studies.

\section{Conclusion} \label{sec:conclusion}
In this paper, we examined whether BTs are sufficient to meet the adaptation needs of modern robotic systems operating in dynamic, uncertain, and open-ended environments. To answer this question, we first identified and classified adaptation needs reported in the robotics literature into six categories: \emph{Knowledge}, \emph{Perception}, \emph{Actuation}, \emph{System}, \emph{Mission}, and \emph{Environment}. We then analyzed the features and limitations of classical BTs in addressing these needs. We reviewed 31 primary studies proposing BT-based approaches for adaptation. We classified these approaches into categories including \emph{generation}, \emph{extension}, \emph{evolution}, and \emph{refinement}. Finally, we strengthened our findings through three validation activities: author validation questionnaires, a survey of robotics researchers and practitioners, and semi-structured interviews.

Our findings demonstrate that classical BTs provide several software engineering advantages, including modularity, hierarchy, reusability, readability, and reactive execution. These properties make them a robust and practical framework for structuring robotic behavior. However, they are not sufficient on their own to address many of the adaptation needs faced by modern robotic systems. In particular, classical BTs provide limited support for runtime behavioral restructuring, reasoning under uncertainty, mission reinterpretation, continuous learning, and coordination with planning, perception, and knowledge-management components. While the BT-based approaches proposed in the literature mitigate some of these shortcomings by introducing additional adaptation mechanisms, each focuses on a limited set of challenges, and none provides comprehensive support across all adaptation categories.

Beyond reviewing existing techniques, this survey establishes a connection between robotic adaptation needs and the capabilities of current BT-based approaches. The resulting classification and mapping provide a practical reference for researchers and practitioners, helping them choose appropriate techniques for different adaptation scenarios. Additionally, it highlights cases where complementary methods are necessary. 

Our analysis suggests that future behavior modeling frameworks should treat adaptation as a core design concern and provide explicit interfaces to planning, reasoning, learning, monitoring, and runtime evolution mechanisms, rather than assuming that adaptation can be handled by the behavioral model alone.

This work also highlights several promising directions for future research. These include developing unified behavior modeling frameworks that combine multiple adaptation mechanisms, strengthening formal guarantees for adaptive behavior, and establishing systematic engineering methods for designing, analyzing, and validating adaptive robotic control architectures. Furthermore, future research should extend the classification of the adaptation needs in robotic systems to include additional concerns related to safety, security, ethics, and regulation. 

\section*{Acknowledgments}
This work has been partially funded by 
(a) the MUR (Italy) Department of Excellence 2023 - 2027, 
(b) the European HORIZON-KDT-JU research project MATISSE ``Model-based engineering of Digital Twins for early verification and validation of Industrial Systems", HORIZON-KDT-JU-2023-2-RIA, Proposal number:  101140216-2, KDT232RIA\_00017, and
(c) the Space It Up project funded by the Italian Space Agency and the Ministry of University and Research CUP: D53C24000580005.

\bibliographystyle{ieeetr}
\bibliography{ref}

@article{ghzouli2023behavior,
  title={Behavior trees and state machines in robotics applications},
  author={Ghzouli, Razan and Berger, Thorsten and Johnsen, Einar Broch and Wasowski, Andrzej and Dragule, Swaib},
  journal={IEEE Transactions on Software Engineering},
  volume={49},
  number={9},
  pages={4243--4267},
  year={2023},
  publisher={IEEE}
}

@article{hezavehi2021uncertainty,
  title={Uncertainty in self-adaptive systems: A research community perspective},
  author={Hezavehi, Sara M and Weyns, Danny and Avgeriou, Paris and Calinescu, Radu and Mirandola, Raffaela and Perez-Palacin, Diego},
  journal={ACM Transactions on Autonomous and Adaptive Systems (TAAS)},
  volume={15},
  number={4},
  pages={1--36},
  year={2021},
  publisher={ACM New York, NY}
}

@incollection{mahdavi2017classification,
  title={A classification framework of uncertainty in architecture-based self-adaptive systems with multiple quality requirements},
  author={Mahdavi-Hezavehi, Sara and Avgeriou, Paris and Weyns, Danny},
  booktitle={Managing trade-offs in adaptable software architectures},
  pages={45--77},
  year={2017},
  publisher={Elsevier}
}

@book{weyns2020introduction,
  title={An introduction to self-adaptive systems: A contemporary software engineering perspective},
  author={Weyns, Danny},
  year={2020},
  publisher={John Wiley \& Sons}
}

@INPROCEEDINGS{ramirez2012taxonomy,
  author={Ramirez, Andres J. and Jensen, Adam C. and Cheng, Betty H. C.},
  booktitle={2012 7th International Symposium on Software Engineering for Adaptive and Self-Managing Systems (SEAMS)}, 
  title={A taxonomy of uncertainty for dynamically adaptive systems}, 
  year={2012},
  volume={},
  number={},
  pages={99-108},
  doi={10.1109/SEAMS.2012.6224396}
}

@article{wen2024auction,
  author    = {Shanghua Wen and Wendi Wu and Ning Li and Ji Wang and Shaowu Yang and Chi Ben and Wenjing Yang},
  title     = {Auction-Based Behavior Tree Evolution for Heterogeneous Multi-Agent Systems},
  journal   = {Applied Sciences},
  year      = {2024},
  volume    = {14},
  number    = {17},
  pages     = {7896},
  doi       = {10.3390/app14177896},
  url       = {https://www.mdpi.com/2076-3417/14/17/7896},
  publisher = {MDPI}
}

@article{scherf2023interactively,
  title={Interactively learning behavior trees from imperfect human demonstrations},
  author={Scherf, Lisa and Schmidt, Aljoscha and Pal, Suman and Koert, Dorothea},
  journal={Frontiers in Robotics and AI},
  volume={10},
  pages={1152595},
  year={2023},
  publisher={Frontiers Media SA}
}

@inproceedings{iovino2023framework,
  title={A framework for learning behavior trees in collaborative robotic applications},
  author={Iovino, Matteo and Styrud, Jonathan and Falco, Pietro and Smith, Christian},
  booktitle={2023 IEEE 19th International Conference on Automation Science and Engineering (CASE)},
  pages={1--8},
  year={2023},
  organization={IEEE}
}

@inproceedings{iovino2023programming,
  title = {On the Programming Effort Required to Generate Behavior Trees and Finite State Machines for Robotic Applications},
  author = {Iovino, Matteo and F{\"o}rster, Julian and Falco, Pietro and Chung, Jen Jen and Siegwart, Roland and Smith, Christian},
  booktitle = {2023 IEEE International Conference on Robotics and Automation (ICRA)},
  pages = {5807--5813},
  year = {2023},
  organization = {IEEE},
  doi = {10.1109/ICRA48891.2023.10161033}
}

@article{Filippone2026Formalisms,
  author={Filippone, Gianluca and Pettinari, Sara and Pelliccione, Patrizio},
  journal={IEEE Transactions on Software Engineering}, 
  title={Formalisms for Robotic Mission Specification and Execution: A Comparative Analysis}, 
  year={2026},
  volume={},
  number={},
  pages={1-32},
  doi={10.1109/TSE.2026.3725356}
}

@article{pezzato2023active,
  title={Active inference and behavior trees for reactive action planning and execution in robotics},
  author={Pezzato, Corrado and Corbato, Carlos Hern{\'a}ndez and Bonhof, Stefan and Wisse, Martijn},
  journal={IEEE Transactions on Robotics},
  volume={39},
  number={2},
  pages={1050--1069},
  year={2023},
  publisher={IEEE}
}

@inproceedings{verma2021automatic,
  title={Automatic generation of behavior trees for the execution of robotic manipulation tasks},
  author={Verma, Parikshit and Diab, Mohammed and Rosell, Jan},
  booktitle={2021 26th IEEE International Conference on Emerging Technologies and Factory Automation (ETFA)},
  pages={1--4},
  year={2021},
  organization={IEEE}
}

@INPROCEEDINGS{fusaro2021human,
  author={Fusaro, Fabio and Lamon, Edoardo and Momi, Elena De and Ajoudani, Arash},
  booktitle={2020 IEEE-RAS 20th International Conference on Humanoid Robots (Humanoids)}, 
  title={A Human-Aware Method to Plan Complex Cooperative and Autonomous Tasks using Behavior Trees}, 
  year={2021},
  volume={},
  number={},
  pages={522-529},
  doi={10.1109/HUMANOIDS47582.2021.9555683}}

@inproceedings{rovida2017extended,
  title={Extended behavior trees for quick definition of flexible robotic tasks},
  author={Rovida, Francesco and Grossmann, Bjarne and Kr{\"u}ger, Volker},
  booktitle={2017 IEEE/RSJ International Conference on Intelligent Robots and Systems (IROS)},
  pages={6793--6800},
  year={2017},
  organization={IEEE}
}

@inproceedings{zhou2024llm,
  title={Llm-bt: Performing robotic adaptive tasks based on large language models and behavior trees},
  author={Zhou, Haotian and Lin, Yunhan and Yan, Longwu and Zhu, Jihong and Min, Huasong},
  booktitle={2024 IEEE International Conference on Robotics and Automation (ICRA)},
  pages={16655--16661},
  year={2024},
  organization={IEEE}
}

@inproceedings{montague2024hierarchical,
  title={A hierarchical approach to evolving behaviour-trees for swarm control},
  author={Montague, Kirsty and Hart, Emma and Paechter, Ben},
  booktitle={International Conference on the Applications of Evolutionary Computation (Part of EvoStar)},
  pages={178--193},
  year={2024},
  organization={Springer}
}

@article{ruiz2022automating,
  title={Automating adaptive execution behaviors for robot manipulation},
  author={Ruiz-Celada, Oriol and Verma, Parikshit and Diab, Mohammed and Rosell, Jan},
  journal={IEEE access},
  volume={10},
  pages={123489--123497},
  year={2022},
  publisher={IEEE}
}

@inproceedings{el2021resume,
  title={To resume or not to resume: A behavior tree extension},
  author={El-Ariss, Wafic and Daher, Naseem and Elhajj, Imad H},
  booktitle={2021 American Control Conference (ACC)},
  pages={770--776},
  year={2021},
  organization={IEEE}
}

@inproceedings{hallawa2020evolving,
  title={Evolving instinctive behaviour in resource-constrained autonomous agents using grammatical evolution},
  author={Hallawa, Ahmed and Schug, Simon and Iacca, Giovanni and Ascheid, Gerd},
  booktitle={International Conference on the Applications of Evolutionary Computation (Part of EvoStar)},
  pages={369--383},
  year={2020},
  organization={Springer}
}

@inproceedings{herranz2022decentralised,
  title={Decentralised negotiation for multi-object collective transport with robot swarms},
  author={Herranz, Guillermo Legarda and Hauert, Sabine and Jones, Simon},
  booktitle={2022 IEEE International Conference on Autonomous Robot Systems and Competitions (ICARSC)},
  pages={186--191},
  year={2022},
  organization={IEEE}
}

@inproceedings{deng2023learning,
  title={Learning behavior trees by evolution-inspired approaches},
  author={Deng, Chuanshuai and Zhao, Chenjing and Liu, Zhenghui and Zhang, Jiexin and Wu, Yunlong and Wang, Yanzhen and Cheng, Hong and Yi, Xiaodong},
  booktitle={Proceedings of the Companion Conference on Genetic and Evolutionary Computation},
  pages={275--278},
  year={2023}
}

@inproceedings{behery2023human,
  title={Human robot collaborative assembly using behavior trees and dynamic tree dispatching},
  author={Behery, Mohamed and Deutsch, Jonas and Trinh, Minh and Koetter, David and Brecher, Christian and Lakemeyer, Gerhard},
  booktitle={ISR Europe 2023; 56th International Symposium on Robotics},
  pages={258--263},
  year={2023},
  organization={VDE}
}

@inproceedings{segura2017integration,
  title={Integration of an automated hierarchical task planner in ros using behaviour trees},
  author={Segura-Muros, Jos{\'e} {\'A}ngel and Fern{\'a}ndez-Olivares, Juan},
  booktitle={2017 6th International Conference on Space Mission Challenges for Information Technology (SMC-IT)},
  pages={20--25},
  year={2017},
  organization={IEEE}
}

@inproceedings{li2022towards,
  title={Towards adaptive behavior trees for robot task planning},
  author={Li, Ning and Jiang, Hao and Li, Chunpeng and Wang, Zhaoqi},
  booktitle={2022 China Automation Congress (CAC)},
  pages={6720--6725},
  year={2022},
  organization={IEEE}
}

@article{chen2024integrating,
  title={Integrating intent understanding and optimal behavior planning for behavior tree generation from human instructions},
  author={Chen, Xinglin and Cai, Yishuai and Mao, Yunxin and Li, Minglong and Yang, Wenjing and Xu, Weixia and Wang, Ji},
  journal={arXiv preprint arXiv:2405.07474},
  year={2024}
}

@inproceedings{alberts2024rebet,
  title={Rebet: Architecture-based self-adaptation of robotic systems through behavior trees},
  author={Alberts, Elvin and Gerostathopoulos, Ilias and Stoico, Vincenzo and Lago, Patricia},
  booktitle={2024 IEEE International Conference on Autonomic Computing and Self-Organizing Systems (ACSOS)},
  pages={1--10},
  year={2024},
  organization={IEEE}
}

@article{scheper2016behavior,
  title={Behavior trees for evolutionary robotics},
  author={Scheper, Kirk YW and Tijmons, Sjoerd and de Visser, Cornelis C and de Croon, Guido CHE},
  journal={Artificial life},
  volume={22},
  number={1},
  pages={23--48},
  year={2016},
  publisher={MIT Press One Rogers Street, Cambridge, MA 02142-1209, USA journals-info~…}
}

@article{abiyev2016robot,
  title={Robot soccer control using behaviour trees and fuzzy logic},
  author={Abiyev, Rahib H and G{\"u}nsel, Irfan and Akkaya, Nurullah and Aytac, Ersin and {\c{C}}a{\u{g}}man, Ahmet and Abizada, Sanan},
  journal={Procedia Computer Science},
  volume={102},
  pages={477--484},
  year={2016},
  publisher={Elsevier}
}

@inproceedings{verdaguer2025boosting,
  title={Boosting Behavior Tree Generation for Robots with Large Language Models and Genetic Programming},
  author={Verdaguer-Gonzalez, Aaron and Dalmau-Moreno, Mag{\'\i} and Merino, Luis and Garc{\'\i}a, N{\'e}stor},
  booktitle={2025 IEEE International Conference on Simulation, Modeling, and Programming for Autonomous Robots (SIMPAR)},
  pages={1--6},
  year={2025},
  organization={IEEE}
}

@inproceedings{yang2023robot,
  title={Robot Behavior Tree Manipulation Using Language Models},
  author={Yang, Zhuo and Jia, Zhizhou},
  booktitle={2023 IEEE 11th Joint International Information Technology and Artificial Intelligence Conference (ITAIC)},
  volume={11},
  pages={1342--1345},
  year={2023},
  organization={IEEE}
}

@inproceedings{wu2022rbt,
  title={Rbt-hci: A reliable behavior tree planning method with human-computer interaction},
  author={Wu, Yunlong and Li, Jinghua and Jin, Haoxiang and Zhang, Jiexin and Wang, Yanzhen},
  booktitle={2022 IEEE International Conference on Robotics and Biomimetics (ROBIO)},
  pages={1637--1642},
  year={2022},
  organization={IEEE}
}

@inproceedings{chen2023co,
  title={Co-Designing Body and Behavior via Planning-based Hierarchical Grammatical Evolution},
  author={Chen, Xinglin and Huang, Da and Li, Minglong and Cai, Yishuai and Cai, Zhongxuan and Yang, Wenjing},
  booktitle={2023 3rd International Conference on Robotics, Automation and Intelligent Control (ICRAIC)},
  pages={102--109},
  year={2023},
  organization={IEEE}
}

@article{colledanchise2018learning,
  title={Learning of behavior trees for autonomous agents},
  author={Colledanchise, Michele and Parasuraman, Ramviyas and {\"O}gren, Petter},
  journal={IEEE Transactions on Games},
  volume={11},
  number={2},
  pages={183--189},
  year={2018},
  publisher={IEEE}
}

@inproceedings{wang2020extending,
  title={Extending behavior trees with market-based task allocation in dynamic environments},
  author={Wang, Tao and Shi, Dianxi and Yi, Wei},
  booktitle={Proceedings of the 2020 4th International Symposium on Computer Science and Intelligent Control},
  pages={1--8},
  year={2020}
}

@inproceedings{lemasurier2024reactive,
  title={Reactive or proactive? how robots should explain failures},
  author={LeMasurier, Gregory and Gautam, Alvika and Han, Zhao and Crandall, Jacob W and Yanco, Holly A},
  booktitle={Proceedings of the 2024 ACM/IEEE International Conference on Human-Robot Interaction},
  pages={413--422},
  year={2024}
}

@inproceedings{hallawa2017instinct,
  title={Instinct-driven dynamic hardware reconfiguration: evolutionary algorithm optimized compression for autonomous sensory agents},
  author={Hallawa, Ahmed and De Roose, Jaro and Andraud, Martin and Verhelst, Marian and Ascheid, Gerd},
  booktitle={Proceedings of the Genetic and Evolutionary Computation Conference Companion},
  pages={1727--1734},
  year={2017}
}

@inproceedings{rostamnia2025towards,
  title={Towards Adaptable and Uncertainty-Aware Behavior Trees},
  author={Rostamnia, Mehran and Filippone, Gianluca and Caldas, Ricardo and Pelliccione, Patrizio},
  booktitle={2025 IEEE/ACM 7th International Workshop on Robotics Software Engineering (RoSE)},
  pages={9--16},
  year={2025},
  organization={IEEE}
}

@techreport{euRobotics2016MAR,
  author       = {{euRobotics aisbl}},
  title        = {Robotics 2020: Multi-Annual Roadmap for Robotics in Europe -- Horizon 2020 Call ICT-2017 (ICT-25, ICT-27 \& ICT-28), Release B},
  institution  = {euRobotics},
  address      = {Brussels, Belgium},
  year         = {2016},
  month        = {December},
  note         = {Version B, 02/12/2016},
  url          = {https://eu-robotics.net/wp-content/uploads/H2020_Robotics_Multi-Annual_Roadmap_ICT-2017B-3.pdf},
}

@inproceedings{shin2024survey,
  title={A Survey of Behavior Tree-Based Task Planning Algorithms for Autonomous Robotic Systems},
  author={Shin, Mingyu and Jung, Soyi},
  booktitle={2024 15th International Conference on Information and Communication Technology Convergence (ICTC)},
  pages={2039--2041},
  year={2024},
  organization={IEEE}
}

@article{ogren2022behavior,
  title={Behavior trees in robot control systems},
  author={{\"O}gren, Petter and Sprague, Christopher I},
  journal={Annual Review of Control, Robotics, and Autonomous Systems},
  volume={5},
  number={1},
  pages={81--107},
  year={2022},
  publisher={Annual Reviews}
}

@article{biggar2020principled,
  title={A principled analysis of behavior trees and their generalisations},
  author={Biggar, Oliver and Zamani, Mohammad and Shames, Iman},
  journal={arXiv preprint arXiv:2008.11906},
  year={2020}
}

@article{iovino2022survey,
  title={A survey of behavior trees in robotics and ai},
  author={Iovino, Matteo and Scukins, Edvards and Styrud, Jonathan and {\"O}gren, Petter and Smith, Christian},
  journal={Robotics and Autonomous Systems},
  volume={154},
  pages={104096},
  year={2022},
  publisher={Elsevier}
}

@article{weyns2023towards,
  title={Towards a research agenda for understanding and managing uncertainty in self-adaptive systems},
  author={Weyns, Danny and Calinescu, Radu and Mirandola, Raffaela and Tei, Kenji and Acosta, Maribel and Bencomo, Nelly and Bennaceur, Amel and Boltz, Nicolas and Bures, Tomas and Camara, Javier and others},
  journal={ACM SIGSOFT Software Engineering Notes},
  volume={48},
  number={4},
  pages={20--36},
  year={2023},
  publisher={ACM New York, NY, USA}
}

@inproceedings{calinescu2020understanding,
  title={Understanding uncertainty in self-adaptive systems},
  author={Calinescu, Radu and Mirandola, Raffaela and Perez-Palacin, Diego and Weyns, Danny},
  booktitle={2020 ieee international conference on autonomic computing and self-organizing systems (acsos)},
  pages={242--251},
  year={2020},
  organization={IEEE}
}

@inproceedings{esfahani2013uncertainty,
  title={Uncertainty in self-adaptive software systems},
  author={Esfahani, Naeem and Malek, Sam},
  booktitle={Software engineering for self-adaptive systems II: International seminar, Dagstuhl Castle, Germany, October 24-29, 2010 Revised selected and invited papers},
  pages={214--238},
  year={2013},
  organization={Springer}
}

@article{DRAGULE2025101330,
title = {Effects of specifying robotic missions in behavior trees and state machines},
journal = {Journal of Computer Languages},
volume = {85},
pages = {101330},
year = {2025},
issn = {2590-1184},
doi = {https://doi.org/10.1016/j.cola.2025.101330},
url = {https://www.sciencedirect.com/science/article/pii/S2590118425000164},
author = {Swaib Dragule and Engineer Bainomugisha and Patrizio Pelliccione and Thorsten Berger},
}

@inproceedings{perez2014uncertainties,
  title={Uncertainties in the modeling of self-adaptive systems: A taxonomy and an example of availability evaluation},
  author={Perez-Palacin, Diego and Mirandola, Raffaela},
  booktitle={Proceedings of the 5th ACM/SPEC international conference on Performance engineering},
  pages={3--14},
  year={2014}
}

@book{colledanchise2018behavior,
  title     = {Behavior Trees in Robotics and AI: An Introduction},
  author    = {Colledanchise, Michele and {\"O}gren, Petter},
  year      = {2018},
  publisher = {CRC Press},
  address   = {Boca Raton, FL},
  isbn      = {9781138593732}
}

@article{dragule2021survey,
  title={A survey on the design space of end-user-oriented languages for specifying robotic missions},
  author={Dragule, Swaib and Berger, Thorsten and Menghi, Claudio and Pelliccione, Patrizio},
  journal={Software and Systems Modeling},
  volume={20},
  number={4},
  pages={1123--1158},
  year={2021},
  publisher={Springer}
}

@article{kurdila2019dynamics,
  title={Dynamics and control of robotic systems},
  author={Kurdila, Andrew J and Ben-Tzvi, Pinhas},
  year={2019},
  publisher={John Wiley \& Sons}
}

@article{durrant2006simultaneous,
  title={Simultaneous localization and mapping: part I},
  author={Durrant-Whyte, Hugh and Bailey, Tim},
  journal={IEEE robotics \& automation magazine},
  volume={13},
  number={2},
  pages={99--110},
  year={2006},
  publisher={IEEE}
}

@inproceedings{best2022resilient,
  title={Resilient multi-sensor exploration of multifarious environments with a team of aerial robots},
  author={Best, Graeme and Garg, Rohit and Keller, John and Hollinger, Geoffrey A and Scherer, Sebastian},
  booktitle={Robotics: Science and Systems (RSS)},
  year={2022}
}

@inproceedings{feng2020center,
  title={Center-of-mass-based robust grasp planning for unknown objects using tactile-visual sensors},
  author={Feng, Qian and Chen, Zhaopeng and Deng, Jun and Gao, Chunhui and Zhang, Jianwei and Knoll, Alois},
  booktitle={2020 IEEE International Conference on Robotics and Automation (ICRA)},
  pages={610--617},
  year={2020},
  organization={IEEE}
}

@article{davchev2022residual,
  title={Residual learning from demonstration: Adapting dmps for contact-rich manipulation},
  author={Davchev, Todor and Luck, Kevin Sebastian and Burke, Michael and Meier, Franziska and Schaal, Stefan and Ramamoorthy, Subramanian},
  journal={IEEE Robotics and Automation Letters},
  volume={7},
  number={2},
  pages={4488--4495},
  year={2022},
  publisher={IEEE}
}

@INPROCEEDINGS{shuai2020research,
  author={Shuai, Sun and Lei, Wang and Zhiping, Li and Peng, Gu and Feifei, Chen and Yuting, Feng},
  booktitle={2020 5th International Conference on Communication, Image and Signal Processing (CCISP)}, 
  title={Research on Parallel System for Motion States Monitoring of the Planetary Rover}, 
  year={2020},
  volume={},
  number={},
  pages={86-90},
  doi={10.1109/CCISP51026.2020.9273460}}

@article{street2023formal,
  title={Formal modelling for multi-robot systems under uncertainty},
  author={Street, Charlie and Mansouri, Masoumeh and Lacerda, Bruno},
  journal={Current Robotics Reports},
  volume={4},
  number={3},
  pages={55--64},
  year={2023},
  publisher={Springer}
}

@inproceedings{askarpour2021robomax,
  title={Robomax: Robotic mission adaptation exemplars},
  author={Askarpour, Mehrnoosh and Tsigkanos, Christos and Menghi, Claudio and Calinescu, Radu and Pelliccione, Patrizio and Garc{\'\i}a, Sergio and Caldas, Ricardo and Von Oertzen, Tim J and Wimmer, Manuel and Berardinelli, Luca and others},
  booktitle={2021 International Symposium on Software Engineering for Adaptive and Self-Managing Systems (SEAMS)},
  pages={245--251},
  year={2021},
  organization={IEEE}
}

@article{cornejo2020survey,
  title={A survey of ontologies for simultaneous localization and mapping in mobile robots},
  author={Cornejo-Lupa, Maria A and Ticona-Herrera, Regina P and Cardinale, Yudith and Barrios-Aranibar, Dennis},
  journal={ACM Computing Surveys (CSUR)},
  volume={53},
  number={5},
  pages={1--26},
  year={2020},
  publisher={ACM New York, NY, USA}
}

@article{aljalbout2025reality,
  title={The reality gap in robotics: Challenges, solutions, and best practices},
  author={Aljalbout, Elie and Xing, Jiaxu and Romero, Angel and Akinola, Iretiayo and Garrett, Caelan Reed and Heiden, Eric and Gupta, Abhishek and Hermans, Tucker and Narang, Yashraj and Fox, Dieter and others},
  journal={Annual Review of Control, Robotics, and Autonomous Systems},
  volume={9},
  year={2025},
  publisher={Annual Reviews}
}

@article{zhang2022dynamic,
  title={Dynamic information fusion in multi-source incomplete interval-valued information system with variation of information sources and attributes},
  author={Zhang, Xiaoyan and Chen, Xiuwei and Xu, Weihua and Ding, Weiping},
  journal={Information Sciences},
  volume={608},
  pages={1--27},
  year={2022},
  publisher={Elsevier}
}

@ARTICLE{Berrio2022camera,
  author={Berrio, Julie Stephany and Shan, Mao and Worrall, Stewart and Nebot, Eduardo},
  journal={IEEE Transactions on Intelligent Transportation Systems}, 
  title={Camera-LIDAR Integration: Probabilistic Sensor Fusion for Semantic Mapping}, 
  year={2022},
  volume={23},
  number={7},
  pages={7637-7652},
  doi={10.1109/TITS.2021.3071647}}

@article{rabiee2023introspective,
  title={Introspective perception for mobile robots},
  author={Rabiee, Sadegh and Biswas, Joydeep},
  journal={Artificial Intelligence},
  volume={324},
  pages={103999},
  year={2023},
  publisher={Elsevier}
}

@article{corso2022risk,
  title={Risk-driven design of perception systems},
  author={Corso, Anthony and Katz, Sydney and Innes, Craig and Du, Xin and Ramamoorthy, Subramanian and Kochenderfer, Mykel J},
  journal={Advances in Neural Information Processing Systems},
  volume={35},
  pages={9894--9906},
  year={2022}
}

@article{elfes2002using,
  title={Using occupancy grids for mobile robot perception and navigation},
  author={Elfes, Alberto},
  journal={Computer},
  volume={22},
  number={6},
  pages={46--57},
  year={2002},
  publisher={IEEE}
}

@article{mulubika2025approach,
  title={An approach towards mobile robot recovery due to vision sensor failure in vSLAM systems using ROS},
  author={Mulubika, Chibaye and Schreve, Kristiaan},
  journal={Robotica},
  volume={43},
  number={2},
  pages={720--742},
  year={2025},
  publisher={Cambridge University Press}
}

@article{cully2015robots,
  title={Robots that can adapt like animals},
  author={Cully, Antoine and Clune, Jeff and Tarapore, Danesh and Mouret, Jean-Baptiste},
  journal={Nature},
  volume={521},
  number={7553},
  pages={503--507},
  year={2015},
  publisher={Nature Publishing Group UK London}
}

@inproceedings{phillips2020planning,
  title={Planning and resilient execution of policies for manipulation in contact with actuation uncertainty},
  author={Phillips-Grafflin, Calder and Berenson, Dmitry},
  booktitle={Algorithmic Foundations of Robotics XII: Proceedings of the Twelfth Workshop on the Algorithmic Foundations of Robotics},
  pages={752--767},
  year={2020},
  organization={Springer}
}

@inproceedings{guida2022simulation,
  title={Simulation of the effects of backlash on the performance of a collaborative robot: a preliminary case study},
  author={Guida, Roberto and Raviola, Andrea and Migliore, Domenico Fabio and De Martin, Andrea and Mauro, Stefano and Sorli, Massimo},
  booktitle={International Conference on Robotics in Alpe-Adria Danube Region},
  pages={28--35},
  year={2022},
  organization={Springer}
}

@article{panda2025integrating,
  title={Integrating attention-based GRU with event-driven NMPC to enhance tracking performance of robotic manipulator under actuator failure},
  author={Panda, Atanu and Ghosh, Lidia and Mahapatra, Subhasish},
  journal={Expert Systems with Applications},
  volume={267},
  pages={125946},
  year={2025},
  publisher={Elsevier}
}

@article{chen2015kinematic,
  title={Kinematic analysis and motion planning of a quadruped robot with partially faulty actuators},
  author={Chen, Xianbao and Gao, Feng and Qi, Chenkun and Tian, Xinghua and Wei, Lin},
  journal={Mechanism and Machine Theory},
  volume={94},
  pages={64--79},
  year={2015},
  publisher={Elsevier}
}

@article{harris2021online,
  title={Online plan modification in uncertain resource-constrained environments},
  author={Harris, Catherine A and Hawes, Nick and Dearden, Richard},
  journal={Robotics and Autonomous Systems},
  volume={140},
  pages={103726},
  year={2021},
  publisher={Elsevier}
}

@article{verma2006scalable,
  title={Scalable robot fault detection and identification},
  author={Verma, Vandi and Simmons, Reid},
  journal={Robotics and Autonomous Systems},
  volume={54},
  number={2},
  pages={184--191},
  year={2006},
  publisher={Elsevier}
}

@article{lu2024event,
  title={Event-triggered resilient joint mobile robot localization and sensor fault estimation},
  author={Lu, Yanyang and Karimi, Hamid Reza and Li, Bin and Chen, Chih-Chiang},
  journal={International Journal of Robust and Nonlinear Control},
  volume={34},
  number={16},
  pages={10971--10989},
  year={2024},
  publisher={Wiley Online Library}
}

@article{khalastchi2018fault,
  title={On fault detection and diagnosis in robotic systems},
  author={Khalastchi, Eliahu and Kalech, Meir},
  journal={ACM Computing Surveys (CSUR)},
  volume={51},
  number={1},
  pages={1--24},
  year={2018},
  publisher={ACM New York, NY, USA}
}

@article{arrichiello2017distributed,
  title={Distributed fault-tolerant control for networked robots in the presence of recoverable/unrecoverable faults and reactive behaviors},
  author={Arrichiello, Filippo and Marino, Alessandro and Pierri, Francesco},
  journal={Frontiers in Robotics and AI},
  volume={4},
  pages={2},
  year={2017},
  publisher={Frontiers Media SA}
}

@article{notomista2021resilient,
  title={A resilient and energy-aware task allocation framework for heterogeneous multirobot systems},
  author={Notomista, Gennaro and Mayya, Siddharth and Emam, Yousef and Kroninger, Christopher and Bohannon, Addison and Hutchinson, Seth and Egerstedt, Magnus},
  journal={IEEE Transactions on Robotics},
  volume={38},
  number={1},
  pages={159--179},
  year={2021},
  publisher={IEEE}
}

@inproceedings{braberman2015morph,
  title={Morph: A reference architecture for configuration and behaviour self-adaptation},
  author={Braberman, Victor and D'Ippolito, Nicolas and Kramer, Jeff and Sykes, Daniel and Uchitel, Sebastian},
  booktitle={Proceedings of the 1st international workshop on control theory for software engineering},
  pages={9--16},
  year={2015}
}

@article{gielis2022critical,
  title={A critical review of communications in multi-robot systems},
  author={Gielis, Jennifer and Shankar, Ajay and Prorok, Amanda},
  journal={Current robotics reports},
  volume={3},
  number={4},
  pages={213--225},
  year={2022},
  publisher={Springer}
}

@article{ilnytska2020loss,
  title={Loss estimation for network-connected UAV/RPAS communications},
  author={Ilnytska, Svitlana I and Li, Fengping and Grekhov, Andrii and Kondratiuk, Vasyl},
  journal={IEEE Access},
  volume={8},
  pages={137702--137710},
  year={2020},
  publisher={IEEE}
}

@online{robmosys_separation_levels_concerns,
  title        = {Separation of Levels and Separation of Concerns},
  author       = {{RobMoSys Consortium}},
  year         = {2017},
  url          = {https://www.robmosys.eu/wiki-sn-05/general_principles:separation_of_levels_and_separation_of_concerns},
  organization = {RobMoSys Project},
  note         = {RobMoSys Wiki, accessed 2026-03-06}
}

@ARTICLE{mengi2023mission,
  author={Menghi, Claudio and Tsigkanos, Christos and Askarpour, Mehrnoosh and Pelliccione, Patrizio and Vázquez, Gricel and Calinescu, Radu and García, Sergio},
  journal={IEEE Transactions on Software Engineering}, 
  title={Mission Specification Patterns for Mobile Robots: Providing Support for Quantitative Properties}, 
  year={2023},
  volume={49},
  number={4},
  pages={2741-2760},
  doi={10.1109/TSE.2022.3230059}}

@article{wilde2024statistically,
  title={Statistically distinct plans for multiobjective task assignment},
  author={Wilde, Nils and Alonso-Mora, Javier},
  journal={IEEE Transactions on Robotics},
  volume={40},
  pages={2217--2232},
  year={2024},
  publisher={IEEE}
}

@article{ghose2025ve,
  title={I’ve Changed My Mind: Robots Adapting to Changing Human Goals during Collaboration},
  author={Ghose, Debasmita and Gitelson, Oz and Jin, Ryan and Abawe, Grace and V{\'a}zquez, Marynel and Scassellati, Brian},
  journal={IEEE Robotics and Automation Letters},
  volume={11},
  number={2},
  pages={1490--1497},
  year={2025},
  publisher={IEEE}
}

@incollection{weyns2019software,
  title={Software engineering of self-adaptive systems},
  author={Weyns, Danny},
  booktitle={Handbook of software engineering},
  pages={399--443},
  year={2019},
  publisher={Springer}
}

@article{zudaire2022assured,
  title={Assured mission adaptation of UAVs},
  author={Zudaire, Sebasti{\'a}n A and Nahabedian, Leandro and Uchitel, Sebasti{\'a}n},
  journal={ACM Transactions on Autonomous and Adaptive Systems (TAAS)},
  volume={16},
  number={3-4},
  pages={1--27},
  year={2022},
  publisher={ACM New York, NY}
}

@misc{larsen2024robotic,
  title={Robotic safe adaptation in unprecedented situations: the RoboSAPIENS project. Res. Direct. Cyber-Phys. Syst. 2, e4 (2024)},
  author={Larsen, PG and others},
  year={2024}
}

@article{petrovska2022defining,
  title={Defining adaptivity and logical architecture for engineering (smart) self-adaptive cyber--physical systems},
  author={Petrovska, Ana and Kugele, Stefan and Hutzelmann, Thomas and Beffart, Theo and Bergemann, Sebastian and Pretschner, Alexander},
  journal={Information and Software Technology},
  volume={147},
  pages={106866},
  year={2022},
  publisher={Elsevier}
}

@inproceedings{sun2021uncertain,
  title={Uncertain-aware safe exploratory planning using Gaussian process and neural control contraction metric},
  author={Sun, Dawei and Khojasteh, Mohammad Javad and Shekhar, Shubhanshu and Fan, Chuchu},
  booktitle={Learning for dynamics and control},
  pages={728--741},
  year={2021},
  organization={PMLR}
}

@article{russell1995modern,
  title={Artificial intelligence: a modern approach},
  author={Russell, Stuart Jonathan and Norvig, Peter and Canny, John F and Malik, Jitendra M and Edwards, Douglas D},
  volume={2},
  number={9},
  year={1995},
  publisher={Prentice hall Englewood Cliffs, NJ}
}

@article{zhang2023adaptive,
  title={Adaptive safety-critical control with uncertainty estimation for human--robot collaboration},
  author={Zhang, Dianhao and Van, Mien and Mcllvanna, Stephen and Sun, Yuzhu and McLoone, Se{\'a}n},
  journal={IEEE Transactions on Automation Science and Engineering},
  volume={21},
  number={4},
  pages={5983--5996},
  year={2023},
  publisher={IEEE}
}

@inproceedings{ghadirzadehsensorimotor,
  author={Ghadirzadeh, Ali and Bütepage, Judith and Maki, Atsuto and Kragic, Danica and Björkman, Mårten},
  booktitle={2016 IEEE/RSJ International Conference on Intelligent Robots and Systems (IROS)}, 
  title={A sensorimotor reinforcement learning framework for physical Human-Robot Interaction}, 
  year={2016},
  volume={},
  number={},
  pages={2682-2688},
  doi={10.1109/IROS.2016.7759417}}

@article{landgraf2021animal,
  title={Animal-in-the-loop: using interactive robotic conspecifics to study social behavior in animal groups},
  author={Landgraf, Tim and Gebhardt, Gregor HW and Bierbach, David and Romanczuk, Pawel and Musiolek, Lea and Hafner, Verena V and Krause, Jens},
  journal={Annual Review of Control, Robotics, and Autonomous Systems},
  volume={4},
  number={1},
  pages={487--507},
  year={2021},
  publisher={Annual Reviews}
}

@inproceedings{honda2024replan,
  title={When to replan? an adaptive replanning strategy for autonomous navigation using deep reinforcement learning},
  author={Honda, Kohei and Yonetani, Ryo and Nishimura, Mai and Kozuno, Tadashi},
  booktitle={2024 IEEE International Conference on Robotics and Automation (ICRA)},
  pages={6650--6656},
  year={2024},
  organization={IEEE}
}

@article{alberts2025software,
  title={Software architecture-based self-adaptation in robotics},
  author={Alberts, Elvin and Gerostathopoulos, Ilias and Malavolta, Ivano and Corbato, Carlos Hern{\'a}ndez and Lago, Patricia},
  journal={Journal of Systems and Software},
  volume={219},
  pages={112258},
  year={2025},
  publisher={Elsevier}
}

@inproceedings{brugali2019non,
  title={Non-functional requirements in robotic systems: Challenges and state of the art},
  author={Brugali, Davide},
  booktitle={2019 IEEE International Conference on Real-time Computing and Robotics (RCAR)},
  pages={743--748},
  year={2019},
  organization={IEEE}
}

@article{chen2026learning,
  title={Learning unknown reward function for drone navigation based on inverse deep reinforcement learning},
  author={Chen, Zhe and Xuan, Junyu},
  journal={Neural Computing and Applications},
  volume={38},
  number={4},
  pages={51},
  year={2026},
  publisher={Springer}
}

@article{kwan2025onboard,
  title={Onboard Mission Replanning for Adaptive Cooperative Multi-Robot Systems},
  author={Kwan, Elim and Qureshi, Rehman and Fletcher, Liam and Laganier, Colin and Nockles, Victoria and Walters, Richard},
  journal={IEEE Robotics and Automation Letters},
  volume={10},
  number={12},
  pages={13225--13232},
  year={2025},
  publisher={IEEE}
}

@inproceedings{bramblett2024robust,
  title={Robust online epistemic replanning of multi-robot missions},
  author={Bramblett, Lauren and Miloradovi{\'c}, Branko and Sherman, Patrick and Papadopoulos, Alessandro V and Bezzo, Nicola},
  booktitle={2024 IEEE/RSJ International Conference on Intelligent Robots and Systems (IROS)},
  pages={13229--13236},
  year={2024},
  organization={IEEE}
}

@article{debie2023swarm,
  title={Swarm robotics: A survey from a multi-tasking perspective},
  author={Debie, Essam and Kasmarik, Kathryn and Garratt, Matt},
  journal={ACM Computing Surveys},
  volume={56},
  number={2},
  pages={1--38},
  year={2023},
  publisher={ACM New York, NY}
}

@article{qiao2024simultaneous,
  title={Simultaneous localization and mapping (SLAM)-based robot localization and navigation algorithm},
  author={Qiao, Junfu and Guo, Jinqin and Li, Yongwei},
  journal={Applied Water Science},
  volume={14},
  number={7},
  pages={151},
  year={2024},
  publisher={Springer}
}

@inproceedings{wolf2023modularity,
  title={Modularity in humanoid robot design for flexibility in system structure and application},
  author={Wolf, Sebastian and Hofmann, Cynthia and Bahls, Thomas and Maurenbrecher, Henry and Pleintinger, Benedikt},
  booktitle={2023 IEEE-RAS 22nd International Conference on Humanoid Robots (Humanoids)},
  pages={1--7},
  year={2023},
  organization={IEEE}
}

@article{tziafas2022enhancing,
  title={Enhancing interpretability and interactivity in robot manipulation: A neurosymbolic approach},
  author={Tziafas, Georgios and Kasaei, Hamidreza},
  journal={The International Journal of Robotics Research},
  pages={02783649251412867},
  publisher={SAGE Publications Sage UK: London, England}
}

@article{darvish2020hierarchical,
  title={Enhancing interpretability and interactivity in robot manipulation: A neurosymbolic approach},
  author={Tziafas, Georgios and Kasaei, Hamidreza},
  journal={The International Journal of Robotics Research},
  pages={02783649251412867},
  year={2026},
  publisher={SAGE Publications Sage UK: London, England}
}

@article{avizienis2004basic,
  title={Basic concepts and taxonomy of dependable and secure computing},
  author={Avizienis, Algirdas and Laprie, J-C and Randell, Brian and Landwehr, Carl},
  journal={IEEE transactions on dependable and secure computing},
  volume={1},
  number={1},
  pages={11--33},
  year={2004},
  publisher={IEEE}
}

@article{ricotti2017biohybrid,
  title={Biohybrid actuators for robotics: A review of devices actuated by living cells},
  author={Ricotti, Leonardo and Trimmer, Barry and Feinberg, Adam W and Raman, Ritu and Parker, Kevin K and Bashir, Rashid and Sitti, Metin and Martel, Sylvain and Dario, Paolo and Menciassi, Arianna},
  journal={Science robotics},
  volume={2},
  number={12},
  pages={eaaq0495},
  year={2017},
  publisher={American Association for the Advancement of Science}
}

@article{paulius2019survey,
  title={A survey of knowledge representation in service robotics},
  author={Paulius, David and Sun, Yu},
  journal={Robotics and Autonomous Systems},
  volume={118},
  pages={13--30},
  year={2019},
  publisher={Elsevier}
}

@article{loquercio2020general,
  title={A general framework for uncertainty estimation in deep learning},
  author={Loquercio, Antonio and Segu, Mattia and Scaramuzza, Davide},
  journal={IEEE Robotics and Automation Letters},
  volume={5},
  number={2},
  pages={3153--3160},
  year={2020},
  publisher={IEEE}
}

@article{lauri2022partially,
  title={Partially observable markov decision processes in robotics: A survey},
  author={Lauri, Mikko and Hsu, David and Pajarinen, Joni},
  journal={IEEE Transactions on Robotics},
  volume={39},
  number={1},
  pages={21--40},
  year={2022},
  publisher={IEEE}
}

@article{du2011robot,
  title={Robot motion planning in dynamic, uncertain environments},
  author={Du Toit, Noel E and Burdick, Joel W},
  journal={IEEE Transactions on Robotics},
  volume={28},
  number={1},
  pages={101--115},
  year={2011},
  publisher={IEEE}
}

@article{krupitzer2015survey,
author = {Christian Krupitzer and Felix Maximilian Roth and Sebastian VanSyckel and Gregor Schiele and Christian Becker},
title = {A survey on engineering approaches for self-adaptive systems},
journal = {Pervasive and Mobile Computing},
volume = {17},
pages = {184-206},
year = {2015},
note = {10 years of Pervasive Computing' In Honor of Chatschik Bisdikian},
issn = {1574-1192},
doi = {https://doi.org/10.1016/j.pmcj.2014.09.009}
}

@article{salehie2009self,
  title={Self-adaptive software: Landscape and research challenges},
  author={Salehie, Mazeiar and Tahvildari, Ladan},
  journal={ACM transactions on autonomous and adaptive systems (TAAS)},
  volume={4},
  number={2},
  pages={1--42},
  year={2009},
  publisher={ACM New York, NY, USA}
}

@misc{Rostamnia2026Replication,
  author       = {Mehran Rostamnia and Ricardo Caldas and Gianluca Filippone and Patrizio Pelliccione},
  title        = {{Adaptation Needs in Robotic Systems: Behavior Trees and Beyond -- Replication Package}},
  year         = {2026},
  howpublished = {\url{https://mehranrostamnia.github.io/adaptation-needs-bt-replication/}},
  note         = {GitHub repository, accessed Aug. 1, 2026}
}

@inproceedings{romeiro2026towards,
  title={Towards Assured Mission Adaptation of Multi-Robot Systems},
  author={Romeiro, Vicente and Caldas, Ricardo and Filippone, Gianluca and Pelliccione, Patrizio and Berger, Thorsten and Rodrigues, Genaina Nunes},
  booktitle={Proceedings of the 21st International Conference on Software Engineering for Adaptive and Self-Managing Systems},
  pages={129--135},
  year={2026}
}

@article{caldas2024runtime,
  title={Runtime verification and field-based testing for ROS-based robotic systems},
  author={Caldas, Ricardo and Garc{\'\i}a, Juan Antonio Pi{\~n}era and Schiopu, Matei and Pelliccione, Patrizio and Rodrigues, Gena{\'\i}na and Berger, Thorsten},
  journal={IEEE Transactions on Software Engineering},
  volume={50},
  number={10},
  pages={2544--2567},
  year={2024},
  publisher={IEEE}
}

@article{luckcuck2019formal,
  title={Formal specification and verification of autonomous robotic systems: A survey},
  author={Luckcuck, Matt and Farrell, Marie and Dennis, Louise A and Dixon, Clare and Fisher, Michael},
  journal={ACM Computing Surveys (CSUR)},
  volume={52},
  number={5},
  pages={1--41},
  year={2019},
  publisher={ACM New York, NY, USA}
}

@book{wohlin2012experimentation,
  title={Experimentation in software engineering},
  author={Wohlin, Claes and Runeson, Per and H{\"o}st, Martin and Ohlsson, Magnus C and Regnell, Bj{\"o}rn and Wessl{\'e}n, Anders and others},
  volume={236},
  year={2012},
  publisher={Springer}
}

\newpage
\appendices

\section{Definitions}
\label{app:definitions}

This appendix reports the definitions adopted in this paper. The purpose of these definitions is to make explicit the terminology used in the taxonomy of adaptation needs, in the mapping between adaptation needs and classical BT features, and in the characterization of adaptive BT approaches.

\begin{definition}[Classical BTs]
A classical Behavior Tree is a hierarchical control structure used to switch between different tasks in an autonomous agent. It represents behavior as a tree of modular nodes. Control-flow nodes, such as sequence, fallback, parallel, and decorator nodes, determine the execution order of action and condition nodes~\cite{colledanchise2018behavior}.
\end{definition}

\begin{definition}[BT Generation]
BT generation refers to automated or semi-automated processes that produce a BT from a higher-level specification, such as human intent, task models, natural-language instructions, demonstrations, plans, or other informal descriptions. In BT generation, no initial BT is taken as the behavior to be modified, although existing BTs, BT fragments, templates, libraries, or design patterns may be used as prior knowledge to guide the generation process.
\end{definition}

\begin{definition}[BT Extension]
BT extension refers to processes that modify the BT formalism itself. This may involve introducing new syntactic constructs into the BT grammar, changing the semantics of existing constructs, or both. Unlike generation, evolution, or refinement, BT extension changes what can be expressed within the BT language or how BT nodes are interpreted.
\end{definition}

\begin{definition}[BT Evolution]
BT evolution refers to processes that start from an initial, often minimal or partial, BT and progressively modify its structure in order to obtain a tree that encodes more complex or improved behavior. These modifications may include adding, replacing, deleting, or reorganizing nodes. Evolution is commonly associated with search-based or optimization-based techniques, such as genetic programming or grammatical evolution, where candidate BTs are iteratively improved according to a fitness criterion.
\end{definition}

\begin{definition}[BT Refinement]
BT refinement refers to processes that take an existing BT as input and update it to improve, correct, or adapt its behavior without changing the underlying BT formalism. Refinement may modify specific nodes, subtrees, parameters, or execution conditions. The resulting BT may preserve the original behavior while improving its execution, reducing it complexity, or it may intentionally update the behavior to better satisfy task, system, or environmental requirements.
\end{definition}

\begin{definition}[Knowledge]
Knowledge is a semantic, mathematical, or computational representation of the robot's beliefs about itself, its task, and the relevant parts of the physical world in which it operates~\cite{paulius2019survey}. Such knowledge may include models of the robot, objects, spatial relations, task constraints, affordances, or prior information about the environment.
\end{definition}

\begin{definition}[Perception]
Perception refers to the robotic capability of processing raw sensor data in order to construct, update, and maintain a representation of the surrounding environment~\cite{durrant2006simultaneous}. This representation may include information about obstacles, objects, robot pose, terrain, humans, or other relevant entities.
\end{definition}

\begin{definition}[Actuation]
Actuation refers to the mechanisms through which a robotic system physically affects itself or its environment. It includes robotic effectors and motion-producing components, such as drive systems, manipulators, transmissions, direct-drive motors, and grippers~~\cite{ricotti2017biohybrid,euRobotics2016MAR}.
\end{definition}

\begin{definition}[Robotic System]
A robotic system is an integrated system composed of one or more robots, sensors, actuators, power units, computational resources, and communication interfaces. These components work together to perform designated tasks autonomously or semi-autonomously within an environment~\cite{kurdila2019dynamics}.
\end{definition}

\begin{definition}[Mission]
A mission is a high-level specification of the objectives, purposes, or behaviors assigned to a robotic system~\cite{dragule2021survey}. It describes what the system is expected to achieve, possibly under given operational constraints, but does not necessarily specify how the behavior should be executed.
\end{definition}

\begin{definition}[Environment]
The environment of a robotic system is the physical, spatial, and contextual domain in which the robot operates and interacts. It includes free and occupied regions, obstacles, surfaces, objects, structural elements, and other entities that may constrain or influence motion, perception, interaction, and task execution~\cite{best2022resilient}.
\end{definition}

\begin{definition}[Uncertainty]
Uncertainty refers to incomplete, inaccurate, ambiguous, changing, or unpredictable information affecting the robotic system, its mission, or its environment. Uncertainty may arise from noisy perception, imperfect models, dynamic environments, actuator errors, communication delays, or underspecified missions. If not properly handled, uncertainty can lead to inefficient execution-time behavior, degraded performance, unsafe behavior, or robot failure.
\end{definition}

\begin{definition}[Adaptation Need]
An adaptation need is a condition indicating that the current behavior, knowledge, configuration, or mission interpretation of a robotic system is no longer adequate with respect to its mission, environment, internal state, or available knowledge. Such a need may arise when the system encounters uncertainty, failures, new constraints, changes in the environment, changes in mission requirements, or discrepancies between expected and observed execution outcomes. Addressing an adaptation need may require modifying the robot's behavior, updating its BT, revising its knowledge, adapting perception or actuation strategies, or changing the system configuration during execution.
\end{definition}

\vfill

\end{document}